\pdfoutput=1
\def\ThesisType{bc}

\def\StudyLanguage{cs}

\def\ThesisTitle{Multilingual Entity Linking Using~Dense~Retrieval}

\def\ThesisAuthor{Dominik Farhan}

\def\YearSubmitted{2024}

\def\Department{Institute of Formal and Applied Linguistics}

\def\DeptType{Institute}

\def\Supervisor{RNDr. Milan Straka, Ph.D.}

\def\SupervisorsDepartment{Institute of Formal and Applied Linguistics}

\def\StudyProgramme{Computer Science}

\def\Dedication{%
I would like to express my gratitude to my supervisor Milan Straka for his guidance, patience, and many fruitful discussions.
His thoughtful insights were invaluable not only for this work but also for my growth in the whole field of machine learning. He has my deepest thanks.

I am grateful to Ondra Sladký for reading parts of the manuscript and suggesting many helpful improvements. I owe many thanks to my girlfriend Majda Dohnalová who read every word I wrote and has always been positive and encouraging. Similarly, I would like to thank my parents and grandparents for all their support.

Last but not least, I appreciate the energy of administrators of the AIC GPU cluster who went to great lengths during one April weekend to make sure that my experiments were smoothly running.
}

\def\Abstract{%
Entity linking (EL) is the computational process of connecting textual mentions to corresponding entities.
Like many areas of natural language processing, the EL field has greatly benefited from deep learning, leading to significant performance improvements.
However, present-day approaches are expensive to train and rely on diverse data sources, complicating their reproducibility. 
In this thesis, we develop multiple systems that are fast to train, demonstrating that competitive entity linking can be achieved without a large GPU cluster. 
Moreover, we train on a publicly available dataset, ensuring reproducibility and accessibility. 
Our models are evaluated for 9 languages giving an accurate overview of their strengths.
Furthermore, we offer a~detailed analysis of bi-encoder training hyperparameters, a popular approach in EL, to guide their informed selection.
Overall, our work shows that building competitive neural network based EL systems that operate in multiple languages is possible even with limited resources, thus making EL more approachable.
}

\def\ThesisKeywords{%
entity linking\sep dense retrieval\sep entity disambiguation\sep multilingual entity linking\sep bi-encoder
}

\def\ThesisAuthorXMP{\ThesisAuthor}
\def\ThesisTitleXMP{\ThesisTitle}
\def\ThesisKeywordsXMP{\ThesisKeywords}
\def\AbstractXMP{\Abstract}

\def\InfoPageFont{\small}  %

\def\ThesisTitleCS{Vícejazyčné propojování entit pomocí vektorového vyhledávání}
\def\DepartmentCS{Ústav formální a aplikované lingvistiky}
\def\DeptTypeCS{Ústav}
\def\SupervisorsDepartmentCS{Ústav formální a aplikované lingvistiky}

\def\ThesisKeywordsCS{%
propojování entit\sep vektorové vyhledávání\sep  vícejazyčné propojování entit\sep bi-enkóder
}

\def\AbstractCS{%
Propojování entit je úloha, ve které jsou zmínky z textu propojovány s~příslušnými entitami.
Stejně jako v mnoha jiných oblastech zpracovaní přirozeného jazyka se i v propojování entit výrazně projevil vliv hlubokého učení, což vedlo k významnému zlepšení výkonu.
V současnosti se ale stávající modely trénují pomalu a spoléhají na nejednotné zdroje dat, což ve výsledku komplikuje reprodukovatelnost.
V této práci vyvíjíme několik systémů, které se učí rychle, čímž ukazujeme, že konkurenceschopných výsledků lze dosáhnout i bez velkého GPU clusteru.
Zároveň trénujeme na konkrétním veřejně dostupném datasetu.
Naše výsledky jsou tak snadno reprodukovatelné.
Modely vyhodnocujeme na devíti jazycích, což nám poskytuje kvalitní přehled o jejich silných stránkách.
Mimo to také podrobně analyzujeme nastavení značného množství hyperparametrů bi-enkóderů --- populárního přístupu pro propojování entit --- čímž zjednodušujeme rozhodování navazujícím pracem.
Náš výzkum ukazuje, že lze vytvářet silné mnohojazyčné systémy na propojování entit i za použití pouze omezených výpočetních zdrojů.
Tím činíme celou úlohu přístupnější. 
}

{
\catcode`\%=12
\global\edef\percenthack{
}

{

\def\sep{\string\sep\space}
\let~=\space

\newwrite\xmp
\immediate\openout\xmp=\jobname.xmpdata
\immediate\write\xmp{\percenthack\space Generated automatically from metadata.tex, please don't edit here.}
\def\xmpitem#1#2{\immediate\write\xmp{\string#1{#2}}}
\xmpitem\Author\ThesisAuthorXMP
\xmpitem\Title\ThesisTitleXMP
\xmpitem\Keywords\ThesisKeywordsXMP
\xmpitem\Subject\AbstractXMP
\xmpitem\Publisher{Charles University}
\immediate\closeout\xmp
}

\documentclass[12pt,a4paper]{report}
\setlength\textwidth{145mm}
\setlength\textheight{247mm}
\setlength\oddsidemargin{15mm}
\setlength\evensidemargin{15mm}
\setlength\topmargin{0mm}
\setlength\headsep{0mm}
\setlength\headheight{0mm}
\let\openright=\clearpage

\usepackage[a-2u]{pdfx}

\usepackage{lmodern}
\usepackage{iftex}
\ifpdftex
\usepackage[utf8]{inputenc}
\usepackage[T1]{fontenc}
\usepackage{textcomp}
\fi

\usepackage{amsmath}        %
\usepackage{amsfonts}       %
\usepackage{amsthm}         %
\usepackage{bm}             %
\usepackage{booktabs}       %
\usepackage{caption}        %
\usepackage{dcolumn}        %
\usepackage{floatrow}       %
\usepackage{graphicx}       %
\usepackage{indentfirst}    %
\usepackage[nopatch=item]{microtype}   %
\usepackage{paralist}       %
\usepackage[nottoc]{tocbibind} %
\usepackage{xcolor}         %

\usepackage{subcaption}
\usepackage{varwidth}
\usepackage{algpseudocode}
\usepackage{algorithm}

\hypersetup{unicode}
\hypersetup{breaklinks=true}

\usepackage{algpseudocode}  %
\usepackage{algorithm}
\usepackage{fancyvrb}       %
\usepackage{listings}       %
\usepackage{makecell}
\usepackage{multirow}
\usepackage{cleveref}
\crefname{introduction}{introduction}{introductions}

\usepackage[natbib,style=authoryear,sorting=nty,maxnames=2,minnames=1,maxbibnames=99,uniquelist=false,backend=bibtex]{biblatex}
\addbibresource{bibliography.bib}

\def\TypeBc{bc}
\def\TypeMgr{mgr}
\def\TypePhD{phd}
\def\TypeRig{rig}

\ifx\ThesisType\TypeBc
\def\ThesisTypeName{bachelor}
\def\ThesisTypeTitle{BACHELOR THESIS}
\fi

\ifx\ThesisType\TypeMgr
\def\ThesisTypeName{master}
\def\ThesisTypeTitle{MASTER THESIS}
\fi

\ifx\ThesisType\TypePhD
\def\ThesisTypeName{doctoral}
\def\ThesisTypeTitle{DOCTORAL THESIS}
\fi

\ifx\ThesisType\TypeRig
\def\ThesisTypeName{rigorosum}
\def\ThesisTypeTitle{RIGOROSUM THESIS}
\fi

\ifx\ThesisTypeName\undefined
\PackageError{thesis}{Unknown thesis type.}{Please check the definition of ThesisType in metadata.tex.}
\fi

\def\LangCS{cs}

\ifx\StudyLanguage\LangCS
\else\ifx\StudyLanguage\LangEn
\else\PackageError{thesis}{Unknown study language.}{Please check the definition of StudyLanguage in metadata.tex.}
\fi\fi

\makeatletter
\def\@makechapterhead#1{
  {\parindent \z@ \raggedright \normalfont
   \Huge\bfseries \thechapter\quad #1
   \par\nobreak
   \vskip 20\p@
}}
\def\@makeschapterhead#1{
  {\parindent \z@ \raggedright \normalfont
   \Huge\bfseries #1
   \par\nobreak
   \vskip 20\p@
}}
\makeatother

\def\chapwithtoc#1{
\chapter*{#1}
\addcontentsline{toc}{chapter}{#1}
}

\theoremstyle{plain}

\theoremstyle{remark}

\ifcsname DeclareCaptionStyle\endcsname
\DeclareCaptionStyle{thesis}{style=base,font=small,labelfont=bf,labelsep=quad}
\captionsetup{style=thesis}
\captionsetup[algorithm]{style=thesis,singlelinecheck=off}
\captionsetup[listing]{style=thesis,singlelinecheck=off}
\fi

\DefineVerbatimEnvironment{code}{Verbatim}{fontsize=\small, frame=single}

\ifcsname lstset\endcsname
\lstset{
  language=C++,
  tabsize=2,
  showstringspaces=false,
  basicstyle=\footnotesize\tt\color{black!75},
  identifierstyle=\bfseries\color{black},
  commentstyle=\color{green!50!black},
  stringstyle=\color{red!50!black},
  keywordstyle=\color{blue!75!black}}
\fi

\ifcsname DeclareNewFloatType\endcsname
\DeclareNewFloatType{listing}{}
\floatsetup[listing]{style=ruled}
\floatname{listing}{Program}
\fi

\ifx\citet\undefined\else

\ExecuteBibliographyOptions{maxcitenames=2}
\DeclareDelimFormat[textcite]{multinamedelim}{\addcomma\space}
\DeclareDelimFormat[textcite]{finalnamedelim}{\space and~}

\fi

\newcommand{\KB}{\mathit{KB}}
\newcommand{\0}{\hphantom{0}}
\newcommand{\softmax}{\operatorname{softmax}}
\newcommand{\gold}{\mathit{gold}}
\newcommand{\onehot}{\bm{1}}
\newcommand{\abr}[2]{\textbf{#1}& #2 \\}

\begin{document}
%

%

%

%

\pagestyle{empty}
\hypersetup{pageanchor=false}
\begin{center}

\centerline{\mbox{\includegraphics[width=166mm]{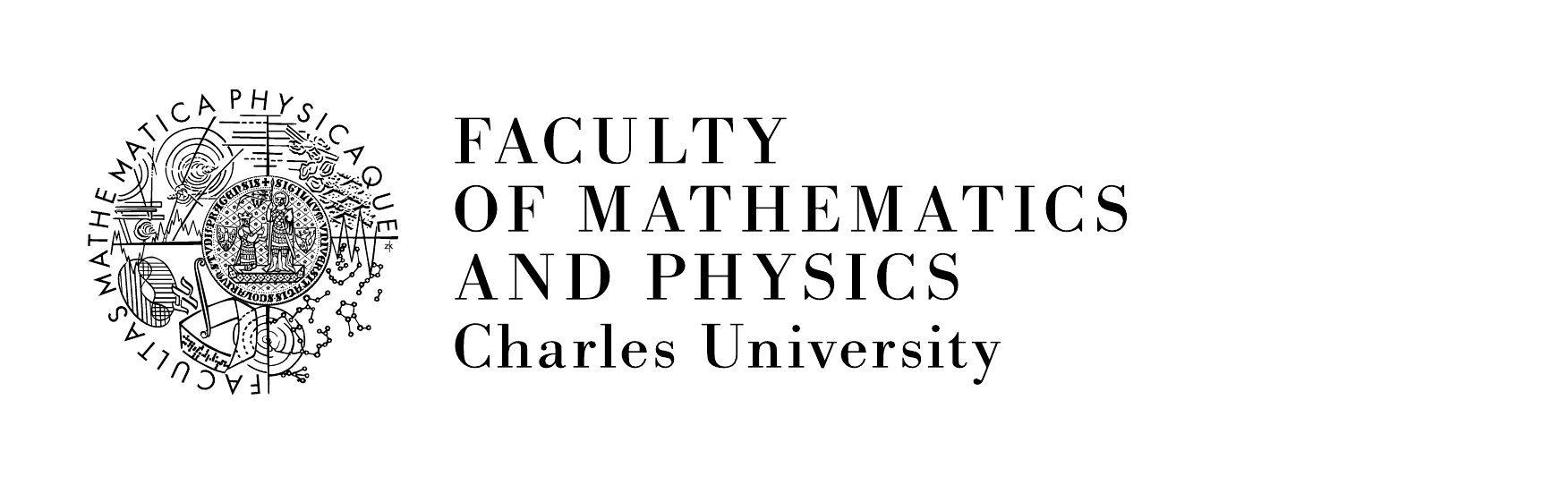}}}

\vspace{-8mm}
\vfill

{\bf\Large\ThesisTypeTitle}

\vfill

{\LARGE\ThesisAuthor}

\vspace{15mm}

{\LARGE\bfseries\ThesisTitle\par}

\vfill

\Department

\vfill

{
\centerline{\vbox{\halign{\hbox to 0.45\hsize{\hfil #}&\hskip 0.5em\parbox[t]{0.45\hsize}{\raggedright #}\cr
Supervisor of the \ThesisTypeName{} thesis:&\Supervisor \cr
\ifx\ThesisType\TypeRig\else
\noalign{\vspace{2mm}}
Study programme:&\StudyProgramme \cr
\fi
}}}}

\vfill

Prague \YearSubmitted

\end{center}

\newpage

\openright
\hypersetup{pageanchor=true}
\vglue 0pt plus 1fill

\noindent
I declare that I carried out this \ThesisTypeName{} thesis on my own, and only with the cited
sources, literature and other professional sources.
I understand that my work relates to the rights and obligations under the Act No.~121/2000 Sb.,
the Copyright Act, as amended, in particular the fact that the Charles
University has the right to conclude a license agreement on the use of this
work as a school work pursuant to Section 60 subsection 1 of the Copyright~Act.

\vspace{10mm}

\hbox{\hbox to 0.5\hsize{%
In \hbox to 6em{\dotfill} date \hbox to 6em{\dotfill}
\hss}\hbox to 0.5\hsize{\dotfill\quad}}
\smallskip
\hbox{\hbox to 0.5\hsize{}\hbox to 0.5\hsize{\hfil Author's signature\hfil}}

\vspace{20mm}
\newpage

\openright

\noindent
\Dedication

\newpage

\openright
{\InfoPageFont

\vtop to 0.5\vsize{
\setlength\parindent{0mm}
\setlength\parskip{5mm}

Title:
\ThesisTitle

Author:
\ThesisAuthor

\DeptType:
\Department

Supervisor:
\Supervisor, \SupervisorsDepartment

Abstract:
\Abstract

Keywords:
{\def\sep{\unskip, }\ThesisKeywords}

\vfil
}

\ifx\StudyLanguage\LangCS

\vtop to 0.49\vsize{
\setlength\parindent{0mm}
\setlength\parskip{5mm}

Název práce:
\ThesisTitleCS

Autor:
\ThesisAuthor

\DeptTypeCS:
\DepartmentCS

Vedoucí bakalářské práce:
\Supervisor, \SupervisorsDepartmentCS

Abstrakt:
\AbstractCS

Klíčová slova:
{\def\sep{\unskip, }\ThesisKeywordsCS}

\vfil
}

\fi

}

\newpage

\pagestyle{plain}

\tableofcontents

\chapter{Introduction}\label{chap:intro}
Humans can easily connect information to names 
and other entities mentioned within a text. 
For instance, consider the sentence, ``To save Troy from destruction, Paris had to be sacrificed.'' 
Anyone who sat through history lessons during high school can explain that this reference to Paris does not refer
to the French capital nor the socialite Paris Hilton, but rather to the mythological figure from the Iliad.
Even though, linking the mention above to the mythological figure may appear straightforward for a human, 
it is a considerable challenge for a machine to decide that it was the shepherd and not the capital that
was sacrificed.

This computational task of using a computer to establish a link between a textual mention of an entity
and its corresponding description is known as entity linking (EL).

Linking textual mentions to facts about entities is useful in many tasks,
including information retrieval and question answering~\citep{khalid, sorokin-gurevych-2018-mixing,lee2020answering, févry2020qa}, dialogue systems~\citep{Slonim2021, Joko_2021}, 
text generation~\citep{puduppully-etal-2019-data}, social media analysis~\citep{tweets2010,Adjali_2020, basaldella2020cometa}, and medicine~\citep{basaldella2020cometa, French2022-ny}.

Similarly to other tasks in natural language processing (NLP), a significant portion of EL research focuses on the English language.
This is reasonable because linking mentions requires a sizable amount of data about those entities.
With over six million articles, English Wikipedia is a good source of both structured data and unstructured texts. 
However, linking is not exclusive to English. 
There have been notable successful attempts~\citep{botha-100, decao2021multilingual, plekhanov2023multilingual} in utilizing multiple languages, indicating that the added complexity of working in multilingual contexts can yield fruitful results.
It is precisely in the domain of multilingual entity linking that our primary interest lies.

In the last few years, entity liking was conquered by deep learning, which is a subfield of machine learning that focuses on training artificial neural networks. 
These networks are great at capturing complex patterns in data that are hard to model with traditional rule-based approaches.
Moreover, the introduction of the Transformer architecture~\citep{NIPS2017_3f5ee243} allowed neural networks to capture long-term dependencies and contextual information. This is useful for linking because the model often needs to quickly process a large window of text surrounding the mention to be able to link it correctly.
Lastly, deep learning models are scalable, allowing us to use large quantities of data.
It is thus not surprising that the most competitive systems of today~\citep{Sevgili_2022} are based on neural networks.

\section*{Our Contributions}
In this thesis, we aim to develop an entity linking system capable of operating across multiple languages.
Apart from the main model we also present several baselines to establish benchmarks and validate the strength of our approach. 
A distinctive aspect of this work is reliance solely on publicly available datasets for training and evaluation, in contrast to previous studies that often withheld their training data.

The implementation and all the experiments are the sole work of the thesis author.
The main contributions are the following:
\begin{itemize}
    \item We fine-tune neural networks for entity linking with relatively limited resources, demonstrating that training competitive EL models on a single GPU in a few days is possible.
    \item Entity linking systems based on deep learning are often complex and configuring numerous hyperparameters is required. Through a series of experiments, we evaluate different parameter configurations, hoping to facilitate better decision-making in this regard and consequently make the work of others easier.
    \item We examine properties of DaMuEL~\citep{kubeša2023damuel}, a recent entity linking dataset,
    show that it can be used to train strong entity linking models and find a few shortcomings that the authors missed.
\end{itemize}

\section*{Structure of the Thesis}

The work is divided into five chapters. 
In \Cref{chap:refs} we provide the necessary background: we build a solid foundation of entity linking and survey recent approaches and popular datasets. 
In \Cref{chap:baselines} we describe three simple improvements to \emph{alias tables}, a popular baseline. Additionally, we establish an upper bound on results one can achieve when training with DaMuEL dataset~\citep{kubeša2023damuel} and evaluating on Mewsli-9~\citep{botha-100}.
In \Cref{chap:finetuning}, we discuss peculiarities associated with training deep neural networks for our task.
We describe a well-known approach to entity linking based on \emph{bi-encoder} models~\citep{gillick-etal-2019-learning,botha-100,wu-etal-2020-scalable,fitzgerald2022moleman} and introduce several modifications.
In \Cref{chap:infra}, we give a slightly technical overview of the infrastructure we built for our experiments.
In \Cref{chap:res}, we evaluate systems from \Cref{chap:baselines,chap:finetuning}.
We also compare them to results from other works and conduct many experiments showing how different parameter settings influence performance.

\chapter{Existing Approaches in Entity Linking}
\label{chap:refs}
\section{Definitions}

To understand Entity Linking (EL), we first need to establish some key terminology.
\subsection{Mention} 
A specific text for which we want to generate a link is called a \textit{mention}, and we use the term \textit{context} to denote the mention together with the surrounding text.
To perform a linking, one ideally needs both
\begin{itemize}
  \item a mention $m$, which is a substring directly corresponding to the entity; and
  \item a context $c$, sometimes also denoted as mention context. In our work, the context is always the mention $m$ and the text surrounding it, and it can be summed up with the maxim ``You shall know your mention by the company it keeps.''
\end{itemize}

To clarify the concept, let us recall the mythological Paris from \Cref{chap:intro} and see how it fits our representation:
\[
m = \text{``Paris''},\, c = \text{``To save Troy from destruction, Paris had to be sacrificed.''}
\]

Generally, we assume that $m$ and $c$ are part of a larger document $d$. Sometimes it is beneficial to contextualize taking $c = d$, but for larger $d$ this is impractical.
Another popular option is to combine the surrounding text with information specific to $d$ (like the title of $d$).
Generally, the most important information needed for linking $m$ is likely close to it. 
Hence, it's best to think of context as a paragraph, a sentence, or some other data that helps to describe $m$ based on $d$.
Nonetheless, how exactly the context looks heavily depends on the approach used.

It is crucial to realize that both $m$ and $c$ are needed to create a robust linking system. 
In theory, we can attempt to link only based on $m$ by looking up an entity with the same or similar name using some string similarity metric. 
However, in practice, this approach is inadequate as $m$ is often ambiguous. Nevertheless, systems utilizing only mentions without contexts can still serve as solid baselines, and when they are provided with enough training examples, they might surpass some more sophisticated approaches.

It is only natural to ask what would happen if we omitted $m$ and tried to link just based on $c$.
Take again the text
\[
\text{``To save Troy from destruction, Paris had to be sacrificed.''}
\]
The linking system wouldn't discern whether we intend to link \textit{Paris},
\textit{Troy}, or something else entirely.

Note that while it is conceptually helpful to think of $m$ and $c$ as a pair that we require to perform the linking,
in practice, we prefer a more concise representation, where the pair is represented just by one item. For example,
\begin{center}
\begin{varwidth}{\textwidth}
``To save Troy from destruction, \verb|[M]|Paris\verb|[M]| had to be sacrificed.''
\end{varwidth}
\end{center}
where \verb|[M]| is a special token that we add to the tokenizer. 

\subsection{Knowledge Base}
For linking a mention, we first need an entity to which we can link.
A~structure that holds information on all possible entities is called a \emph{knowledge base (KB)}.
Typically, it contains for each entity: a label, an ID, and structured and unstructured data that describe it.
In many studies, an important part of KB is a so-called \emph{knowledge graph (KG)} --- for example Freebase~\citep{freebase}, or Wikidata~\citep{wikidata2014}.
KG holds structured knowledge about entities.
Information is typically represented as \emph{claims}, which contain properties and relations to other entities. 
Because relations to other entities can be viewed as typed, directed edges, the structure is called a \emph{knowledge graph}.

In our work, we limit ourselves to working with the unstructured part of KB.
We are not interested in knowledge graphs.
Instead, we compile a set of \textit{textual descriptions} for each entity $e = (d_1, d_2, \dots, d_n)$.
Generally, these descriptions are independent chunks of text, and their number per entity can vary based on the system.
One can imagine them for example as Wikipedia pages in different languages.\footnote{\url{https://www.wikipedia.org/}}
We chose this representation for three primary reasons:
\begin{itemize}
    \item It is simpler to gather unstructured textual descriptions of an entity than a set of relations.
    \item Using text descriptions grants our models more autonomy in determining
      the significance of information. The idea is that our system learns to construct important relations by itself.
    \item Even someone with a limited understanding of our system can easily add
      new descriptions and, consequently, new entities to the KB.
\end{itemize}

This narrowing to only textual descriptions is not a novel approach, there are numerous other works utilizing it \citep{gillick-etal-2019-learning,botha-100,fitzgerald2022moleman}.

\subsection{What is Entity Linking?}
With the above-described formalism, it is perhaps the right time to ask ``What exactly is entity linking?''

The answer is anything but simple. 
In a fairly broad view, entity linking is about learning to map mentions to entities in a knowledge base.

However, to make this a usable definition, we need to answer the following two questions:
\begin{enumerate}
    \item Are the mentions given to us (for example as spans in a document) or do we have to find them by ourselves?
    \item What happens when the entity corresponding to the queried mention is not in the KB?
\end{enumerate}

\subsection{End-to-End Entity Linking}
To answer the first question, we need to understand what is \textit{mention detection (MD)} and \textit{entity disambiguation (ED)}.
The primary objective of MD is to detect possibly all entity mentions in a given document.
Subsequently, ED connects mentions to entities in a knowledge base.
Traditionally, entity linking is composed of those two components, and we denote this composition as \emph{end-to-end entity linking} in our text.
Nevertheless, the vast majority of EL works concentrate only on the second step ED~\citep{Sevgili_2022}.

In this thesis, we do not focus on MD.
Mentions are already detected in our data, so we only disambiguate them.
Thus, we use \emph{entity linking} and \emph{entity disambiguation} interchangeably.

\subsection{NIL entity linking}
To answer the second question, consider the following two scenarios. During linking, it might happen that:
\begin{itemize}
    \item an entity corresponding to a mention is not present in a KB, or
    \item that MD system made a mistake and a span of text it annotated does not correspond to a mention of an entity.
\end{itemize}

In both of these cases, it could be useful to have the ability to predict a so-called \textit{NIL} entity, which signals that we cannot link the input to any entity in a given KB.

Similarly to end-to-end entity linking, the problem of NIL entities is underrepresented in research. Part of this is due to its fragmentation. How NIL entities are treated varies between systems and datasets, and often they are simply not considered~\citep{zhu2023learn}.

Our evaluation dataset and the primary metric we decided to use have no way to accommodate NIL entities, so their treatment is out of our scope.
Nevertheless, all the systems we describe in \Cref{chap:baselines} and \Cref{chap:finetuning} could be extended to predict them.

Let us sum up the above paragraphs. End-to-end entity linking contains two steps
\begin{itemize}
    \item \emph{mention detection}, and
    \item \emph{entity disambiguation}.
\end{itemize}
The datasets we work with contain already annotated mentions, so we focus only on the second step.
Also, they do not contain examples of NIL entities, thus we always assume that a query can be linked.

\subsection{Multilinguality}\label{subsec:multilinguality}
The last piece of our puzzle is multilinguality.
EL is most often performed only for one language.
To be more specific, let $L^{\KB}$ be the set of all languages used in a KB and let all mentions $M$ come from a context language $l^c$.
A~large part of current research assumes that $L^{\KB} = \{l^c\}$, and the one language that is used above all others is English.

One attempt to accommodate different languages is \emph{cross-lingual EL} (XEL). 
It assumes that $L^{\KB} = \{l\}$ and $l \neq l^c$. 
Commonly, the language of the KB is English. 
Although XEL allows for using more than one language, it is still heavily limited by the choice of $l$.
That is why \emph{multilingual entity linking}~\citep{botha-100} was introduced.
Its goal is resolving mentions $M$ without any assumptions on the relation between $L^M$ (languages of $M$) and $L^{\KB}$.

This setup is useful because it gives us much more freedom to create a KB.
For example, there are many entities, for which we can find detailed descriptions in some uncommon language and only brief or none in English.
On the other hand, linking between multiple different languages brings its own new challenges.

\section{Data}
There are three kinds of data we require in order to build and evaluate an EL system: a KB, train data, and test data.
Below, we describe some notable options with special emphasis on multilingual EL.

\subsection{Wikidata}
\label{subsection:wikidata}
\begin{figure}[t]
\centering
\includegraphics[width=.6\linewidth]{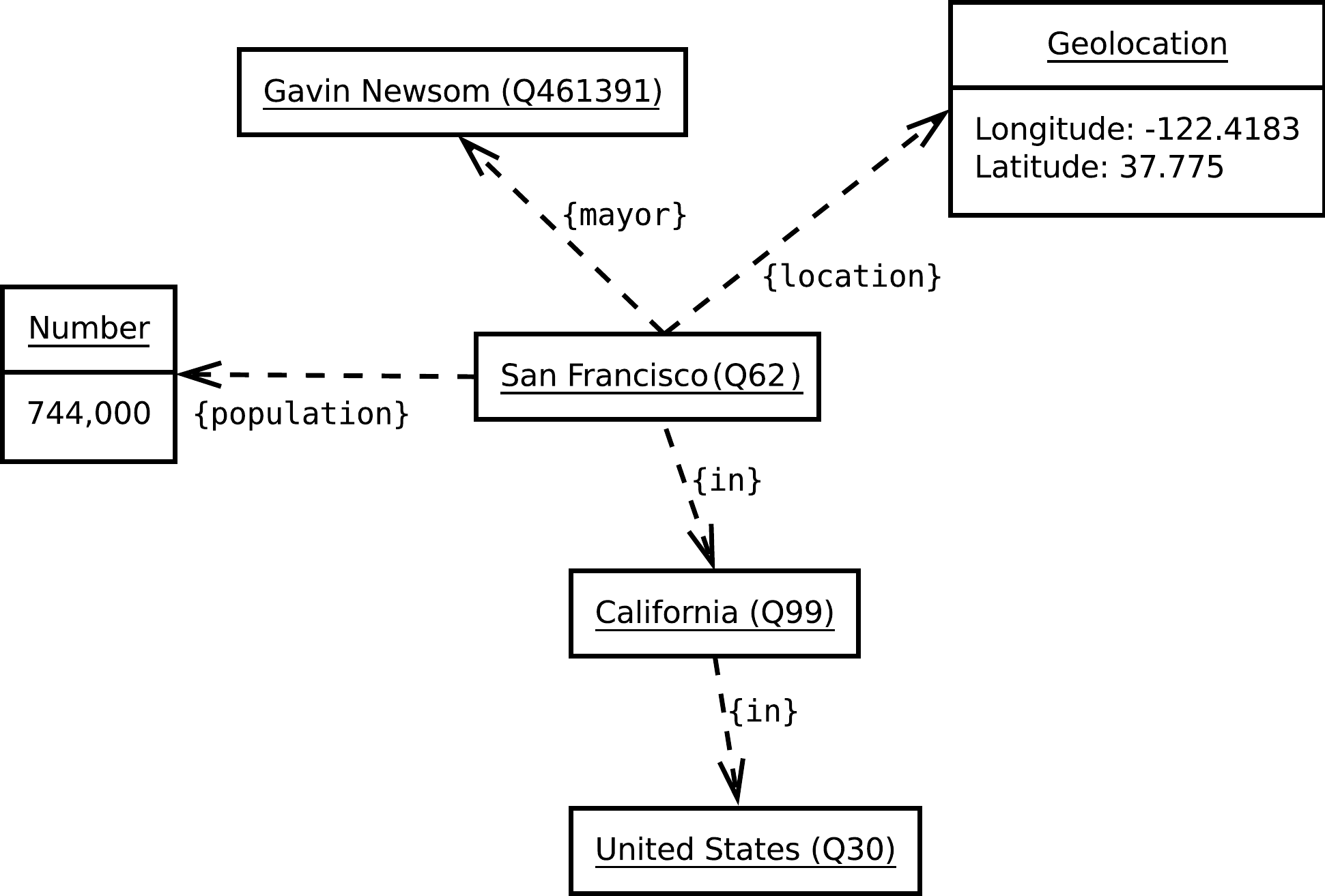}
\caption{An example of Wikidata statements for San Francisco. By Jeblad, CC BY-SA 3.0, via Wikimedia Commons,
\url{https://commons.wikimedia.org/wiki/File:Linked_Data_-_San_Francisco.svg}.}
\label{fig:sf_statements}
\end{figure}

Wikidata is a free, multilingual knowledge base built on top of Wikipedia.
It is the backbone of many EL datasets.
It consists of a staggering 109M entities.\footnote{\url{https://www.wikidata.org/wiki/Wikidata:Statistics}}\footnote{Although in Wikidata entities are denoted as \textit{items}, we decided to use our terminology for consistency with the rest of the text.}
Each entity contains the following fields:
\begin{itemize}
    \item a \emph{label}, usually the most common name (for example \textit{San Francisco});
    \item a short \emph{description} (\textit{consolidated city and county in California, United States});
    \item a unique identifier, a so-called \emph{QID} (\textit{Q62});
    \item optional aliases, alternative names (\textit{The City by the Bay, SF, \dots});\footnote{We also use the term \emph{alias} in other places to mean \textit{an alternative entity name} but not necessarily \textit{an alternative entity name from Wikidata}.} 
    \item optional \emph{statements}, which connect the entity to various facts and other entities in a language-agnostic way.
    A statement consists of a property-value pair (\textit{population: }\textit{$805704$}).
    A property-value pair might have additional qualifiers that specify it further (\textit{point-in-time: 1 July 2010}). See \Cref{fig:sf_statements} for an example with multiple statements).
\end{itemize}

\subsection{Mewsli-9}
Mewsli-9 is the de facto standard benchmark in present-day multilingual EL.
It has been introduced by \citet{botha-100} in the same work in which they also proposed the multilingual EL.
The dataset is built on top of Wikinews in 9 languages.\footnote{\url{https://www.wikinews.org/}}
The authors purposefully use a varied set of languages encompassing five language families and six orthographies.
Mewsli-9 contains the following languages: Arabic (ar), German (de), English (en), Spanish (es), Persian (fa), Japanese (ja), Serbian (sr), Tamil (ta), and Turkish (tr).

Authors build their system in a way that enables ranking entities based on a similarity to a queried mention. 
They suggest using recall-at-k ($R@K$) to evaluate their system for each of the nine languages separately.
To define this metric, let  $M^l$ be the set of Mewsli-9 mentions in language $l$, and $g(m)$ the correct entity for the mention $m$. 
Denote the $K$ most similar entities according to the examined system as $\varphi(m, K)$, and let $[x \in S]$ evaluate to $1$ when $x$ is in $S$ and $0$ otherwise. Now, we define $R@K$ as
follows:
\[
  R@K = \frac{\displaystyle\sum\nolimits_{m\in M^l}\big[g(m) \in \varphi(m, K)\big]}{\displaystyle\big|M^l\big|}.
\]

Thus, to get the $R@K$, we calculate the ratio of mentions recognized as the $K$ top probable by the system.
The paper evaluates $R@1$ and $R@10$.

Interestingly, the authors decided to allow linking mentions to disambiguation pages.
These pages are part of Wikipedia, and their purpose is to resolve conflicts between articles with the same or very similar titles. 
The purpose of EL is to \textit{disambiguate} mentions; it is not clear how disambiguating to disambiguation pages achieves that goal.
We return to this problem and how it affects our work in \Cref{section:damuel_mewsli_intersection}.

\subsection{TAC KBP}
The Text Analysis Conference included EL in their knowledge base population challenge multiple times. 
Several datasets sprung from those years.
\textit{The TAC KBP Reference Knowledge Base}~\citep{TACsimpson2014} is built from English Wikipedia and consists of more than 800k entities.
Each entity consists of a canonical name, a title for the Wikipedia page, a type, data from its infobox, and a stripped version of the text of the Wikipedia article.
It can be used for monolingual linking, but also easily for cross-lingual because TAC also provides training and evaluation data with queries that link to a KB in Chinese~\citep{TACChinese2016} and Spanish~\citep{TACSpanish2016}.

\subsection{AIDA CoNLL-YAGO}
AIDA CoNLL-YAGO is a dataset from a seminal work by~\citet{hoffart-etal-2011-robust}.
It is based on the English part of the 2003 CoNLL named entity recognition shared task~\citep{tjong-kim-sang-de-meulder-2003-introduction}. 
The dataset consists of annotations of $1393$ Reuters articles. 
Proper nouns from these articles are linked to entities from YAGO2 \citep{yago1,HOFFART201328}, a  KB built on Wikipedia, WordNet, and GeoNames.

The dataset consists of 3 parts: train, testa, and testb. The first two are used for training and validation, the third serves as a test set.

\subsection{DaMuEL}
\label{subsec:damuel}
DaMuEL~\citep{kubeša2023damuel} is a large (approx 1 TB) multilingual dataset spanning $53$ languages based on Wikidata and Wikipedia. It consists of two components: \emph{language agnostic part} and \emph{language specific part}.

The agnostic part contains information that is the same across all languages, whereas the specific part consists of $53$ sub-parts (one for each language).
The specific part is the one we are interested in.
It contains Wikidata labels and descriptions, Wikipedia pages, and mentions annotated with QIDs to which they link.

The mentions in DaMuEL are constructed in one of two ways. 
Either they were already present in Wikipedia as a form of hyperlink that links from one page to another, or they were expanded by the authors.
The Wikipedia Manual specifies that only the first occurrence of an entity should be linked.\footnote{\scriptsize\url{https://en.wikipedia.org/wiki/Wikipedia:Manual_of_Style/Linking\#MOS:REPEATLINK}}
However, one page can contain many more mentions of the same entity. 
The authors came up with a heuristic that adds new links based on an already linked mention.
The heuristic takes an article with an already linked mention, finds the entity of that mention in Wikidata, and matches strings inside the article that are the same (by exact or lemma match) as either the mention or names from Wikidata (labels, aliases, etc).
This heuristic is used only for named entities, because for other entities it could produce a huge number of false positive links.
Alongside constructing the dataset, the authors performed morphological analysis and also found types for named entities.

\subsubsection{Specific parts}\label{subsubsec:damuel_spec}
Below, we describe the general structure of language-specific parts of DaMuEL because we use them as the training data for all our models.
Each part consists of $500$ .xz archives of approximately the same size.
All parts are in the same format.
Each line contains data on one entity as key-value pairs serialized in JSON Lines.\footnote{\scriptsize\url{https://jsonlines.org/}}
Below, we list the keys most relevant to our work:
\begin{itemize}
    \item \textit{QID}, \textit{label}, and \textit{description}; all from Wikidata
    \item \textit{wiki}; present only when the entity has a Wikipedia article in the given language. Takes the form of a dictionary. We again list the most relevant keys:
    \begin{itemize}
        \item \textit{title}, and \textit{text}; from the Wikipedia article,
        \item \textit{tokens};
        \item \textit{links}; each containing a mention span, a QID, and a source (information whether it is an original wikilink or it was created by the authors).
    \end{itemize}
\end{itemize}

\section{Neural Entity Linking}\label{section:neural_el}
The deep learning revolution of the past decade brought several advancements to EL, and in the last few years, non-neural approaches stopped being competitive~\citep{Sevgili_2022}.
In this section, we cover some notable multilingual approaches that are the results of these advancements.

\subsection{Bi-encoders}
\label{subsec:be-encoders}
\begin{figure}
\centering
\includegraphics[width=.6\linewidth]{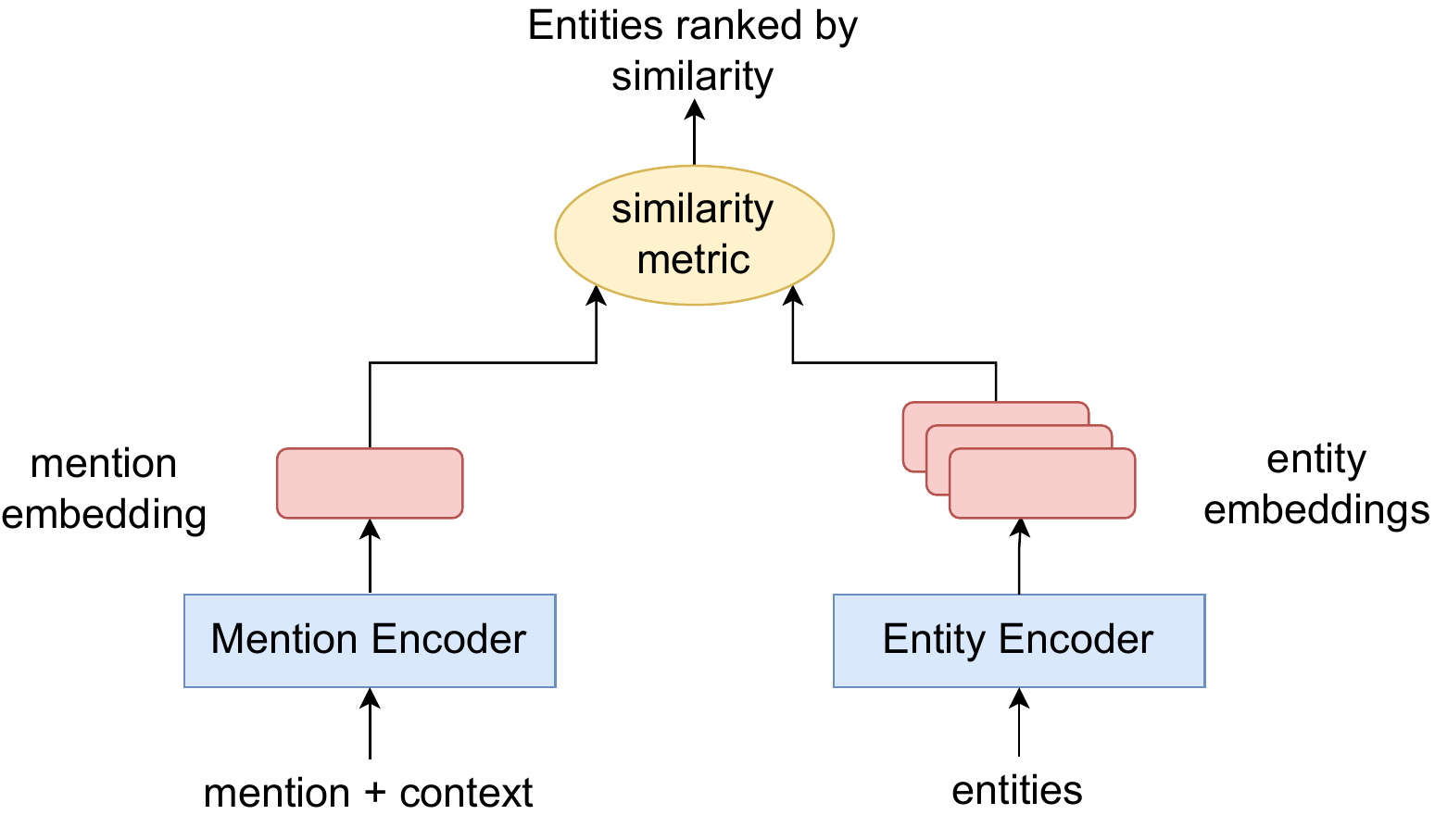}
  \caption{A general layout of bi-encoder models used in EL.}
\label{fig:bi_encoders}
\end{figure}

Many ED approaches use deep learning models to encode both entities and mentions into high-dimensional vectors using the bi-encoder architecture (see \Cref{fig:bi_encoders}).

These fixed-length high-dimensional vectors are called \emph{embeddings}.
Models are trained to produce similar embeddings for semantically similar texts.
Embedding similarity is most often defined as the angle between them.
This allows for efficient semantic comparison between two texts.
Nowadays, many models produce embeddings that work across languages.
Thus, a sentence in one language is embedded in a vector very similar to one generated from the same sentence in a different language.

In EL that utilizes bi-encoders, two models are learned: one to encode mentions based on the context and the other to encode entities.
The entity encoding models sometimes combine a vast number of different facts because KBs often carry much more information than just textual descriptions.

The cosine of the angle between an entity and a mention is used to measure the similarity.
Let $\bm{e}_m$ and $\bm{e}_e$ be the embeddings for a mention and an entity, respectively, and let $\|\bm{e}\|$ denote the norm of an embedding $\bm{e}$. 
Then we can calculate the cosine of the angle between $\bm{e}_m$ and $\bm{e}_d$ as
\[
    \cos(\bm{e}_m, \bm{e}_e) = \frac{\bm{e}_m ^T \bm{e}_e}{\|\bm{e}_m\| \cdot \|\bm{e}_e\|}.
\]

One particular strength of bi-encoders is that embeddings of entities can be calculated beforehand; therefore, one only needs to calculate the $\bm{e}_m$.
Finding an entity most similar to a given $\bm{e}_m$ can then be done quickly using a highly optimized similarity search library like FAISS~\citep{douze2024faiss} or ScaNN~\citep{avq_2020}.

\subsubsection{Learning Dense Representations for Entity Retrieval (DEER)}
\citet{gillick-etal-2019-learning} created Dual Encoder for Entity Resolution (DEER) and showed that bi-encoders are a viable approach for EL in English.
They successfully utilize in-batch-sampled softmax~\citep{jozefowicz2016exploring}. 
To provide the negatives to the softmax, the authors sample the negatives randomly~\citep{henderson2017efficient,gillick2018endtoend}, but later, as the learning progresses, the model itself is used to generate negatives that are more challenging than random ones.
The authors call their method hard negative mining.

We also employ in-batch-sampled softmax and hard negative mining, thus, we decided to describe it together with our work in~\Cref{chap:finetuning}.

\subsubsection{Entity Linking in 100 Languages (EL100)}\label{subsub:el100}
\citet{botha-100}, the same paper that introduced Mewsli-9, build on ideas from DEER.
Hard negative mining as well as in-batch-sampled softmax are used.
The bi-encoder utilizes multilingual BERT~\citep{devlin2019bert} --- a transformer-based neural network~\citep{NIPS2017_3f5ee243} --- to embed textual descriptions of entities and mentions with contexts.

Each entity is represented by exactly one description.
To obtain it, the authors create candidate descriptions from Wikipedia pages or from Wikidata descriptions when a Wikipedia page does not exist. 
This yields multiple possible descriptions whenever an entity has a Wikipedia/Wikidata entry in multiple languages.
To choose the best of those, the authors use a simple heuristic:
they define $n_e(l)$ to be the number of mentions of the entity $e$ in the language $l$, and $n(l)$ the number of mentions in $l$.
Then, they order candidate descriptions for $e$ by $n_e(l)$.
If this produces a tie (because there are multiple languages with the same number of mentions of $e$), they use $n(l)$ to break it and take the description corresponding to the language with higher $n(l)$.

\subsubsection{MOLEMAN}
\citet{fitzgerald2022moleman} continue in the footsteps of EL100 but stop differentiating between entity descriptions and training data contexts.
Suddenly, contexts are part of the KB, and it is possible to link to them.
The idea is that many mention contexts provide a better entity characterization than having only one high-quality description.
The resulting system, called MOLEMAN, beats EL100 on Mewsli-9, but it comes with the heavy burden of a larger memory footprint.

\subsection{(m)GENRE}
GENRE (short for Generative ENtity REtrieval) by~\citet{decao2021autoregressive} is an approach different from the prevalent bi-encoders.
The authors propose to use autoregressive generation to produce unique entity labels.
They assign score to each entity $e$ conditioned on mention and context pair $x = (m, c)$ as follows:
\[
s(e \mid x) = p_{\theta}(n \mid x) = \prod_{i=1}^{|n|} p_{\theta}(n_i \mid n_{<i}, x),
\]
where $n$ is the unique identifier of $e$, $n_i$ its $i$-th token, and $\theta$ are learned parameters of the model.

Calculating the above for all entities given $x$ would make the inference expensive.
Therefore, the authors propose using beam search --- a heuristic search algorithm that traverses a search tree in a breadth-first-search fashion but expands no more than $k$ most promising nodes per level. 

Additionally, to the usage of a beam search, the authors also constrain the search to only generate valid identifiers.
To quickly decide whether a given path in a search space corresponds to a valid identifier, they store all known identifiers in a trie.
The trie occupies usually hundreds of MBs or a few GBs of space, which is a significant improvement compared to bi-encoder models, where the index with all entity embeddings often spans tens of GBs.

The GENRE paper also describes how to extend their system to end-to-end EL. However, because we focus only on the disambiguation part of the end-to-end formulation, we do not describe it here.

The same month as the GENRE paper was published, \citet{decao2021multilingual} also proposed its multilingual formulation called mGENRE.
To extend GENRE to mGENRE, the authors compare multiple strategies that aim at utilizing entity names from different languages. 
Overall, the best-performing option seems to be allowing the model to generate not only the name but also its language (for example \textit{Sistema de posicionamento global-ES} for GPS in Spanish; both the name and language code are generated). 
Each entity can then have multiple identifiers that consist of a name in a specific language together with the language identifier.
Predicting not only the name but also the language is useful because one name can point to different entities in different languages.
The GENRE formula is then changed to
\[
s(e \mid x) = \sum_{n^l \in I_e} p_\theta(n^l, l \mid x),
\]
where $I_e$ is the set of all per-language-unique names $e$.

The mGENRE system shows strong results on several datasets and is the current state of the art for Mewsli-9.

\section{Text Embedding Models}
\label{sec:models}
\citet{feng2022languageagnostic} show that the combination of dual encoder finetuning,
additive softmax, and a large language model can yield excellent results when producing sentence embeddings.
The result of their research is LaBSE, which can produce high-quality sentence embeddings for more than
$109$ languages.

Finetuning BERT to produce sentence embeddings takes significantly fewer resources than attempting to achieve the same feat
from scratch. 
If we limit our EL data only to text, our problem starts to resemble that of producing a sentence embedding.
We can look at mentions with contexts and entities with descriptual texts as large sentences that we need to embed, 
so that mentions lie close to their corresponding entities.
Because fine-tuning pre-trained language models to produce sentence embeddings saves substantial amount of resources,
it is only natural to ponder whether the same can be achieved when fine-tuning sentence embedding models
to disambiguate entities.

It is tempting to fine-tune LaBSE for our task. Disappointingly, the large number of parameters (471M) and the $756$-di\-men\-sional embedding size make working with it complicated given our limited resources.
In the end, we decided to settle for the LEALLA family of models~\citep{mao2023lealla} for our fine-tuning experiments.
This family contains three models: small, base, and large (see~\Cref{tab:lealla_overview} for comparison with LaBSE).
All of these much smaller models were produced by knowledge distillation from LaBSE.

\begin{table}[h]
  \centering
  \begin{tabular}{lcc}
    \hline
    Model & Parameters  & Embedding dimension \\
    \hline
    LEALLA-small & \069M & 128 \\
    LEALLA-base & 107M&  192 \\
    LEALLA-large & 147M& 256 \\
    LaBSE & 471M& 756  \\
    \hline
  \end{tabular}
  \caption{LEALLAs and LaBSE: parameters and embedding sizes.}
  \label{tab:lealla_overview}
\end{table}

Note that although LEALLA models are smaller than LaBSE, they are not necessarily faster. 
See~\Cref{sec:leallavslabse} for comparison.

\chapter{Baselines}
\label{chap:baselines}
Before we start finetuning bi-encoders for EL, we present several simpler models.
We do so for two reasons:
\begin{itemize}
  \item As far as we know, DaMuEL has never been used in a publicly available EL system. We aim to explore its strengths and weaknesses by evaluating it with models whose strength is not dependent on the particular fine-tuning set up.
    \item We can validate our advanced results by using those from this chapter as baselines.
\end{itemize}

\section{Upper Bounds on Mewsli-9 Recall When Using DaMuEL}
\label{section:damuel_mewsli_intersection}
Before we build our first EL system, we establish an upper bound on recalls we can get when using DaMuEL as a KB and linking from Mewsli-9.
In doing so, we treat each mention from Mewsli-9 separately and check whether it has an entity in DaMuEL.
This allows us to establish an upper bound on $R@K$.
We provide a per-language breakdown of upper bounds in \Cref{tab:damuel_mewsli_intersection}.
We give two columns of results.
In the first column, we consider all DaMuEL entities.
In the other column, only the entities from the specific DaMuEL part (\Cref{subsubsec:damuel_spec}) corresponding to the language are considered.
The differences in upper bounds between the two columns are negligible.
This leads us to the conclusion that when linking a Mewsli-9 language, we can use only entities from the corresponding DaMuEL part and still achieve high recalls.

\begin{table}[t]
  \centering
  \begin{tabular}{lcc}
    \hline
    \makecell[c]{Mewsli-9\\language} & \makecell[c]{Mentions from\\all [\%]} & 
    \makecell[c]{Mentions from\\the specific part [\%]}  \\
    \hline
    ar & 95.6 & 95.6\\
    de & 96.1 & 95.9\\
    en & 94.3 & 94.3\\
    es & 95.3 & 95.0\\
    fa & 98.3 & 98.3\\
    ja & 96.8 & 96.7\\
    sr & 97.4 & 97.3\\
    ta & 98.6 & 98.6\\
    tr & 96.0 & 96.0\\
    \hline
  \end{tabular}
  \caption{Upper bounds on recalls when linking Mewsli-9 to DaMuEL. In the second column, we use all DaMuEL entities, in the third only entities from the corresponding language-specific part are used. Our results show that the vast majority of entities from a particular Mewsli-9 language can be linked to entities in the corresponding DaMuEL counterpart but not all of them.}
  \label{tab:damuel_mewsli_intersection}
\end{table}

Another observation we make is that we cannot achieve a perfect recall on any of the languages. 
This means that some entities are missing from DaMuEL.
Subsequently, we discovered that if we take the \textit{set} of all entities of DaMuEL and the \textit{set} of all entities of Mewsli-9, only $93 \%$ of the entities from the second set are in the first.
A gap of $7 \%$ cannot be explained by the fact that the datasets were gathered three years apart. 
In fact, DaMuEL is three years younger, so we would expect it to contain all the Mewsli-9 entities and more.
The $5$ lowest missing QIDs are Q83, Q86, Q182, Q514, and Q569.
Some of the QIDs are disambiguation pages (like \url{https://www.wikidata.org/wiki/Q182}), while others are not.
We discussed this with one of the authors of DaMuEL, who explained that the omission of disambiguations is intentional; however, the disappearance of the other entities is a mistake that they hope to correct soon.
To quote from a conversation with Milan Straka (one of the authors):
\begin{quote}
    One of the things
    that we deliberately didn't put in DaMuEL are disambiguations. 
    For one thing, they're not
    real-world entities (although they are entities in terms of Wikipedia),
    and there shouldn't be reasonable mentions of them in the wiki (because the real
    mentions in the text should lead to a specific entity, not a disambiguation
    one level above it).
\end{quote}

Disambiguation pages are only about $69$\% of missing QIDs, thus we must look for another reason for the missing entities.
After further deliberation with the authors, we concluded that these pages are missing
because the process that removed non-real-world entities was a bit too aggressive.
Both the missing disambiguations and mistakenly removed entities should be addressed in a new release, which is, however, not yet available.

Nevertheless, our work's crucial aim is to create an EL system trained with freely and easily available data. 
DaMuEL satisfies this property. 
Additionally, because there are plans to address the aforementioned issues, and we already have plenty of experience with it from 
our prior work on EL,\footnote{\url{https://github.com/ufal/linpipe/blob/kbelik/doc/dev/kbelik/specification.md}} we decided to use it as our training dataset regardless.

\section{General Picture of Entity Linking System}\label{sec:general_framework}
In a bird's eye view, many EL systems (and all those we describe in this and the following chapter) can be seen as specific instances of a simple yet general framework.
This framework consists of two components: 
\begin{itemize}
    \item a KB that holds information on entities,
    \item and similarity function $s$ which scores entities and queries.
\end{itemize}
The most important part of designing a good EL system is to choose a suitable KB and $s$.

Although this general framework is simplistic, we encourage readers to bear it in mind when studying the systems that follow.

\section{Alias Table}\label{section:alias_table}
An alias table is a map connecting an entity name (alias) to an entity.
Systems based on them are simple yet effective.
Nowadays, they are mostly used as baselines or to retrieve candidates for more sophisticated approaches.
They are built on the observation that entity names are often unique.

Let us frame this system in the general view presented in \Cref{sec:general_framework}.
For each entity, the KB contains a list of possible names.
We call these names \emph{aliases}.
The scoring function $s$ simply checks whether the entity has an alias that exactly matches the mention.

Recall that in~\Cref{subsection:wikidata} we explain that Wikidata uses the term \textit{alias}
to mean an alternative entity name.
Here, we use an alias in the same way, but we would like to emphasize that our aliases are \textit{not} based on those from Wikidata.
Instead, to create the aliases we iterate over the training data and gather for each entity all mentions that link to it.
This is better because, with large enough training data, we obtain more aliases than if we utilized those already present in Wikidata.

In practice, the systems based on alias tables do not represent KB as a map that connects entities to a list of aliases, but as a map that connects aliases to their entities.
This allows for efficient linking as shown in~\Cref{alg:alias_table}.
\begin{algorithm}[t]
\begin{algorithmic}
\Function{LinkFromAliasTable}{$m$, $\mathit{table}$}
	\If{$m \in \mathit{table}$}
		\State \Return $\mathit{table}[m]$
	\Else
        \State \Return $\mathit{NIL}$
	\EndIf
\EndFunction
\end{algorithmic}
\caption{Link a $m$ to its alias table entity, return $\mathit{NIL}$ when linking fails.}
\label{alg:alias_table}
\end{algorithm}

Thus far, we have not discussed what happens when an alias corresponds to more than one entity.
However, in reality, this happens quite often.
Nevertheless, the solution is simple.
Assuming that the train and test data contain a similar distribution of alias-entity occurrences, we decide to prioritize entities that appear more often under the given alias.
For example, when training on Wikipedia, Paris occurs more often as the city than the Troyan prince.
We then assume that the same holds for the test.
Thus, when an alias table is asked to give an entity for the mention \textit{Paris}, it prioritizes the city over the prince.

You may recall that all the systems mentioned in the part on Neural Entity Linking~(\Cref{section:neural_el}) could retrieve multiple entities and rank them.
Yet, the above-described alias table returns at most one entity.
To allow for a fair comparison with other models, we extend alias tables to retrieve more than one entity.
To do so, the condition that each alias maps to just one entity is relaxed.
Instead, we say that it maps to a collection of entities of size at most $K$, where $K$ is some predefined constant.
This allows us to easily evaluate not just $R@1$ but also any $R@K$.
The construction is described in~\Cref{alg:alias_table_construct}.

In \Cref{subsec:at_damuel}, we use alias tables to compare DaMuEL and Mewsli-9.

\begin{algorithm}[t]
\begin{algorithmic}
\Function{GetAliasEntityCounts}{$d$} \Comment{Expects train data as alias-entity pairs.}
    \State $\mathit{counts}$ $\gets$ map with counters
    \For{$(a, e) \in d$}
        \State $\mathit{counts}[a]\mathit{.Increase}(e)$
    \EndFor
    \State \Return $\mathit{counts}$
\EndFunction
\Statex
\Function{ConstructTableFromCounts}{$\mathit{counts}$, $K$}
    \State $\mathit{table}$ $\gets$ empty map
    \For{$(a, \mathit{counter}) \in \mathit{counts}$}
        \State $\mathit{table}[a]$ $\gets$ $\mathit{counter.GetTop}(K)$
    \EndFor
    \State \Return $\mathit{table}$
\EndFunction
\Statex
\Function{ConstructTable}{$d$, $K$}
    \State $\mathit{counts}$ $\gets$ \Call{GetAliasEntityCounts}{$d$}
    \State $\mathit{table}$ $\gets$ \Call{ConstructTableFromCounts}{$\mathit{counts}$, $K$}
    \State \Return $\mathit{table}$
\EndFunction
\end{algorithmic}
\caption{Construct an alias table. Expects $d$ to be a list of alias-entity training pairs, $K$ specifies how many entities to keep per each alias.}
\label{alg:alias_table_construct}
\end{algorithm}

\section{Beyond Exact Matching}
\label{sec:beyond_exact_matching}
An apparent problem of systems based on alias tables is that they link only when a mention exactly matches an alias.
This does not account for the fact that mentions can be misspelled or inflexed to a form that is not present in the training data.
We propose three attempts to solve this.

\subsection{Lemmatization}\label{subsec:lemmatization}
A \emph{lemma} of a word is its one canonical form.
This allows us to group different forms (plurals, inflections, \dots).
Because of that, we can link a mention even when we do not have an alias that matches it exactly.

To demonstrate, consider the sentence, ``There have been several environmental disasters in Istanbul so far,'' which in Turkish is, ``\textit{İstanbul'da} bugüne dek birkaç çevre faciası yaşanmıştır.''\footnote{We provide our example in Turkish because it uses inflections significantly more than English. Thus, there is a greater chance that the queried form will not be in the training set.}
Let us now say that we need to link the mention denoting the biggest European city to its entity.
To achieve it with an alias table, the mention must be present during training in the same form as given above.
We might have many different forms of \textit{İstanbul} in our training data, but if none exactly matches \textit{İstanbul'da} (meaning \textit{in Istanbul}), the correct entity cannot be retrieved.
If we mapped everything to corresponding lemmas, we could answer easily assuming that \textit{İstanbul} appears in any form in the training data at least once.

Observe that this setting is similar to alias tables (\Cref{section:alias_table}).
The representation of aliases is different, but the scoring function is still looking for an exact match.

We expect that highly inflexive languages, such as Turkish, Arabic, or those within the Slavic language family, would benefit from lemmatization.
However, we also anticipate that lemmatization may not perform as well when compared to the following two alternatives.
Hence, we have chosen not to explore this baseline further.

\subsection{String Similarity}\label{subsec:ss_alias_table}
A different approach from lemmatization~(\Cref{subsec:lemmatization}) is to keep the KB the same as for alias tables~(\Cref{section:alias_table}) but change the scoring function.
The idea is to use a suitable string similarity metric to compare mentions to their aliases.

The \emph{Indel distance} (short for \textit{in}sertions-\textit{del}etions) is defined as the minimum number of insertions and deletions needed to convert one string to another. For example \textit{floor} and \textit{flower} have an Indel distance of $3$ because we can transform the former to the latter by removing \textit{o} and adding \textit{we}.
Observe that although the nouns denote something very different, they appear similar under Indel distance. 
This lack of understanding of the meaning is a problem of all approaches we encountered.
We tackle it immediately in the following section.

For our string similarity experiments, we ended up choosing the \emph{normalized Indel distance}.
For two strings $s_1$ and $s_2$, it is defined as
\[
    \frac{\mathit{Indel}(s_1, s_2)}{\mathit{len}(s_1) + \mathit{len}(s_2)}.
\]

We chose this distance due to its simplicity and availability of an efficient implementation~\citep{max_bachmann_2021_5584996}.

We present our experiments in \Cref{subsec:ss_results}.

\subsection{Pre-trained Embeddings}
\label{subsec:we_alias_table}
Pre-trained embeddings are our most ambitious attempt to improve the alias table.
We use the LEALLA-small model~\citep{mao2023lealla} to embed mentions and aliases to $128$-dimensional vectors and rank them based on the cosine similarity.
To do so, we utilize the ScaNN library~\citep{avq_2020}, which can quickly retrieve similar aliases based on a dot product and works efficiently with multiple CPUs.

\begin{figure}[t]
\centering
\includegraphics[width=.8\linewidth]{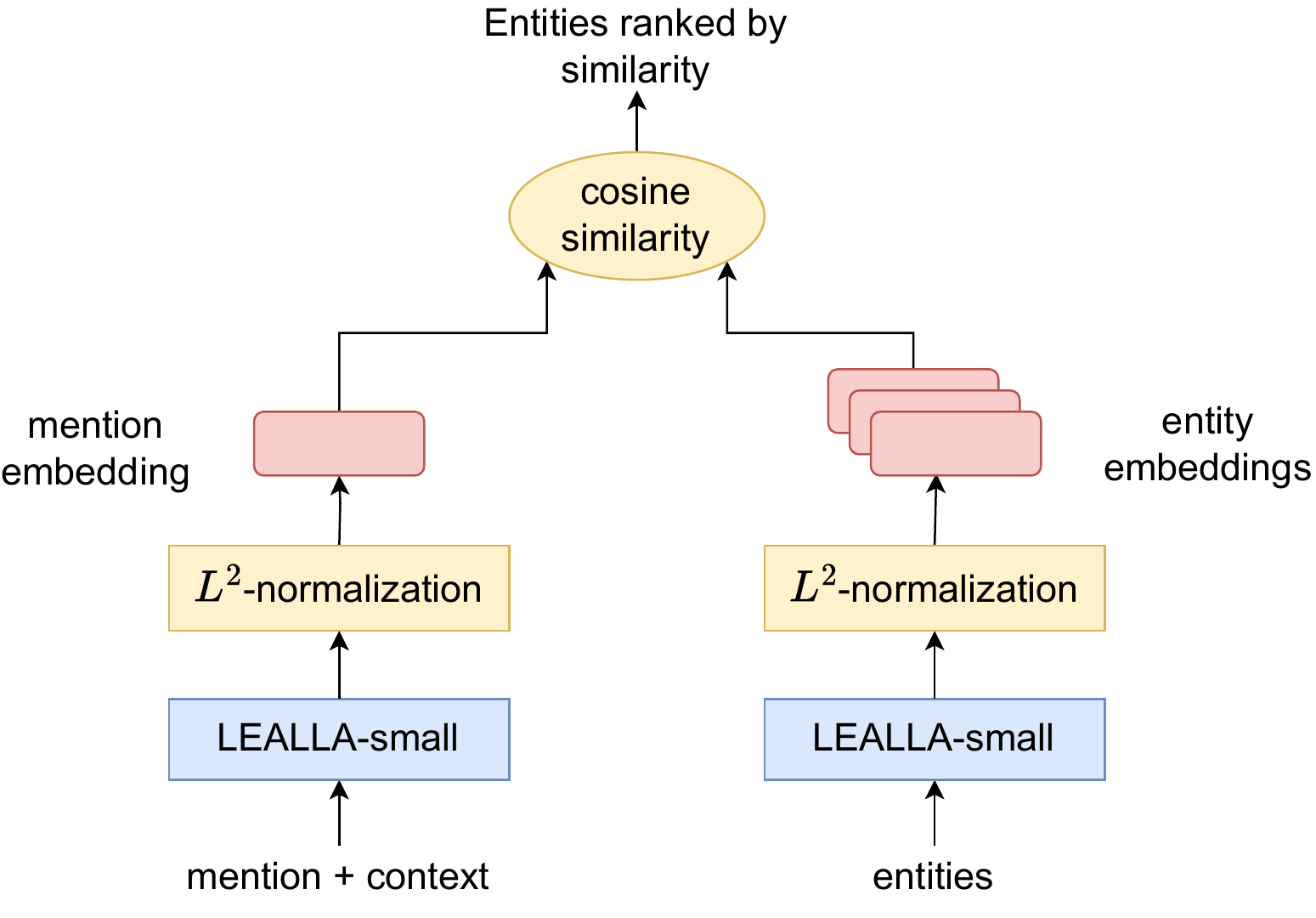}
\caption{An illustration of EL system based on aliases and an embedding model.}
\label{fig:no_traning}
\end{figure}

Because the dot product of unit vectors is the same as the cosine of the angle between them, we $L^2$-normalize the embeddings before feeding them to the ScaNN index. See~\Cref{fig:no_traning} for an illustration of the process.

One major advantage of embeddings is that they encode \textit{semantics}.
Unlike earlier approaches, none of which recognize that terms like ``United
Kingdom,'' ``UK,'' and ``Britain'' often refer to the same entity.

Our results together with more details about our approach are discussed in \Cref{subsec:we_alias_table_res}

\subsubsection{Embeddings and Model}
How well our alias embeddings encompass semantic information should depend on the quality of our models.
It is natural to expect that all LEALLAs are strong enough to provide a high-quality embedding on any aliases, thus using the smallest one for our alias experiments should be enough.
However, in \Cref{subsec:model_compar} we show that the story is far more complex.

\section{Precision of ScaNN}\label{sec:scann}
Comparing a mention to each alias in the KB takes $\mathcal{O}(d n)$ time, where $d$ represents the embedding dimension and $n$ is the number of aliases.
DaMuEL contains over $30$ million entities.
Additionally, we usually need to evaluate several tens of thousands of mentions from Mewsli.
Consequently, brute forcing the similarities is impractical.

For that reason, we use the ScaNN library (ScaNN), which uses quantization to approximate a solution of \emph{maximum inner product search (MIPS)}.
We normalize the vectors; therefore MIPS reduces to maximizing cosine similarity.
Using approximate searchers in EL is not a new idea~\citep{fitzgerald2022moleman, plekhanov2023multilingual}.

To investigate the potential decrease in recalls when using the ScaNN library, we compare brute force search with ScaNN search.
We present our results in \Cref{sec:scann_vs_bruteforce}.

\chapter{Adding Context}
\label{chap:finetuning}
So far, we have not yet utilized contexts in our approaches to EL.
Nonetheless, contexts are central to creating a robust system.
In this chapter, we explain bi-encoder models that utilize both the mention and its surrounding context.
We describe our model, compare it to approaches from~\Cref{subsec:be-encoders}, explain the training process, and discuss how different hyperparameters influence the results.

\section{Lightweight Bi-Encoder Entity Disambiguation}
Traditionally, bi-encoder models used in MEL require a long time to train.
They are the work of large research teams with abundant resources therefore for them, the training time is usually not a problem.
Below, we provide several simplifications intending to make the training easier and more accessible.

\subsection{One Model}
\begin{figure}[t]
\centering
\includegraphics[width=.8\linewidth]{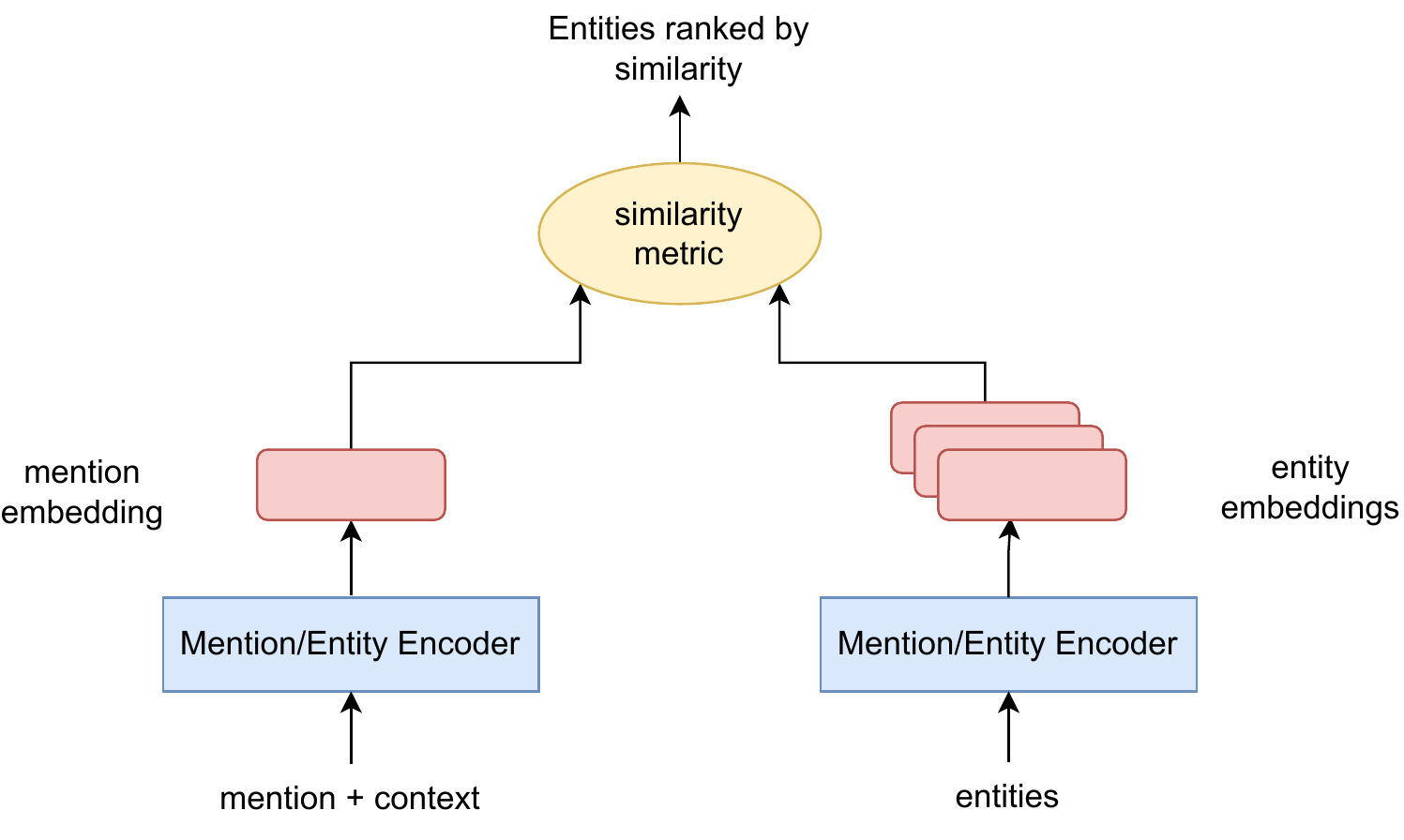}
\caption{Overview of the proposed bi-encoder. Observe that just one model is used.}
\label{fig:one_model}
\end{figure}

Notable prior works use different models for embedding entity descriptions and mention contexts.
This is motivated by the fact that these two can be structurally and semantically very different.
An entity description is a precise text pertaining to just the entity, whereas the mention context is seldom a description of anything.
Nonetheless, we note that both the mention encoder and the entity encoder need to obtain similar knowledge during training: given textual data, they need to gain the ability to distill information that explains well a given mention/entity and encode this information to an embedding.
This motivates our first idea, which is to use the same model to embed mention contexts and entity descriptions~(\Cref{fig:one_model}).

Using just one model offers several benefits.
First, we can train the system in the traditional setting, where each entity has just one KB description, but if we need to, we can extend it to the setting, where the index is populated with mention contexts, not descriptions.
This results in an improvement in inference without the need to work with a large KB of mention contexts during training.
Additionally, this extension makes adding new entities easier because it removes the need to supply a description of a new entity.
Therefore, a user of the system can add new entities by simply providing mentions of the sought-after entity.
Using just one model also results in a slightly lower memory footprint albeit not much because we are still required to keep activations to calculate the backward pass.

On the other hand, one might argue that our model loses the ability to distinguish between mentions and entities.
This could hinder it because the information it needs to extract from the input might differ for mentions and entities.
However, we disagree and hypothesize that the model can still understand whether it is presented with an entity description or a mention context, simply because contexts and descriptions do not look the same.
Hence, the model has full freedom to decide whether it should treat entities and mentions differently.
Lastly, if we were ever to conclude that the model is not strong enough to distinguish the two, and we believed that the ability to distinguish is important, we could always prepend the text with special tokens. 

Note that using one model is not an entirely novel idea in MEL. 
\citet{fitzgerald2022moleman} successfully use just one model to build a system, where the KB is populated with contexts.

\subsection{Models from a Similar Problem}
\label{subsec:similar_problem}
In the last decade, transfer learning and its subtype fine-tuning brought tremendous advancements to different deep learning fields~\citep{Silver2016AlphaGo,devlin2019bert,tan2020efficientnet}.
Many prior works on the Mewsli-9 dataset fine-tune multilingual BERT or BART~\citep{lewis-etal-2020-bart}.
However, since then, several small models trained for tasks similar to ours have been published.
From those, we chose the LEALLA family~(\Cref{sec:models}) for all our fine-tuning experiments.
Since these models are trained to align sentence pairs that are translations of each other, we expect them to converge faster than other more general models.

\citet{botha-100} use only the first $4$ layers of BERT.
We also contemplated using just the first few layers of LEALLA models, but we decided not to because the models are already quite small.

\section{How to Fine-tune a Bi-encoder}
In this section, we aim to:
\begin{itemize}
    \item frame EL as a multiclass classification;
    \item show how to structure the training examples to make the training efficient;
    \item explain how to fine-tune our model using backpropagation;
    \item give details on our fine-tuning regime and some hyparameters we use.
\end{itemize}

\subsection{EL as Multiclass Classification}\label{subsec:el_as_multiclass}
Multiclass classification is a problem where the task of the model is to assign an item to one of $k \geq 2$ possible classes.
We can formulate EL as exactly this type of problem.
The classified item is a mention-context pair $(m, c)$ and the classes correspond to entities in the KB.
Generally, in deep learning, neural networks tasked with multiclass classification apply a \emph{softmax} activation function to the outputs of the last layer.
Inputs to the softmax layer are called \emph{logits}.
\begin{figure}[t]
\centering
\includegraphics[width=.8\linewidth]{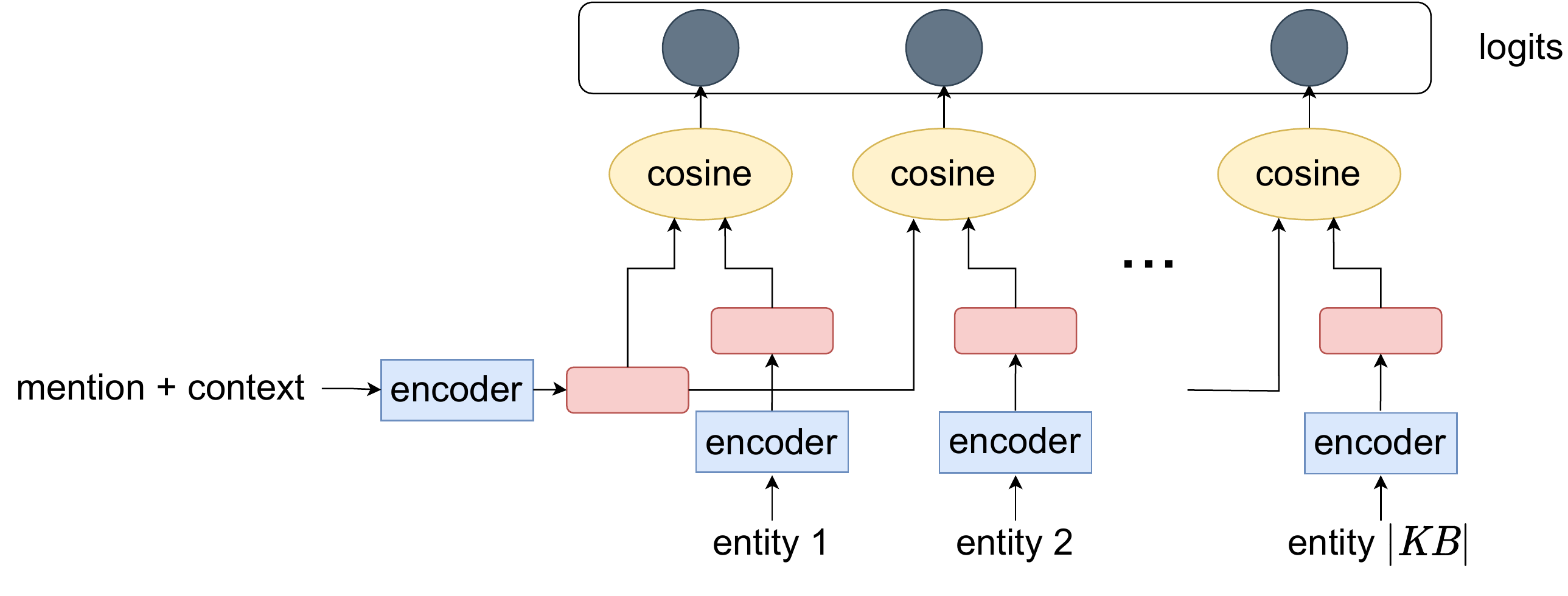}
\caption{A diagram showing how to use bi-encoder to produce logits for softmax.}
\label{fig:softmax}
\end{figure}

We can treat similarities produced by the bi-encoder as logits~(\Cref{fig:softmax}).
Yet, applying softmax to them is not straightforward.
To see why, let $\bm{s}$ be the vector of similarities, where the $i$-th element represents the similarity between queried $(m, c)$ and $i$-th item in the KB.
The softmax of the $i$-th element of $\bm{s}$ is then calculated as
\[
    \softmax(\bm{s})_i = \frac{e^{s_i}}{\sum_{j=1}^{|\KB|}e^{s_j}}.
\]
Evaluating the denominator requires evaluating all the $|\KB|$ similarities.
Because a KB often contains millions of entities, evaluating softmax for each training mention is infeasible. 
Nonetheless, the softmax's expansiveness has been long known and successfully tackled with different approximating tricks~\citep{pmlr-vR4-bengio03a, mikolov2013distributed}.

We follow the approach of~\citet{gillick2018endtoend}.
During training, we do not calculate logits for the whole KB, instead, we do so only for a small set of entities; a set that contains one positive (the KB entity corresponding to $m$) and $\mathit{neg}$ number of negative entities (different from the one corresponding to $m$).\footnote{How to choose the negatives and what is a good value for $\mathit{neg}$ is discussed later.}

\subsubsection{In-Batch Sampled Softmax}
\label{subsubsub:in_batch_sampled_softmax}
The examples we train on must provide a good approximation of the whole KB, otherwise we get a poor gradient estimate.
To decrease the noise of our estimate, we can always increase $\mathit{neg}$.
However, there is another trick that we can employ to increase the number of logits per mention.
The idea is to utilize \emph{in-batch sampled softmax}~\citep{gillick2018endtoend, henderson2017efficient} and let examples in the batch interact.
For each mention, our batches contain a positive and $\mathit{neg}$ negative entities.
The number of logits can be easily increased if we allow calculating similarities with all entities in the batch, not just those collected for the mention in question.
Therefore, for every mention, we get $b\cdot (\mathit{neg} + 1)$ logits, where $b$ is the batch size.
This provides a much better approximation of KB than the case where we have just $\mathit{neg} + 1$ logits per mention.

\begin{figure}[t]
\centering
\includegraphics[width=.9\linewidth]{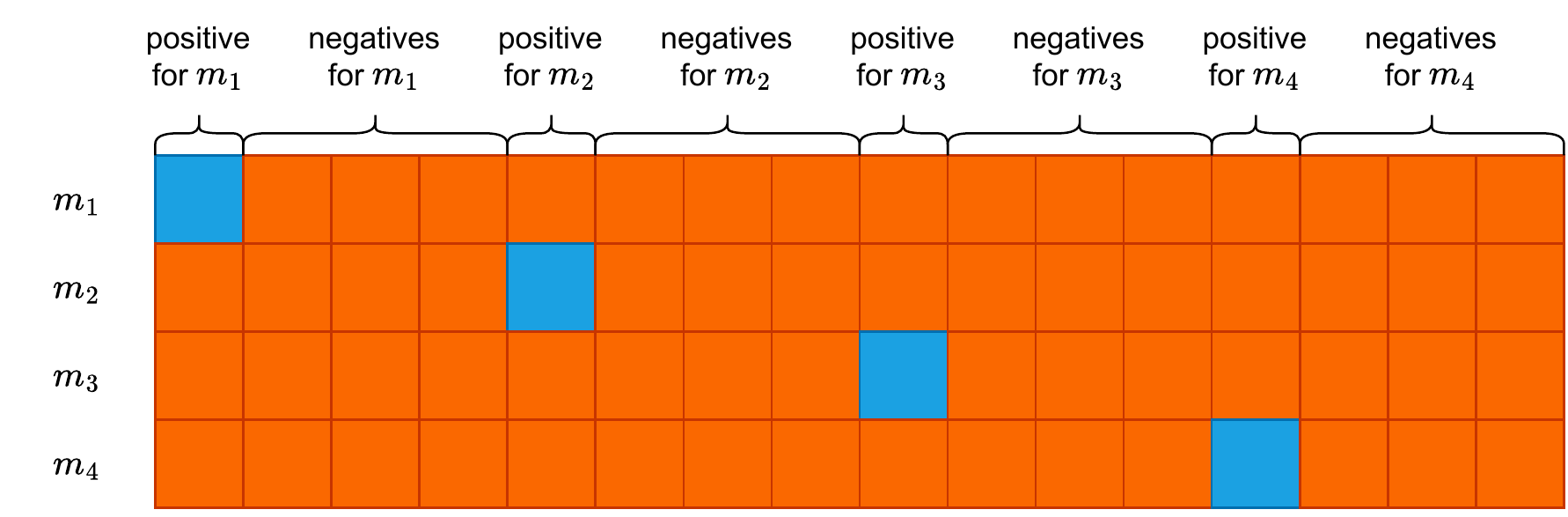}
\caption{The resulting similarity matrix after a batch is fed through a bi-encoder. Blue corresponds to similarities that the optimizer should maximize, and orange to those that should be minimized. The softmax would be evaluated per each row. We used $\mathit{neg}=3$ and $b=4$ in this illustration.}
\label{fig:batch}
\end{figure}

With the in-batch sampled softmax, the model produces a $b\times b(\mathit{neg}+1)$ matrix per batch.
The $i$-th row corresponds to similarities between the $i$-th mention and all entities in the batch. See~\Cref{fig:batch} for an example.

\subsubsection{Loss and Scaling of Similarities}
\label{subsub:loss}

To train our model, we use the cross-entropy objective with scaling
\[
    L(\bm{s}, \mathit{gold}) = - \log \softmax(\mathit{scaling}(\bm{s}))_{\gold},
\]
where $\bm{s}$ is a row of similarities from the batch matrix, and $\gold$ is the index of the sought-after entity.
Please note that we scale our similarities before we put them through the softmax.
Without an appropriate scaling, the model cannot learn (we demonstrate this in \Cref{subsec:lm_res}).
The encompassing idea is to multiply similarities by some coefficient $a$, so that they are re-scaled from the range between $-1$ and $1$ to possibly unbounded logits (or some larger range when $a$ is a constant).
Many prior works also utilize this approach.
Some predecessors use a fixed value for $a$~\citep{chidambaram-etal-2019-learning, feng2022languageagnostic}, while others change it adaptively~\citep{8953896, gillick-etal-2019-learning}.
We provide intuition on the scaling below.

Softmax is bad at capturing small-scale differences. 
When similarities are bound to a small range, even a relatively significant difference results only in a small effect on the softmax-produced probabilities.
Let us demonstrate this with an example.
Imagine that we input $512$ cosine similarities to the softmax and let us consider two cases.
\begin{enumerate}
    \item The model is impossibly confident and predicts a cosine similarity score of the correct entity to be $0.99$ and gives $-0.99$ to all others.
    \item The model is unsure but still predicts correctly. It assigns  $0.2$ similarity to the correct entity and $-0.2$ to all others.
\end{enumerate}
Now, let us compare the corresponding softmax activations.
These two cases are significantly different; hence we would expect softmax activations to also differ.
However, that does not happen.
\begin{enumerate}
    \item In the first case, the softmax assigns the probability $1.4\%$ to the correct entity and $0.19\%$ to each of the others.
    \item In the second, we get $0.29\%$ for the correct entity and $0.2\%$ for the rest.\footnote{The given probabilities for both of the cases do not sum exactly to $100$. This is because we do not want to clutter the text with too many decimals.}
\end{enumerate}

In both cases, the model appears extremely uncertain.
This is in stark contrast with the fact that in the first example, the model cannot possibly separate the entities more.

Now let us examine the case, when we multiply all similarities by a scaling multiplier.
We use $10$ in the following calculation, but as we show in \Cref{subsec:lm_res}, the exact value is not that important.
\begin{enumerate}
    \item Suddenly, we obtain a vector where the probability of the correct entity nears $100$ and all others are close to $0$.
    \item Here the values are still close to each other, thus, capturing the inability of the model to decide, yet the positive entity is much strongly separated from the herd. 
    We have $9.65\%$ for the correct entity and $0.18\%$ for the rest. 
\end{enumerate}
Thus, the scaled softmax gives a better representation of the model's confidence.

We have yet to cover why it is important that the softmax captures confidence well.
This can be easily understood when we examine the gradient update.
For this, let us step aside from scaling and imagine a general classifier $f(\bm{x} \mid \bm{{\theta}})$.
Assume that the model output is passed through a softmax activation and that the model is trained using cross-entropy.
For readability, we use $\bm{z} = f(\bm{x} \mid \bm{{\theta}})$ and $\bm{o} =\softmax(\bm{z})$.
By the chain rule, we get for some parameter $\theta_i$,
\[
    \frac{L(\bm{o}, \gold)}{\partial \theta_i}
    =
    \frac{L(\bm{o}, \gold)}{\partial \bm{z}}
    \frac{\bm{z}}{\partial \theta_i}.
\]
Using the formula for cross-entropy and softmax, we can simplify the first term and obtain
\[
    \frac{L(\bm{o}, \gold)}{\partial \theta_i}
    =
    \big(\bm{o} - \onehot_{\gold}\big)
    \frac{\bm{z}}{\partial \theta_i},
\]
where $\onehot_{\gold}$ is a vector with $1$ at the $\gold$ index and $0$s everywhere else.\footnote{You can find more detailed derivation of this formula in: \url{https://ufal.mff.cuni.cz/~straka/courses/npfl138/2324/slides/?03\#63}.}

From the above formula, we see that the confidence of the model influences the difference between $\bm{o}$ and $\onehot_{\gold}$, which then affects the gradient.
The result is that the model learns more from the examples, where it is unsure or incorrect.
Learning more from incorrectly predicted examples speeds up training, which is why the scaling factor is important.

To wrap this up, we use scaling to overcome the fact that cosine similarities are bound to a small range.
\citet{gillick-etal-2019-learning} learn the scaling factor.
We observe that the exact value of the scaling is not important (\Cref{subsec:lm_res}), as long the scaling factor is large enough to push the similarities apart.
Hence, we decided not to introduce additional complexity in the model, and we use a fixed value.

\begin{figure}[t]
\centering
\includegraphics[width=.8\linewidth]{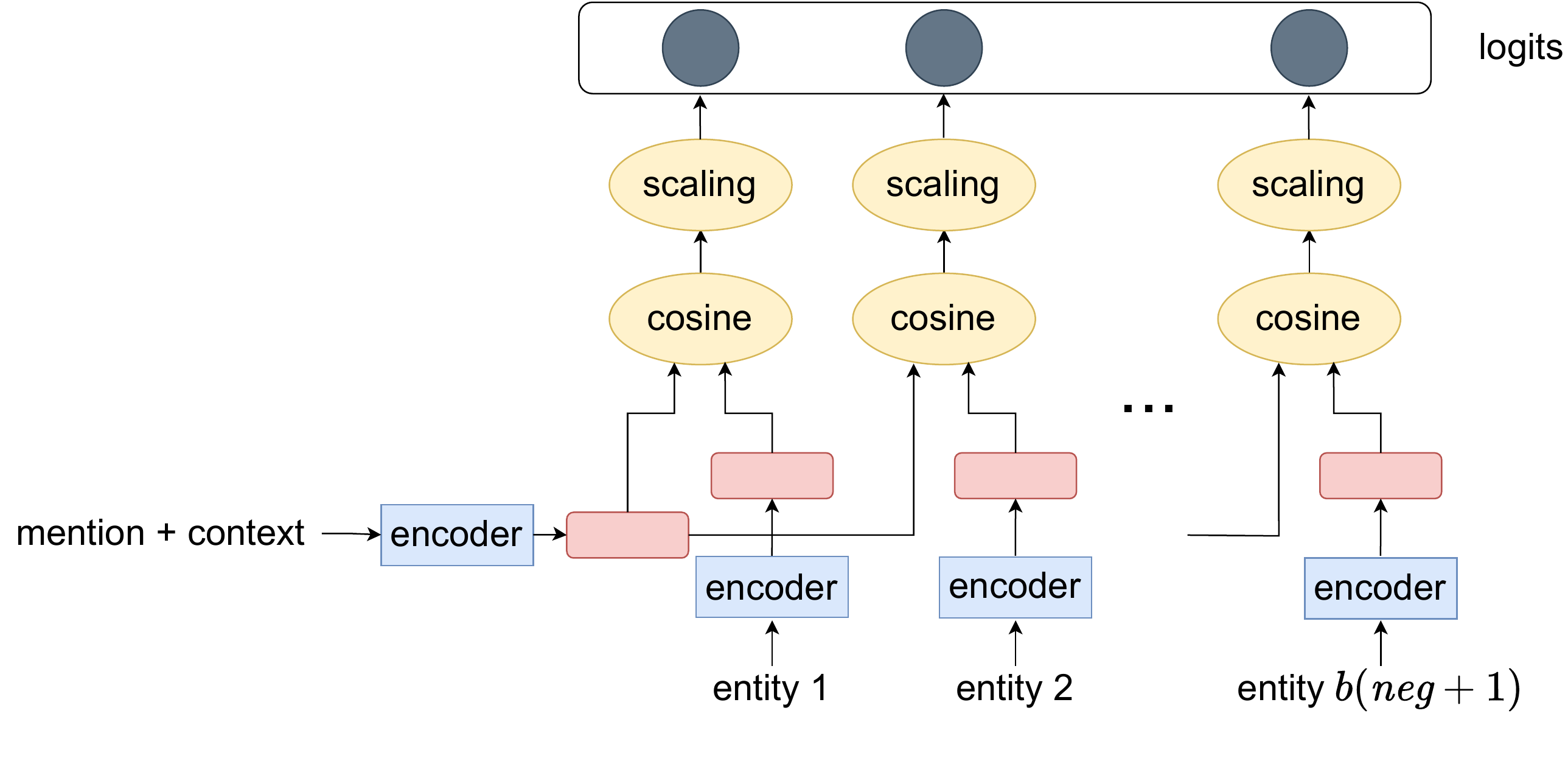}
\caption{A diagram showing how to produce logits for in-batch sampled softmax. The cosine similarities are scaled to enable learning.}
\label{fig:in_batch_sampled_softmax}
\end{figure}

You can see a diagram of the system with in-batch sampled softmax and similarity scaling in~\Cref{fig:in_batch_sampled_softmax}.

\subsubsection{Hard Negative Mining}
In ED, it is possible to attain fairly high recalls just by comparing the mention with entity labels, as showcased by our experiments with alias tables~(see~\Cref{subsec:at_damuel}).
We hypothesize that the bi-encoder learns exactly this when presented with random negatives.
As a consequence, it disregards the context completely.
For this reason, we need to use a bit more sophisticated approach when sampling negatives to a batch.
To force the model to exploit the contextual information, we utilize \emph{hard negative mining}~\citep{gillick-etal-2019-learning}.
The idea is to train the model on examples that it finds to be very similar (hard).

When building negatives for a particular mention and context, we embed them with the encoder, search for the most similar entities different from the entity of the mention, and use them as the negatives.
To retrieve the negative entities, we utilize the ScaNN index, which holds embeddings of the whole KB.
However, the creation of an index is time-consuming: it requires embedding all entities, and non-trivial time for the ScaNN library to construct the searcher.
Thus, we cannot keep the index up to date with the encoder.
Instead, the index is rebuilt only a few times during the training.

\citet{gillick-etal-2019-learning} mine only the entities that get higher cosine similarity than the gold entity.
In our work, we employ a more elaborate sampling scheme which we describe in~\Cref{subsec:token_index}.
Another notable difference is that we introduce hard negative mining from the beginning of training because our model is already pre-trained to align sentence pairs~(\Cref{subsec:similar_problem}).
Prior works use it only during the later stages of fine-tuning when the model acquires a sufficient understanding of the task.

\section{Choosing Hyperparameters}
Several hyperparameters influence the quality of our system.
In this section, we describe experiments that explore various settings. 

\subsection{Rebuilding the Index}
The longer the model trains, the better negatives it should produce during hard negative mining.
\citet{gillick-etal-2019-learning} show that the first index rebuild has a great impact on the quality of negatives and that the effect of subsequent rebuilds quickly diminishes.
We provide the results of our experiments in \Cref{subsec:index_rebuilding_res}.

\subsection{Batch Size and Queried Negatives}
Batch size is a crucial hyperparameter in any deep-learning pipeline.
Here, the significance is even greater because it directly influences the number of entities to which a mention is compared during training.

Together with the batch size, we also aim to evaluate the effect of $\mathit{neg}$, the number of sampled negatives.
This parameter has a similar role as batch size because it also influences the number of seen entities, yet in a different way. 
The value of $\mathit{neg}$ influences the number of similar entities for each of the mentions, whereas batch size makes the set of entities to which we compare more diverse.

The results of our experiments, which involved varying batch sizes and the $\mathit{neg}$ parameter, are detailed in \ref{subsec:bs_res} and \ref{subsec:neg_res}, respectively.

\subsection{Model Comparison}
The LEALLA-family has $3$ models of various sizes~(\Cref{tab:lealla_overview}).
It is natural to expect an increase in performance as the capacity of the model increases.
We investigate this in \Cref{subsec:lm_res}

\section{Cross-Lingual Transfer}
\label{sec:cross_lingual_transfer}
There are over seven thousand living languages.\footnote{According to \emph{Ethnologue: Languages of the World} there are 7,164 living languages today. \url{https://www.ethnologue.com/}}
This linguistic diversity poses a significant challenge for any NLP task one might want to solve for multiple languages. Even gathering training data for a small subset of the languages can be daunting.
One viable solution is \emph{cross-lingual transfer}, a method based on the assumption that a multilingual model represents various languages within a unified semantic space.\footnote{Not to be confused with cross-lingual entity linking from \Cref{subsec:multilinguality}}.
As noted by \citet{macková2020reading}, the network is incentivized to represent similar languages with the same part simply because it saves some capacity.
Generally, this should allow us to take a multilingual model trained on languages $A$ and $B$, fine-tune it on a task \textit{only} in $A$, and then use it to solve the task in $B$.
Studies have validated the effectiveness of this approach on EL~\citep{schumacher2021crosslingual} and many various other tasks~\citep{pires-etal-2019-multilingual,lewis-etal-2020-mlqa}.

In our case, we do not lack training data for the languages we plan to evaluate on.
Nonetheless, for us, the important benefit of cross-lingual transfer is that it saves time during training.
When training in just one language, we deal with a noticeably smaller index and fewer mentions.
Thus, we are interested in exploring if training on one language and evaluating on the whole Mewsli-9 benchmark produces competitive results.
We evaluate this idea in \Cref{sec:cros_lingual_transfer_res}.
We would like to emphasize that in our approach, we still use at least some entity representations from the unseen languages during evaluation, however, we do not train on them.
\chapter{Infrastructure}
\label{chap:infra}
In this chapter, we describe the most important components of our infrastructure.

\section{Tokenization}\label{sec:tokenization}
Before we can feed a text to LEALLA, we tokenize it with the appropriate tokenizer.
We use a fixed number of tokens for both descriptions and contexts.
When the corresponding text is too short, we pad the tokenizer's output.
We adopt slightly different schemes for tokenizing descriptions and mentions with context.

\subsection{Descriptions}
All our descriptions of entities consist of two parts: a label and a description.
The description part is either the start of the corresponding Wikipedia page or its Wikidata description.
We prioritize pages over descriptions because the former contains much more information.
However, the page is not always available.
The label is marked with special tokens --- \verb|[M]| --- and concatenated to the description part.
The resulting string input to the tokenizer then looks like this:

\label{subsec:infra_descriptions_example}
\begin{quote}
``\verb|[M]|Chancellor (Poland)\verb|[M]| Chancellor of Poland (Polish: Kanclerz - Polish pronunciation: [...], from Latin: cancellarius) was one of the highest officials in the historic Crown of the Kingdom of Poland. This office functioned from the''
\end{quote}

\subsection{Mentions with Context}
All the datasets we work with represent mentions in the form of a text and a mention span.
We add \verb|[M]| tokens around the mention and include as many preceding and following characters (the context) as needed to obtain the chosen number of tokens.
Often in our data, one text contains multiple different mentions (for example a
Wikipedia page with many links), thus, we can save some time by tokenizing the
text only once and adding the \verb|[M]| tokens as needed.

We aim to put the mention in the middle of the extracted text so that the result contains information preceding and succeeding the mention.
Nevertheless, putting it in the middle is not always possible, because there might not be enough tokens on one of the sides.
Therefore, whenever the context window overlaps with the text's start or end, we try to enlarge it to the other side.

A mention corresponding to the example entity from~\Cref{subsec:infra_descriptions_example}
might look like
\begin{quote}
    ``(and sejmik) who presided over the proceedings and was elected from the body of deputies evolved in the 17th century.) Next, the \verb|[M]|kanclerz\verb|[M]| (chancellor) declared the king's intentions to both chambers, who would then debate separately till the ending ceremonies. After 1543 the''
\end{quote}

\section{Fine-tuning}
\label{sec:finetuning_infrastructure}
\begin{figure}[t]
\centering
\includegraphics[width=.8\linewidth]{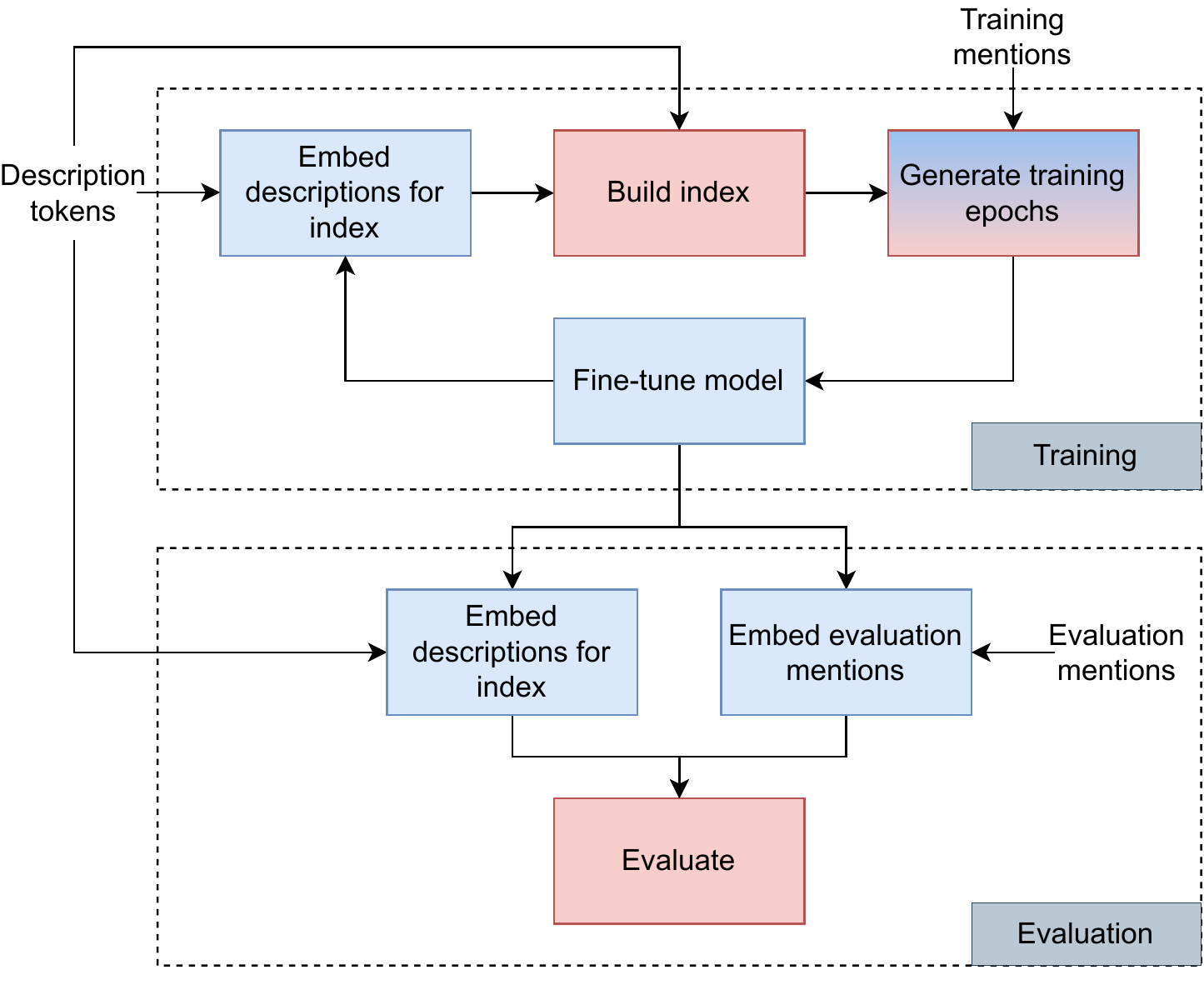}
  \caption{The diagram contains the training part and the evaluation part. All steps requiring a ScaNN searcher are marked red, and those working with the embedding model are marked blue.}
\label{fig:finetuning_loop_diagram}
\end{figure}

The fine-tuning consists of multiple steps~(\Cref{fig:finetuning_loop_diagram}).
The majority of those steps concern itself with the construction of training examples.
In \Cref{subsec:el_as_multiclass} we describe how we employ hard-negative-mining to get negative examples for the batch.
This requires having an index that can efficiently retrieve those negative examples.

A typical fine-tuning procedure thus consists of the following steps:
\begin{itemize}
    \item index construction;
    \item batch construction;
    \item fine-tuning.
\end{itemize}

The quality of the negative examples from the index directly depends on the quality of the model.
Therefore, the above steps are repeated multiple times. 
Later, in \Cref{subsec:index_rebuilding_res} we explore what number of rounds is optimal.

Below, we describe the above-mentioned steps in more detail.

\subsection{Index}
\label{subsec:token_index}
Our goal is to have a data structure that takes embeddings of some mention with context and returns tokens of the most similar entity descriptions.
To achieve our goal, we use an instance of a ScaNN index, which we wrap to a class that facilitates the $\text{embedding}\rightarrow\text{tokens}$ querying.

A typical query (\Cref{alg:query}) consists of an embedding, its QID, $\mathit{neg}$, and $k$; the expected answer contains tokens of \textit{negative} entity descriptions.

For training, we want the entities that are shown to the model to be as diverse as possible.
For this reason, we do not retrieve the $\mathit{neg}$ most similar entities, but we instead do this in a two-step fashion.
First, we retrieve the $k$ most similar items from the ScaNN index.
From them, we sample uniformly at random without replacement the required $\mathit{neg}$ entities.

Note that in (\Cref{alg:query}) we remove occurrences of neighbors that correspond to the same entity (with calls to \textproc{RemoveDuplicates}).
Currently, we prevent multiple occurrences of the same entity in the index by limiting the number of per entity representations to one during training, thus \textproc{RemoveDuplicates} does not do anything.
However, in future work, we plan to explore systems that use \textit{more than one description per entity} (these are often sourced from different languages).
This slightly complicates the query execution.
If we executed the query naively, the returned $\mathit{neg}$ items might contain multiple instances that correspond to the same entity.
To demonstrate this with an example, imagine we have an index with descriptions gathered from all Wikipedias, and we are constructing negative examples for \textit{mythical Paris}.
The above-described query can easily return $20$ almost identical descriptions of \textit{Paris-city}, albeit in $20$ different languages.
This might teach the model the difference between the city and the mythological figure, but it does not give any information on the similarity to the socialite Paris Hilton.
For efficient training, we hypothesize that seeing a greater number of different entities is better.
Therefore, before we sample negatives, we remove successive occurrences of the same entity from the result of the ScaNN query.

It is possible that after we filter out the successive entities, we are left with less than $\mathit{neg}$ entries.
When this happens, we multiply $k$ by two.
We repeat this process until we get the desired number of negatives.

\begin{algorithm}[t]
\begin{algorithmic}
\Function{QueryNegatives}{$e_{\mathit{mc}}$, $\mathit{qid}$, $\mathit{neg}$, $k$} 
\State $\mathit{neighbors}$ $\gets$ \Call{RetrieveNegativesFromScaNN}{$e_{\mathit{mc}}$, $k$, $\mathit{qid}$}
\State $\mathit{neighbors}$ $\gets$ \Call{RemoveDuplicates}{$\mathit{neighbors}$}
\While{\Call{Len}{$\mathit{neighbors}$} < $\mathit{neg}$}
    \State $k$ $\gets$ $2 \cdot k$
    \State $\mathit{neighbors}$ $\gets$ \Call{RetrieveNegativesFromScaNN}{$e_{\mathit{mc}}$, $k$, $\mathit{qid}$} 
    \State $\mathit{neighbors}$ $\gets$ \Call{RemoveDuplicates}{$\mathit{neighbors}$}
\EndWhile
\State \Return \Call{SampleTokens}{$\mathit{neighbors}$, $\mathit{neg}$}
\EndFunction
\Statex
\Function{RetrieveNegativesFromScaNN}{$e_{\mathit{mc}}$, $k$, $\mathit{qid}$}
\State $\mathit{neighbors}$ $\gets$ \Call{RetrieveNeighborsFromScaNN}{$e_{\mathit{mc}}$, $k$}
\State \Return \Call{FilterNegatives}{$\mathit{neighbors}$, $\mathit{qid}$}
\EndFunction
\end{algorithmic}
\caption{Query negatives for batch construction. The input should consist of an embedding of mention with context, the corresponding QID, the requested number of negatives, and the retrieval parameter $k$. Returns the tokens of $\mathit{neg}$ entities.}
\label{alg:query}
\end{algorithm}

\subsection{Generating and Fine-tuning}
For simplicity, we split batch construction and fine-tuning into two distinct steps, although we
acknowledge that they could run together, which would offer different runtime-memory tradeoffs.

When generating batches, training mentions are embedded with the current best version of the model.
For each mention, a required number of positives and negatives is retrieved from the index.
A batch consists of the following triplets:
\begin{itemize}
    \item tokens of mentions,
    \item list of tokens of positive and negative entities,
    \item matrix of targets; each row of the matrix is a probability
      distribution (corresponds to the matrix from \Cref{fig:batch}). 
\end{itemize}
Finally, batches are concatenated into epochs, which are compressed and saved to disk.

During fine-tuning, we process the epochs produced in the previous step.
We train our model using PyTorch~\citep{Ansel_PyTorch_2_Faster_2024}.
In all our experiments we optimize with Adam~\citep{Kingma2014AdamAM} and the following parameters: $\mathit{lr} = 1\cdot 10^{-5}$, $\beta_1=0.9$, and $\beta_2=0.999$.

\subsection{Evaluation}\label{subsec:eval}
During evaluation, the model is used to embed descriptions and evaluation mentions.
To find the most similar entity, we try to utilize ScaNN and brute force.
To make the brute-force efficient, we approach calculating similarities as a matrix multiplication with queries in one matrix and entities in the other.
This allows us to calculate similarities in parallel on a GPU.
We also tune the parameters of ScaNN to make it more precise at the cost of some slowdown.

\section{Implementation}
All our code can be obtained from GitHub.\footnote{\url{https://github.com/Yokto13/multilingual-entity-linking}}
Additionally, for archival purposes, we provide the same code as a supplement to this thesis.

The root of the repository contains \path{README.md}, the required Python packages in \path{requirements.txt}, and directories \path{src} and \path{tests}.
In \path{src/finetunings}, we provide our fine-tuning pipeline.
It contains several subdirectories that roughly correspond to the setup we gave in this chapter.
For running the fine-tuning, please consult \path{src/run_finetuning.sh} and \path{src/run_finetuning_round.sh}.

Apart from the fine-tuning pipeline, we also provide an implementation of the mention-only approaches from \Cref{chap:baselines}.
The majority of those can be found in \path{src/baselines} or \path{src/scripts}.
The former also contains scripts for calculating the upper bounds and intersections, whose results are in \Cref{section:damuel_mewsli_intersection} and \Cref{appendix:intersection}.
The remaining directories in \path{src} contain data structures and some utility functions that are required for our experiments.

All our code runs under Python version 3.10.12;\footnote{\url{https://docs.python.org/3.10/index.html}} some scripts also require PySpark version 3.5.0.\footnote{\url{https://spark.apache.org/docs/3.5.0/}}

\chapter{Results and discussion}\label{chap:res}

\section{ScaNN vs Brute-force}\label{sec:scann_vs_bruteforce}
To see how much precision we lose due to ScaNN, we propose a simple experiment.
We embed entity labels from DaMuEL and all Mewsli-9 mentions with the plain LEALLA-base model.
Then, for each mention, we retrieve the closest entity label either using ScaNN or brute force.
We run this experiment for each language separately (mentions from one Mewsli-9 language are compared only to entities from the corresponding DaMuEL language).
This ensures that we have at most one representation per entity, and we do not have to resolve conflicts that naturally arise every time an entity has a label in more than one DaMuEL language.
We use this baseline experiment once more in \Cref{subsec:cross_baseline}, where we describe it in more detail.

The larger the language, the harder it is to brute force it without encountering problems with the GPU's memory (see \Cref{subsec:eval} for details on brute force).
Therefore, to make our computation easier, we skip brute-force evaluation for English and German.
We present the results in \Cref{tab:scann_vs_bruteforce}.
Losses from using approximate searching are minimal.
Consequently, we see ScaNN as sufficient for our experiments and we use it for all subsequent evaluations.

\begin{table}[t]
  \centering
  \begin{tabular}{lcccc}
    \hline \multirow{2}{*}{Language} & \multicolumn{2}{c}{ ScaNN} & \multicolumn{2}{c}{Bruteforce} \\
    \cline{2-5}
            &  R@1 &  R@10 &  R@1 &  R@10 \\
    \hline
        ar & 66.3 & 84.1 & 66.3 & 84.3\\
        es & 58.9 & 77.8 & 58.9 & 78.2\\
        fa & 66.5 & 82.4 & 66.7 & 82.4\\
        ja & 66.5 & 78.2 & 66.6 & 78.4\\
        sr & 58.2 & 88.4 & 58.2 & 88.6\\
        ta & 63.7 & 76.8 & 63.8 & 77.4\\
        tr & 71.4 & 85.4 & 71.5 & 85.7 \\
    \hline
  \end{tabular}
  \caption{Baseline With Labels \Cref{subsec:cross_baseline} evaluated using ScaNN and bruteforce.}
  \label{tab:scann_vs_bruteforce}
\end{table}

\section{LEALLAs vs LaBSE}\label{sec:leallavslabse}
In~\Cref{sec:models} we described several models for producing text embeddings: LaBSE and LEALLA family.
We note that LaBSE might be faster even though it has several times more parameters than LEALLAs.
Here, we explore this in more detail.

One could erroneously assume that LEALLAs, having fewer parameters, should be faster.
However, a careful examination of the LEALLA paper reveals that the models have more layers than LaBSE.
LEALLA's authors observe that on their task, a deep architecture with relatively small layers yields better results than a shallow architecture with larger layers.
Consequently, their models are smaller than LaBSE in terms of parameters, but contain two times more layers, making the forward pass slightly more expensive (see \Cref{tab:lealla_speed} for results).
On the other hand, the smaller number of parameters of LEALLA models allows for running them with larger batch sizes, which makes our speed comparison inconclusive.

In this and all other experiments, we use models hosted on Hugging Face.\footnote{\url{https://huggingface.co/}}

\begin{table}[t]
  \centering
  \begin{tabular}{lcc}
    \hline
    Model & bs 4096, time [ms] & bs 16384, time [ms] \\
    \hline
    LEALLA-small & 1961 & 1752 \\
    LEALLA-base & 1972 &  1770 \\
    LEALLA-large & 1980 & 1788 \\
    LaBSE & 1861 & OoM \\
    \hline
  \end{tabular}
  \caption{LEALLAs and LaBSE: time needed to embed all $56716$ mentions from the Spanish part of Mewsli-9 for two different batch sizes. Each mention consists of $64$ tokens. We ran each experiment $3$ times, averaged the result, and rounded it. Running LaBSE with the large batch size was unsuccessful and raised an out-of-memory error. Only the time needed to pass the data through the model is shown. In reality, much more time is spent on extracting mentions of the right size, their tokenization, and batch construction. All experiments used NVIDIA GeForce RTX 3090 (24 GB) and PyTorch.}
  \label{tab:lealla_speed}
\end{table}

\section{DaMuEL as Entity Linking Dataset}
In \Cref{section:damuel_mewsli_intersection} we noticed that DaMuEL lacks some entities that are present in Mewsli-9.
Since DaMuEL has not yet been used in a publicly available EL system, we do not know how much the missing entities might affect performance.
Hence, in this section, we compare it using the alias table (\Cref{section:alias_table}) to the foundational work by \citet{botha-100}.
Our results show that DaMuEL is comparable in regard to R@1 but lacks in R@10.
To our knowledge, we are the first to undertake a comprehensive analysis of DaMuEL in this manner.

Additionally, apart from evaluating the alias table in \Cref{subsec:at_damuel} we also experiment with the string similarity baseline from \Cref{subsec:ss_alias_table} in \Cref{subsec:ss_results}.

\subsection{Alias Table}\label{subsec:at_damuel}
We use an alias table described in~\Cref{section:alias_table} to evaluate DaMuEL with Mewsli-9.
We aim to find out how well DaMuEL compares to the unpublished training data from the Mewsli paper.

We extract aliases from each of the $53$ specific DaMuEL parts by processing all the links~(see~\Cref{subsec:damuel} for refresh of DaMuEL's structure) and adhering to the following points:
\begin{itemize}
    \item To allow fair comparison, we use only mentions with links originating in Wikipedia, not those added by DaMuEL's authors.
    \item Less than $10$ mentions were defined as empty strings. We examine these by hand and find out that the corresponding Wikipedia text contains some unusual special characters. We expect that these characters broke the tokenization process of DaMuEL so that the span does not correspond to a mention. Because these mentions are a tiny fraction of the dataset, we skip them. 
\end{itemize}

We create one table from the extracted aliases, which we use to link all mentions from all Mewsli-9 languages. 
We calculate both R@1 and R@10.
Our results are presented in~\Cref{tab:alias_table_dam_vs_mew}.

In R@1 DaMuEL seems to be on par with the dataset from~\citet{botha-100}.
In R@10 it is outperformed.
We believe that the underperformance on R@10 can be explained by the missing entities~(see \Cref{subsec:damuel} for details).
A significant part of them are disambiguations, which should influence R@10 much more than R@1.

Note the surprisingly weak performance of DaMuEL in Persian.
Our initial assumption was that there are problems with charsets.
Therefore, we normalized Persian with Parsivar~\citep{mohtaj-etal-2018-parsivar} but to no avail.
Later, our poor results with LEALLA-based embeddings alias tables~(\Cref{subsec:we_alias_table}) allowed us to reject the normalization.
Word embeddings should be invariant to charsets, yet the Persian with embeddings is still significantly worse~(results in~\Cref{subsec:we_alias_table_res}).
There is also a possibility, which we do not explore, that the Persian in DaMuEL is preprocessed in some way that even the LEALLA's tokenizer is baffled.

Interestingly, the Persian is the smallest Mewsli-9 language.
It contains a mere $535$ mentions, which is less than $0.2\%$ of all Mewsli-9, therefore evaluating on it might be noisy.
We hope to explore this in future work on DaMuEL.

Additionally, we provide evaluations of $R@K$ for $K > 10$ and results with uncased aliases in \Cref{appendix:alias_tables}.

\begin{table}[t]
  \centering
  \begin{tabular}{lcccc}
    \hline
     \multirow{2}{*}{Language} & \multicolumn{2}{c}{ DaMuEL} & \multicolumn{2}{c}{ EL100} \\
    \cline{2-5}
                     &   R@1       &  R@10      &  R@1       &  R@10      \\ 
    \hline
    ar         &        87.4         &      89.7        &    \textbf{89.0}            &       \textbf{93.0}             \\ 
    de       &          \textbf{86.3}       &         90.3      &      86.0              &      \textbf{91.0}              \\ 
    en      &            77.6         &      84.6         &      \textbf{79.0}              &      \textbf{86.0}              \\ 
    es      &            \textbf{82.4}     &          89.0     &      82.0              &      \textbf{90.0}              \\ 
    fa      &           72.9      &         76.8     &      \textbf{87.0}              &      \textbf{92.0}              \\ 
    ja      &            \textbf{82.7}        &      89.2       &      82.0              &      \textbf{90.0}              \\ 
    sr      &            \textbf{87.2}        &      90.3         &      87.0              &      \textbf{92.0}              \\ 
    ta     &             \textbf{81.2}       &       82.8      &      79.0              &      \textbf{85.0}              \\ 
    tr     &             \textbf{81.7}  &        \textbf{88.3}     &      80.0              &      88.0              \\ 
    \hline
    micro-avg        &  82.7 &         88.2          &        \textbf{82.8}          &  \textbf{89.4}                 \\ 
    macro-avg        &  82.2     &     86.8               &   \textbf{83.4}              &      \textbf{89.7}              \\ 
    \hline
  \end{tabular}
  \caption{DaMuEL compared to EL100 on Mewsli-9 using the alias table.}
  \label{tab:alias_table_dam_vs_mew}
\end{table}

\subsection{String Similarity}
\label{subsec:ss_results}
For the string similarity baseline~(\Cref{subsec:ss_alias_table}) we use normalized Indel distance from RapidFuzz~\citep{max_bachmann_2021_5584996}, which is a fast string matching library for C++ and Python that builts on top of FuzzyWuzzy.\footnote{The original FuzzyWuzzy library can be found here: \url{https://github.com/seatgeek/fuzzywuzzy}.}

We construct the table per language to save some computational resources during the evaluation.
This means that the aliases for a table for German are constructed only from mentions from the German part of DaMuEL, etc.
We call such alias tables \emph{one language alias tables} to emphasize the difference to other alias table experiments where we use all DaMuEL languages.
Observe that when gathering aliases, we process each language separately but otherwise use the same process as in \Cref{subsec:at_damuel}.

Querying a mention by calculating similarities to all aliases is expensive.
However, from \Cref{tab:alias_table_dam_vs_mew} we note that the majority Mewsli-9 queries can be answered by exact match (this corresponds to the Indel distance of 0).
Therefore, we evaluate string similarity only for mentions we cannot solve by matching exactly.

To validate our string similarity results we cannot use the values from \Cref{tab:alias_table_dam_vs_mew} because that alias table is constructed using all DaMuEL languages.
Hence, we also compute an alias table per each of the nine Mewsli languages.
We use the term \emph{one language alias table} (OLAT) to distinguish it from the alias table from \Cref{subsec:at_damuel}.

We show the results of the string similarity in ~\Cref{tab:similarity}.
We observe relatively consistent improvements across all languages except for Tamil, where the improvement is much larger than for the other languages: $8.9$ (R@1) and $11.4$ (R@10).
Interestingly, on Tamil, our simple baseline beats by two points the bi-encoder model from the Mewsli-9 paper~\citep{botha-100}.
\begin{table}[t]
  \centering
  \begin{tabular}{lcccccc}
    \hline
     \multirow{2}{*}{Language} & \multicolumn{2}{c}{ String similarity} & \multicolumn{2}{c}{ OLAT} & \multicolumn{2}{c}{ Difference}\\ 
    \cline{2-7}
                     &   R@1       &  R@10      &  R@1 &  R@10  &  R@1 &  R@10    \\ 
    \hline
    ar & 89.7 & 93.5 & 87.5 & 89.6 & 2.2 & \03.9\\
    de & 88.6 & 92.8 & 86.0& 89.1& 2.6 & \03.7 \\
    en & 80.2 & 88.6 & 77.1& 83.8& 3.1 & \04.8\\
    es & 83.8 & 90.4  &81.8 & 87.4& 2.0 & \03.0 \\
    fa & 75.7 & 80.0 &71.6& 75.3& 4.1& \04.7 \\
    ja & 85.4 & 93.5 &82.7& 89.1& 2.7 & \04.4 \\
    sr & 89.3 & 92.7 & 86.4& 88.8& 2.9 & \03.9\\
    ta & 90.1 & 94.2 & 81.2& 82.8& 8.9 & 11.4 \\
    tr & 85.2 & 92.7 &81.0& 87.0&4.2 & \05.7 \\
    \hline
  \end{tabular}
  \caption{Results for the alias table where string similarity metric is used instead of an exact match. We evaluate on languages from Mewsli-9. \emph{OLAT} denotes the one language alias table; it serves as a baseline to see how much of an improvement string similarity brings. 
  }
  \label{tab:similarity}
\end{table}

\section{Pre-trained Embeddings}
\label{subsec:we_alias_table_res}
Here, we evaluate the alias table improved with pre-trained embeddings as described in~\Cref{subsec:we_alias_table}.

We again evaluate per language.
Consequently, we call our approach one language pre-trained embeddings alias table (OLPEAT).
We gather the aliases the same way as in the string similarity experiment (\Cref{subsec:ss_results}).
After that, we embed them with LEALLA-small to $128$-dimensional vectors.
We evaluate on all Mewsli-9 languages and use ScaNN~\citep{avq_2020} to find the most similar alias embeddings.

We present the results in~\Cref{tab:pretrained_embeddings}.
Generally, languages with a smaller number of aliases perform better.
This seems reasonable because in those languages it is less likely that an entity label appears in all possible morphological variations.
In \Cref{sec:beyond_exact_matching} we hypothesize that highly inflected languages should get a greater performance boost. 
However, our results do not seem to neither prove or disprove it; both Turkish and Serbian perform well, yet an improvement in Arabic is small.

\begin{table}[t]
  \centering
  \begin{tabular}{lcccccc}
    \hline
     \multirow{2}{*}{Language} & \multicolumn{2}{c}{ OLPEAT} & \multicolumn{2}{c}{ OLAT} & \multicolumn{2}{c}{ Difference}\\ 
    \cline{2-7}
                     &   R@1       &  R@10      &  R@1 &  R@10  &  R@1 &  R@10    \\ 
    \hline
    ar & 90.1 & 93.3 & 87.5 & 89.6 & \02.6& \03.7\\
    de & 88.8 & 92.9 & 86.0& 89.1& \02.8& \03.8\\
    en & 80.1 & 88.2 & 77.1& 83.8& \03.0& \04.4\\
    es & 83.9 & 90.3 &81.8 & 87.4& \02.1& \02.9\\
    fa & 83.7 & 89.7 &71.6& 75.3& 12.1&14.4\\
    ja & 85.2 & 92.6 &82.7& 89.1& \02.5& \03.5\\
    sr & 91.1 & 95.0 & 86.4& 88.8& \04.7& \06.2\\
    ta & 91.2 & 95.5 & 81.2& 82.8& 10.0&12.7\\
    tr & 86.0 & 93.3 &81.0& 87.0& \05.0& \06.3\\
    \hline
  \end{tabular}
  \caption{Results for one language word embeddings alias table (OLPEAT) on Mewsli-9. 
  We compare it with the standard alias table built separately for each language (OLAT).
  OLPEAT items were embedded with LEALLA-small. ScaNN was used for the comparison.
  }
  \label{tab:pretrained_embeddings}
\end{table}

\subsection{String similarity vs Pre-trained Embeddings}
In \Cref{sec:beyond_exact_matching} we discuss different approaches for improving alias tables.
We experimented with the two we deem the most promising above.
In this part, we briefly compare them.
See \Cref{tab:similarity_and_olpeat} for the results.

\begin{table}[t]
  \centering
  \begin{tabular}{lcccc}
    \hline
     \multirow{2}{*}{Language} & \multicolumn{2}{c}{ String similarity} & \multicolumn{2}{c}{ OLPEAT}\\
    \cline{2-5}
                     &   R@1       &  R@10      &  R@1 &  R@10\\ 
    \hline
    ar & 89.7 & \textbf{93.5} &  \textbf{90.1} & 93.3\\
    de & 88.6 & 92.8 & \textbf{88.8}& \textbf{92.9} \\
    en & \textbf{80.2} &  88.6& 80.1& \textbf{88.2} \\
    es &  83.8 &\textbf{90.4} & \textbf{83.9} & 90.3\\
    fa & 75.7 & 80.0 & \textbf{83.7}& \textbf{89.7}\\
    ja & \textbf{85.4} & \textbf{93.5} & 85.2&  92.6\\
    sr & 89.3 & 92.7 & \textbf{91.1}&  \textbf{95.0}\\
    ta & 90.1 & 94.2 & \textbf{91.2}& \textbf{95.5}\\
    tr & 85.2 & 92.7 & \textbf{86.0}& \textbf{93.3}\\
    \hline
  \end{tabular}
  \caption{String similarity (\Cref{subsec:ss_results}) compared to one-language-pre-trained-embeddings alias table (\Cref{subsec:we_alias_table_res}).}
  \label{tab:similarity_and_olpeat}
\end{table}

The embedding-based system demonstrates superior performance. 
Notably, string similarities \textit{only outperform embeddings in Japanese}.
This can be potentially attributed to the complexity of the Japanese script. 
Calculating Indel distance remains consistent across different scripts, so it should not matter as much whether we do it for Japanese, Spanish, or Serbian. 
On the other hand, for a deep learning model, the effectiveness of text embeddings for Japanese significantly relies on how much and how well it is represented in the training dataset.
It is possible that LEALLA-small was not exposed to Japanese to the extent that it completely understands the complexity of the script.

The approach utilizing embeddings outperforms others not only in recalls but also in efficiency. 
Excluding the time needed to embed DaMuEL, none of the evaluations with OLPEAT exceed three hours. 
In contrast, evaluations based on string similarity for larger languages, such as German or English, require more than 12 hours.\footnote{With the exception of English both OLPEAT and string similarity evaluations were conducted on a single machine equipped with $30$ Intel(R) Xeon(R) Silver 4110 CPU @ 2.10GHz. The English language, which required more memory, ran on a machine equipped with $20$ Intel(R) Xeon(R) Gold 6230 CPU @ 2.10GHz.} 
However, it is important to note that such a comparison is somewhat biased against string similarity, as it overlooks the nonnegligible time required to generate the embeddings and the huge time to train the LEALLA-small model.

\subsection{Comparing Models in Spanish}
\label{subsec:model_compar}
To find out whether the choice of model influences results on the OLPEAT~(\Cref{subsec:we_alias_table_res}) approach we evaluate for Spanish also with other LEALLA models.
The results (\Cref{tab:different_models_pretrained_embeddings}) surprise us.
Naturally, one would expect that a model with a larger number of parameters would be slightly stronger than a one with less.
Yet, our results point in the opposite direction.
The smallest model and the base model score both high in R@1 and R@10, and the largest model is outperformed by a large margin by both of its peers.
We did not find out what the reason is for the weak performance of LEALLA-large.
However, we encounter this phenomenon again during fine-tuning in \Cref{subsec:size_finetune}.

\begin{table}[t]
  \centering
  \begin{tabular}{lcc}
    \hline
    Model &  R@1 &  R@10 \\
    \hline
    LEALLA-small & 83.9& 90.3 \\
    LEALLA-base & 83.9 & 90.3\\
    LEALLA-large & 64.7 & 70.2  \\
    \hline
  \end{tabular}
  \caption{Comparison of different LEALLA models on OLPEAT in Spanish. Contrary to our expectation the the \textit{large} model trails significantly behind the other two.}
  \label{tab:different_models_pretrained_embeddings}
\end{table}

\section{Hyperparameter search}
\label{sec:hyperparameter_search}
In this section, we report the results of our search for the best hyperparameters.

We use the fine-tuning setup from~\Cref{sec:finetuning_infrastructure}.
To decrease the size of the index and make the whole fine-tuning pipeline faster, we train only on the Spanish part of DaMuEL and evaluate on the Spanish mentions from Mewsli-9.
We chose Spanish due to its substantial yet still smaller data size than English, and because we have a basic proficiency in the language, which facilitates easier debugging.

Additionally, in this particular section, we decided to exclude from our KB all the entities that do not have a Wikipedia page.
There are multiple reasons to do so.
Perhaps the most important one is that it makes any operation with the index substantially faster.
Also, all links inside a specific DaMuEL part always lead to a Wikipedia page.
Thus, we cannot see an entity that does not have a Spanish page as a positive example during training (interestingly, one could still learn about non-page entities because they can be mined as hard negatives).
Moreover, we would like to remark that omitting Wikidata items without a page is not uncommon.
For example \citet{botha-100,decao2021multilingual} remove any items without a page from their entire training.
However, it is important to note that the aforementioned works used KBs constructed from more languages than Spanish.
Consequently, more entities had a page in at least one language, thus the omission might not cost them much.
Lastly, we observed that from the Mewsli-entities that have a corresponding entity in Spanish DaMuEL only around $2\%$
cannot be disambiguated to a Spanish Wikipedia page.
Therefore, we concluded that tuning hyperparameters only with entities with a page is a sufficient approximation of the classical task, in which we would build a KB from
all possible Spanish entities.
We would like to add that we later train on all the entities in
\Cref{sec:cros_lingual_transfer_res}.

Some parameters are the same for all fine-tuning experiments; we present these in~\Cref{tab:finetuning_params}.
Note that we use a different number of steps for the first and subsequent rounds of fine-tuning.
In our experience, the learning converges quickly in the first round because the batches are simple to solve.
We chose these numbers of steps because we did not qualitatively see any improvement in recalls when training longer.

\begin{table}[t]
  \centering
  \begin{tabular}{lr}
    \hline
    context size & 64\\
    steps per 1\textsuperscript{st} round & $2 \cdot 10$\rlap{$^4$}\\
    steps per 2\textsuperscript{nd} and subsequent rounds & 10\rlap{$^5$}\\
    logit multiplayer & $50$\\
    retrieval parameter $k$ & $100$\\
    \hline
  \end{tabular}
  \caption{Hyperparameters shared across all our experiments.}
  \label{tab:finetuning_params}
\end{table}

\subsection{Index Rebuilding}
\label{subsec:index_rebuilding_res}
Here, we evaluate what is the optimal number of index rebuilds in our settings.
We use a batch size of $32$ and query $7$ negative descriptions per mention.

\begin{figure}[t]
\centering
\includegraphics[width=.8\linewidth]{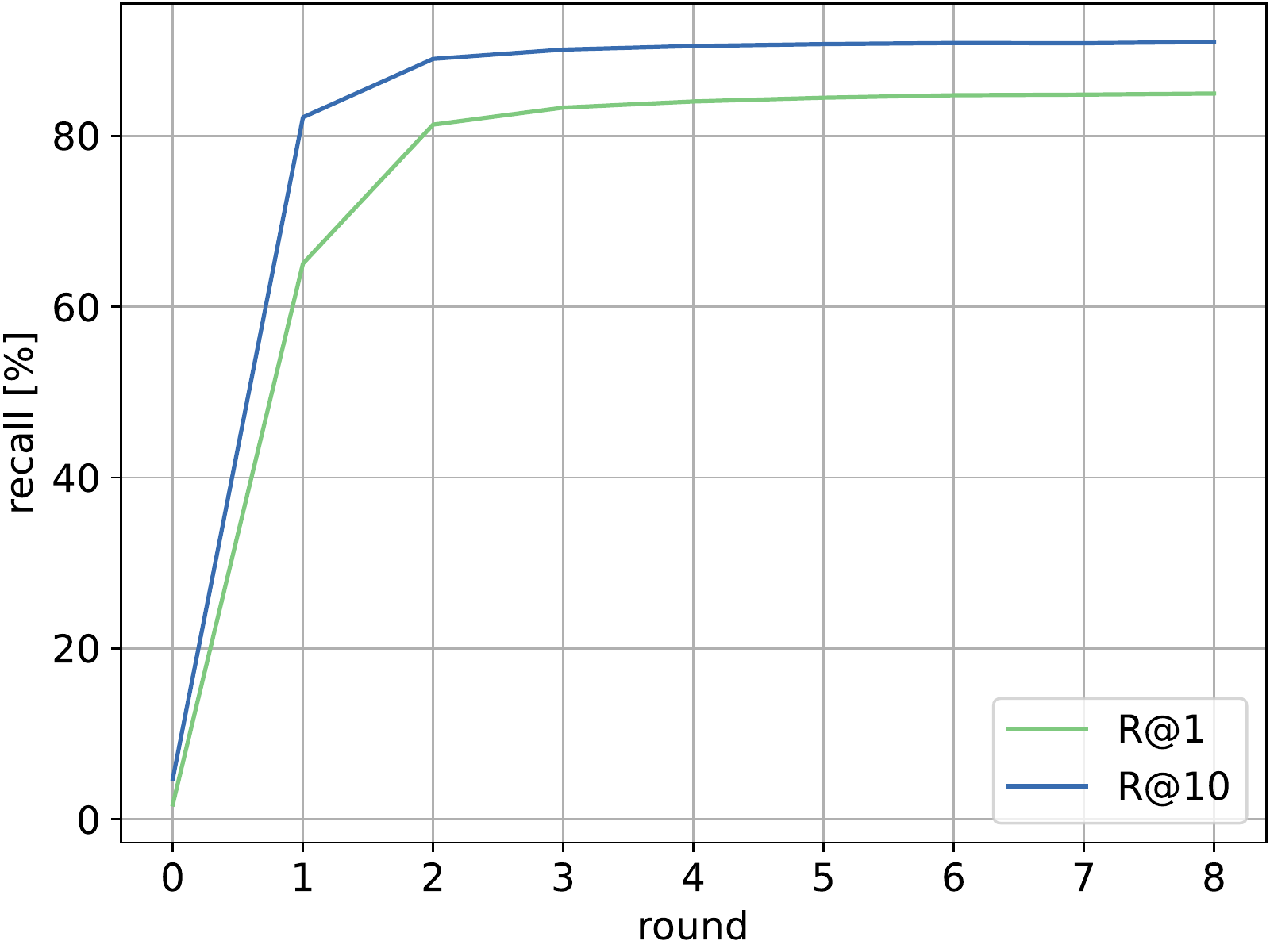}
  \caption{Experiment evaluating the required number of \textit{finetuning rounds}. We start indexing rounds at $1$ and evaluate when each fine-tuning round finishes. The $0$th evaluation corresponds to evaluation without any training. The greatest improvement is after the first two rounds, then the curve flattens.}
\label{fig:index_experiment}
\end{figure}

Our results~(\Cref{fig:index_experiment}) show that in our regime four to five rounds are enough for the recall to stabilize (corresponds to $520\cdot10^3$ learning updates).
The biggest improvement is after the first round, where the model most likely learns to disambiguate based on the mention and entity label, and the recalls jump from values around $5\%$ to around $70 \%$. 
This is comparable to recalls we can get by embedding entity labels and using them as KB entries (\Cref{tab:cross_per_lang}).
The second round gives another performance boost.
In the subsequent rounds, the improvement in recall quickly diminishes.

The performance of our model without any fine-tuning --- which corresponds to round $0$ in \Cref{fig:index_experiment} --- is worse than we initially anticipated.
When context is involved, aligning entities and mentions is hard (see \Cref{appendix:context_embeddings} for detailed evaluation).
This is not a significant problem because the model quickly improves.
However, it does mean that the hard negatives are of low quality until the index is first rebuilt with the fine-tuned model.
If we had better hard negatives, we could likely learn even faster.
Therefore, we propose a simple improvement, which we, however, do not evaluate in this thesis: during the initial round, it is better to mine the hard negatives based solely on the mention, disregarding the context.
\Cref{subsec:we_alias_table_res,subsec:cross_baseline} speak for this approach because they demonstrate that the model can align mentions and entity names well even without training.

\begin{figure}[t]
\centering
\includegraphics[width=.8\linewidth]{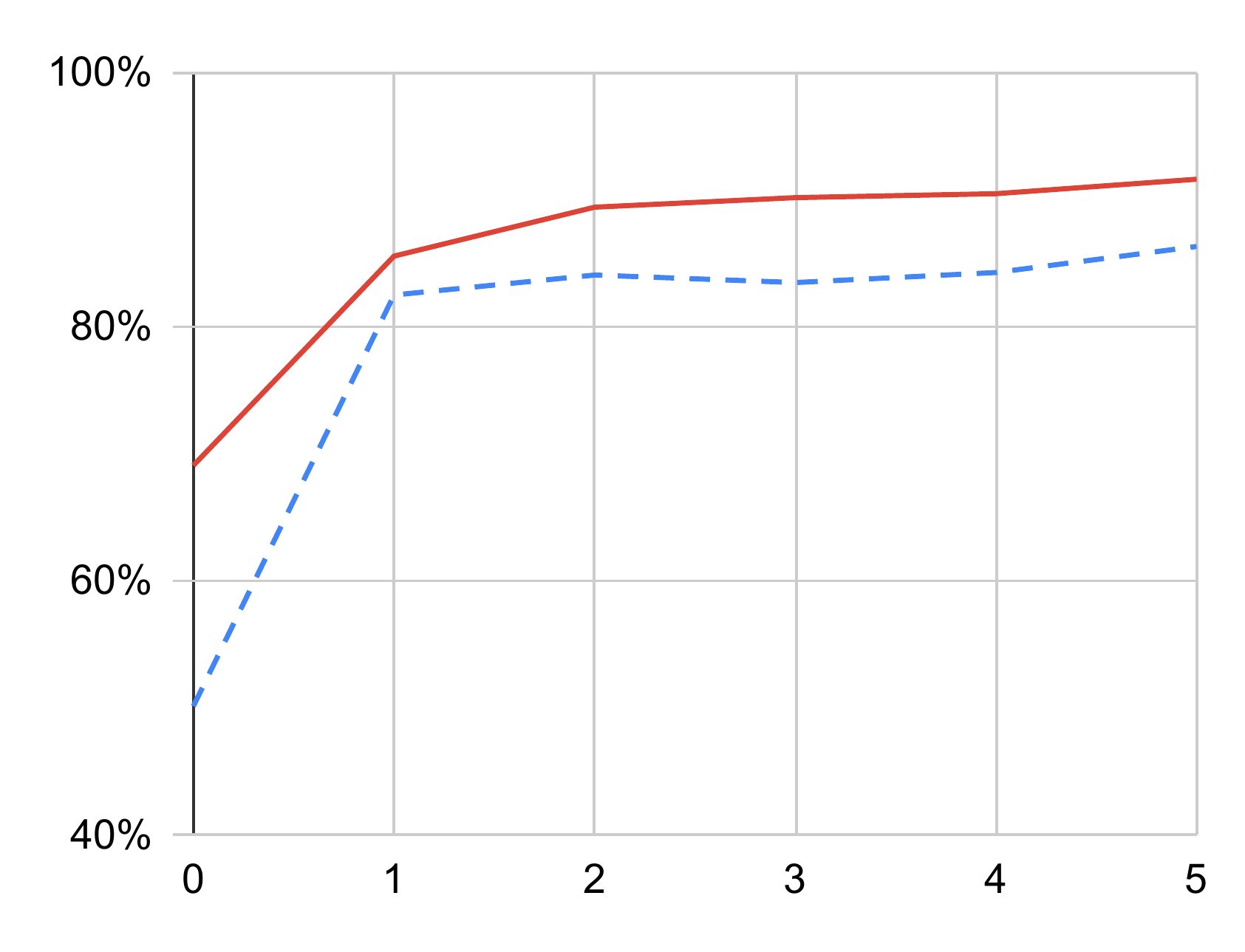}
\caption{Experiment evaluating the required number of \textit{hard negative mining rounds} from \citet{gillick-etal-2019-learning}. The solid line corresponds to R@1 at the Wikinews~\citep{gillick-etal-2019-learning} dataset and the dashed at TACKBP-2010~\citep{Ji2010OverviewOT}. A direct comparison with our result is impossible because the paper uses a different dataset and the condition that decides when to rebuild the index is different. Also, their model is trained with random negatives before the start of hard negative mining. Nonetheless, the shape is similar to ours.}
\label{fig:gillick_index_experiment}
\end{figure}

In \Cref{fig:gillick_index_experiment} we provide a figure from~\citet{gillick-etal-2019-learning}, which is similar to ours. 
The plot has a familiar shape, but a direct comparison is impossible due to the fact that our approach differs in data, the underlying model, and slightly in the training procedure and round-stopping condition. 

\subsection{Batch Size}\label{subsec:bs_res}
\begin{figure}[t]
    \centering
    \includegraphics[width=.49\textwidth]{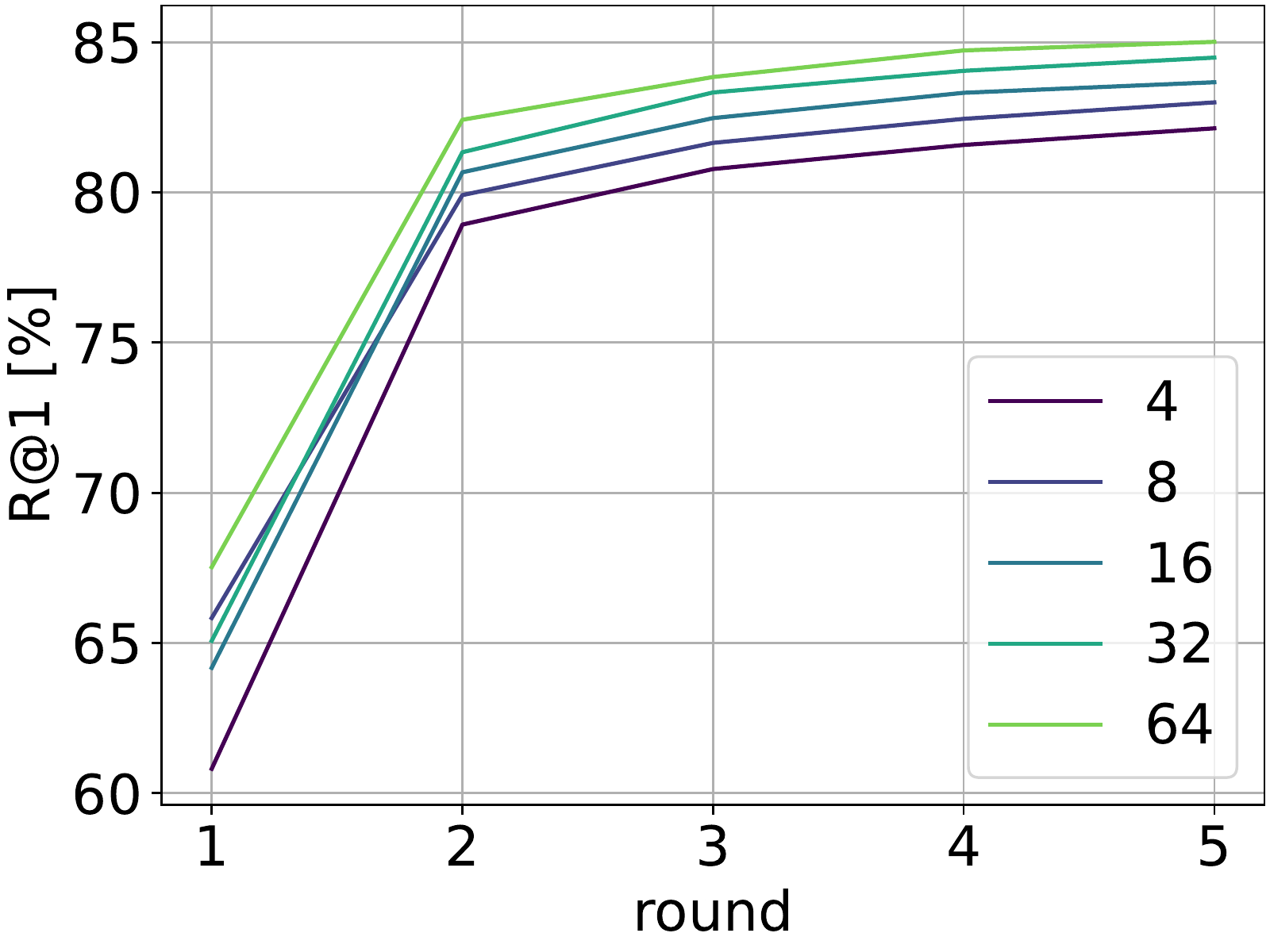}\kern.02\textwidth
    \includegraphics[width=.49\textwidth]{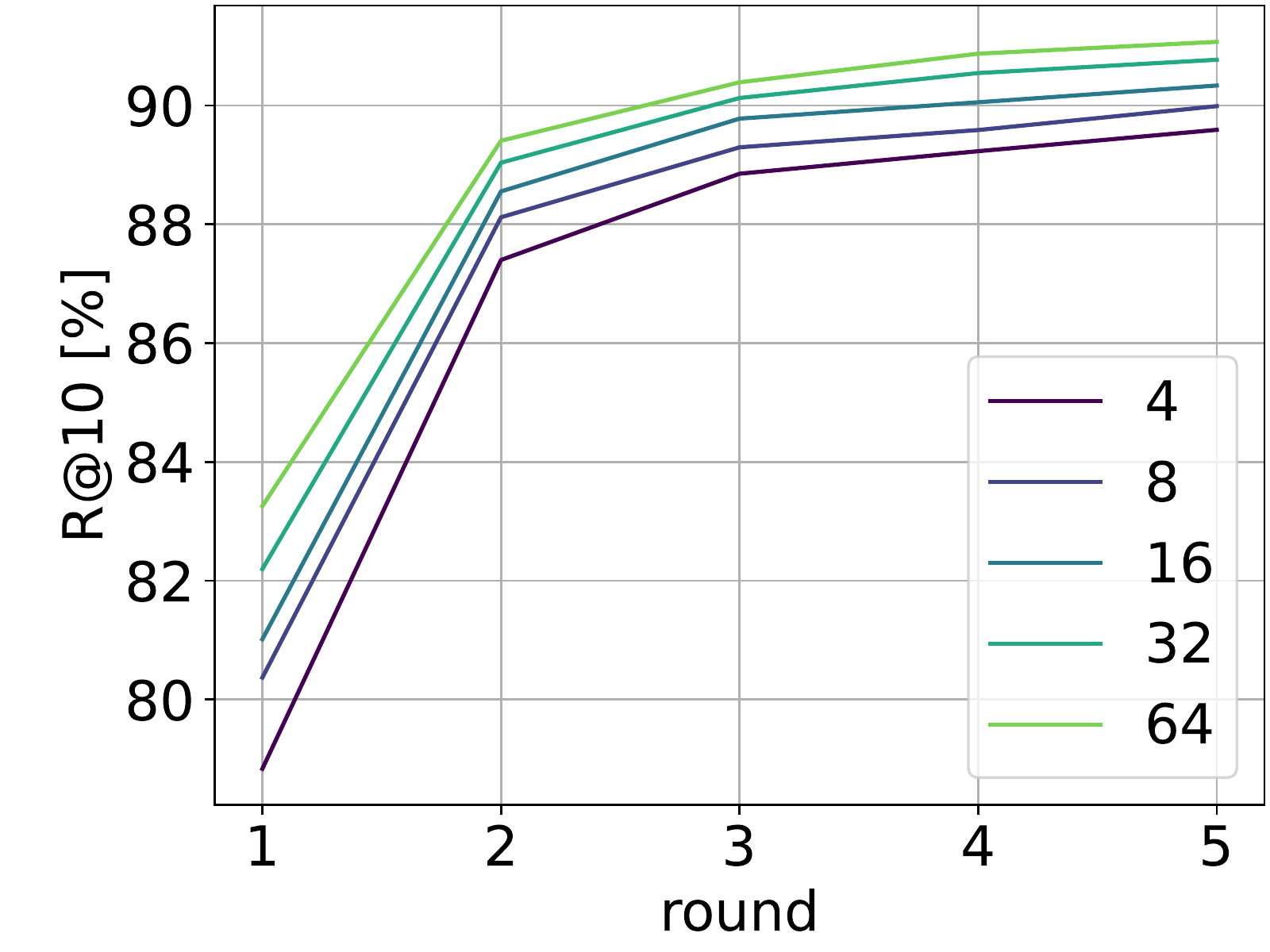}
    \caption{Results of experiments varying the batch size for R@1 and R@10. There is a consistent improvement in both R@1 and R@10 when increasing the batch size.}
    \label{fig:bs}
\end{figure}

We experiment with five different batch sizes (4, 8, 16, 32, 64) and always use $\mathit{neg}=7$. Our results are in \Cref{fig:bs}.
Unsurprisingly, we see that recall increases when the batch size grows.
With our batch sizes, we do not reach a point where an increase in batch size does not produce an increase in recalls.
Yet we can see that gap between recalls decrease, and from it we conclude that there exists a batch size where the model saturates.
However, this batch size can be quite large.
When training bi-encoders it is not uncommon to use batches with several thousand mentions (for example \citet{botha-100, fitzgerald2022moleman} use $8192$ while training a model similar to ours).

Another thing to note in this experiment is that the fine-tuning is quite stable.
Even with the batch size of $4$, the model trains well, and after five rounds it is only approximately $2$ points worse than the best model.

We also experimented with a batch size of $1$, which proved to be insufficient, and the model did not train.

\subsection{Number of Hard Negatives}\label{subsec:neg_res}
\begin{figure}[t]
        \includegraphics[width=.49\textwidth]{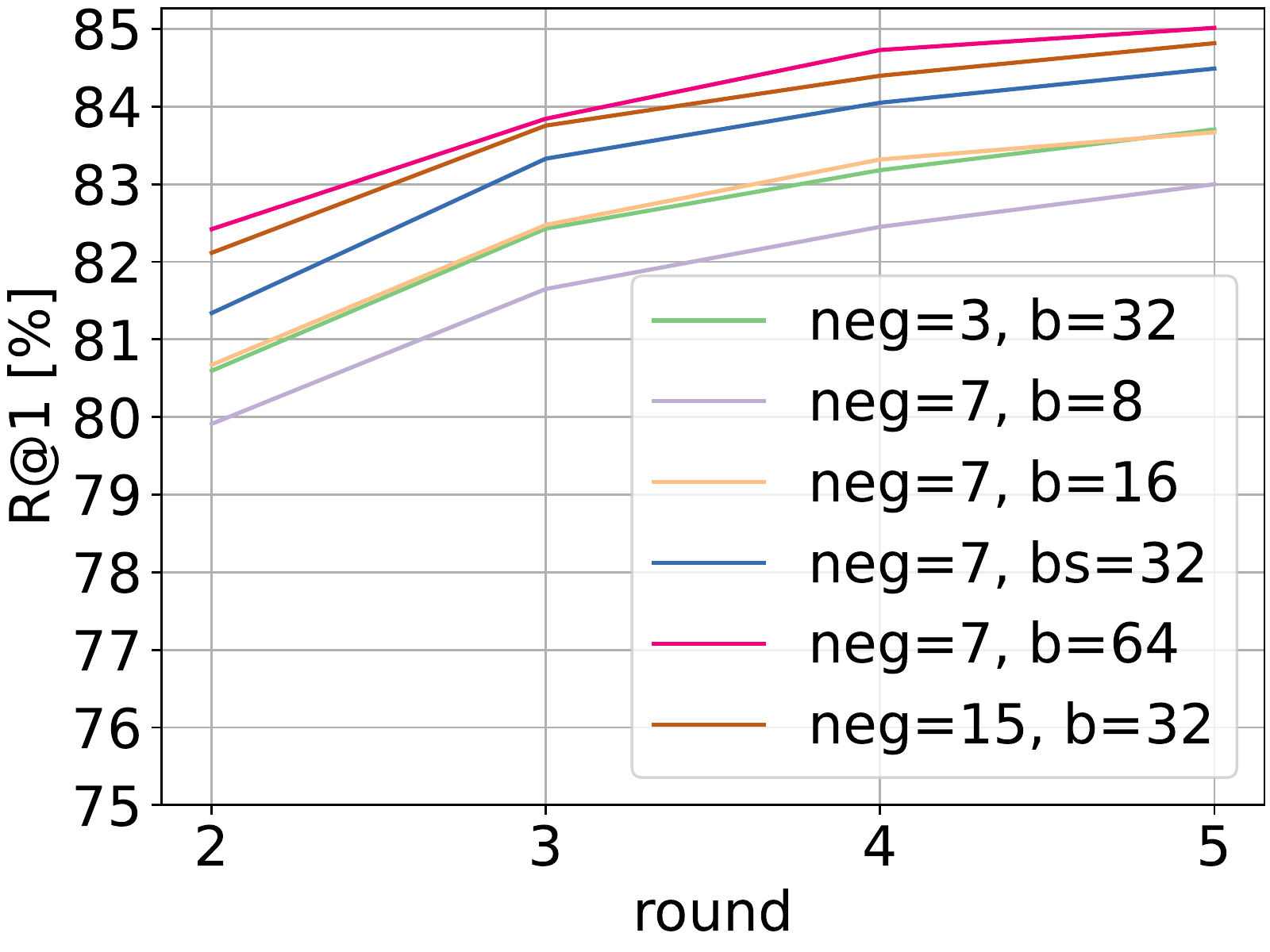}\kern0.02\textwidth
        \includegraphics[width=.49\textwidth]{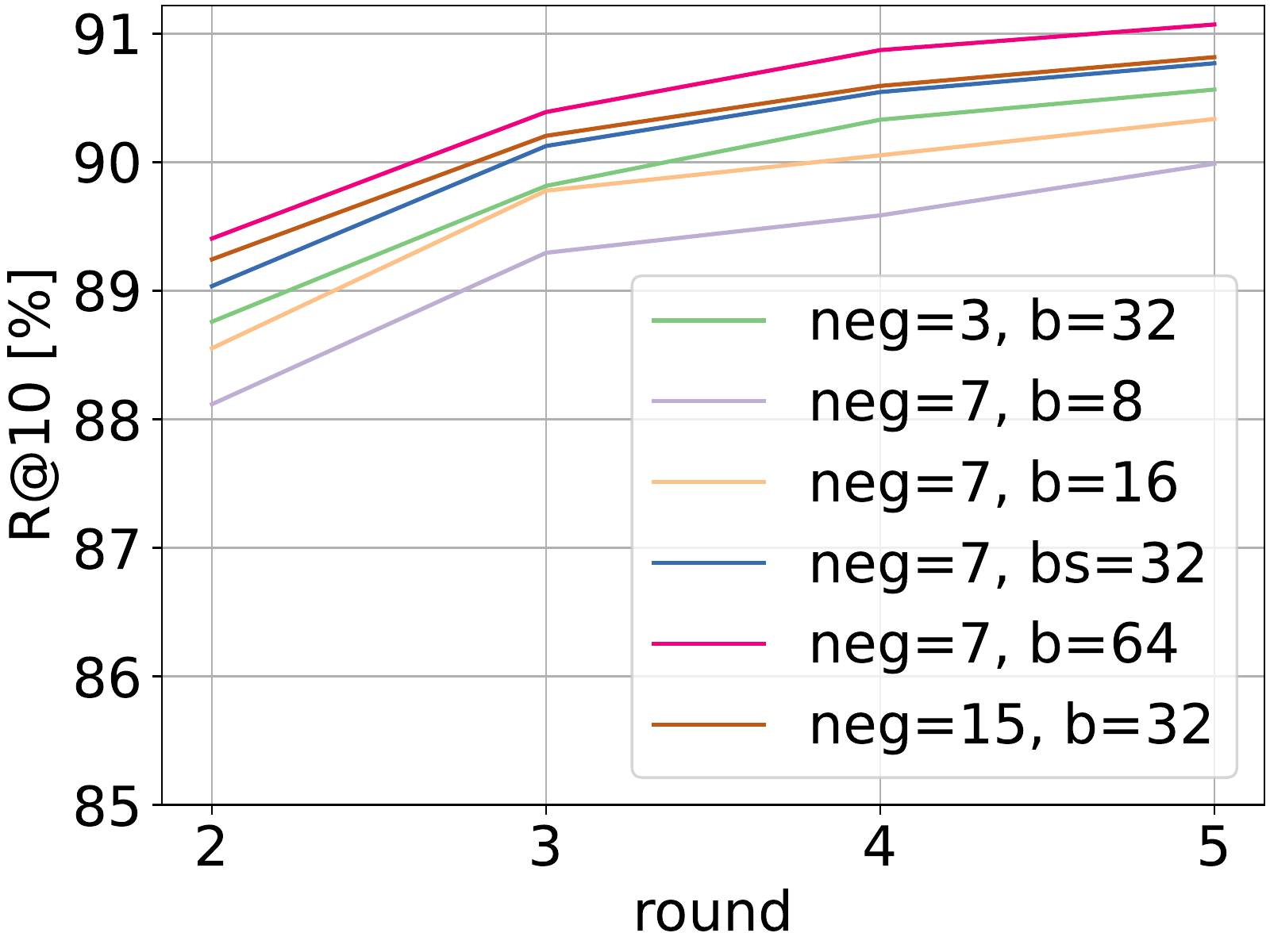}
    \caption{Results of experiments varying the batch size and $\mathit{neg}$ for R@1 and R@10. We compare multiple different values of the batch size and the number of negative descriptions retrieved from the index. Note that the $x$-axis starts from the second round; this is to gain enough precision on the $y$-axis so that all the small differences can be observed.}
    \label{fig:negs}
\end{figure}

Apart from the default $\mathit{neg}$ value of $7$, we also experiment with $\mathit{neg}=3$ and $\mathit{neg}=15$.
We present our results in \Cref{fig:negs}.

Although it clearly holds that both higher batch size (which we denote as $b$ in this subsection) and  $\mathit{neg}$ improve recalls, the striking observation is that for R@1 it nearly does not matter which one of those we increase.
All that is important is the number of entities to which a mention is compared.
Recall that after feeding a batch through the system, we are left with $b \times b (1 + \mathit{neg})$ output matrix (\Cref{fig:in_batch_sampled_softmax}).
It seems that the number of columns of this matrix is a decent indicator of the success we can expect at R@1.

To be more specific, let us observe the results of our experiment.
In the evaluation for R@1, the curve corresponding to the $b = 16$ and $\mathit{neg} = 7$ is nearly the same as for $b=32$ and $\mathit{neg} = 3$.
Both output matrices contain $128$ similarities per each mention. 
Similarly, the curves for $\mathit{neg}=15, b=32$ and $\mathit{neg}=7, b=64$ are very close, albeit here the higher batch size starts to dominate.
Interestingly, for R@10 the picture is different and the larger batch size wins.

Notice that the number of columns of the matrix is linear in both $b$ and $\mathit{neg}$, but the number of items is quadratic in $b$ and linear in $\mathit{neg}$.
This leads to an important result.
When working with constrained resources where memory is scarce, it might be better to increase $\mathit{neg}$ over the batch size.

\subsection{Logit Multiplier}\label{subsec:lm_res}
\begin{figure}[t]
    \centering
    \includegraphics[width=0.49\textwidth]{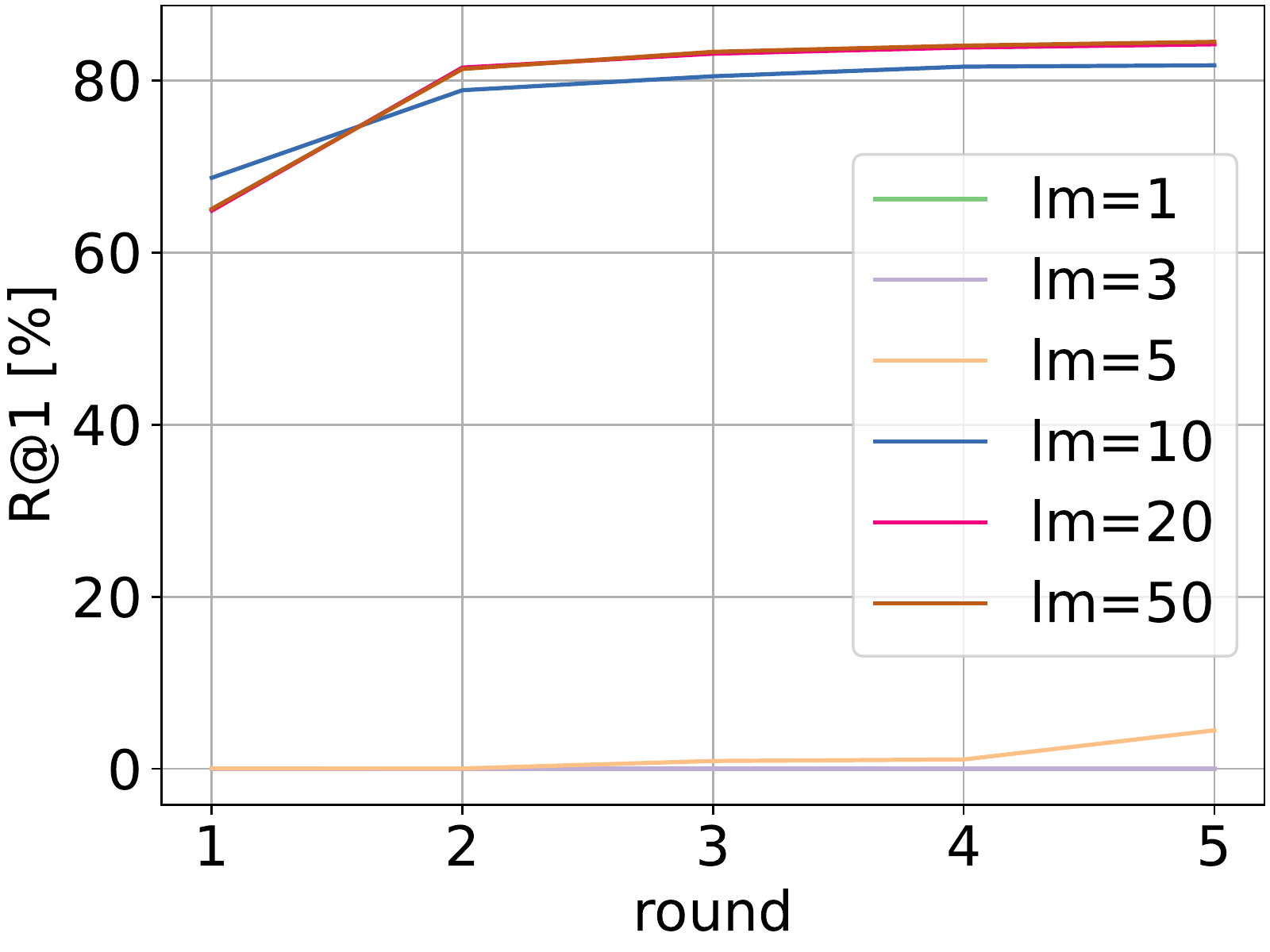}\kern.02\textwidth
    \includegraphics[width=0.49\textwidth]{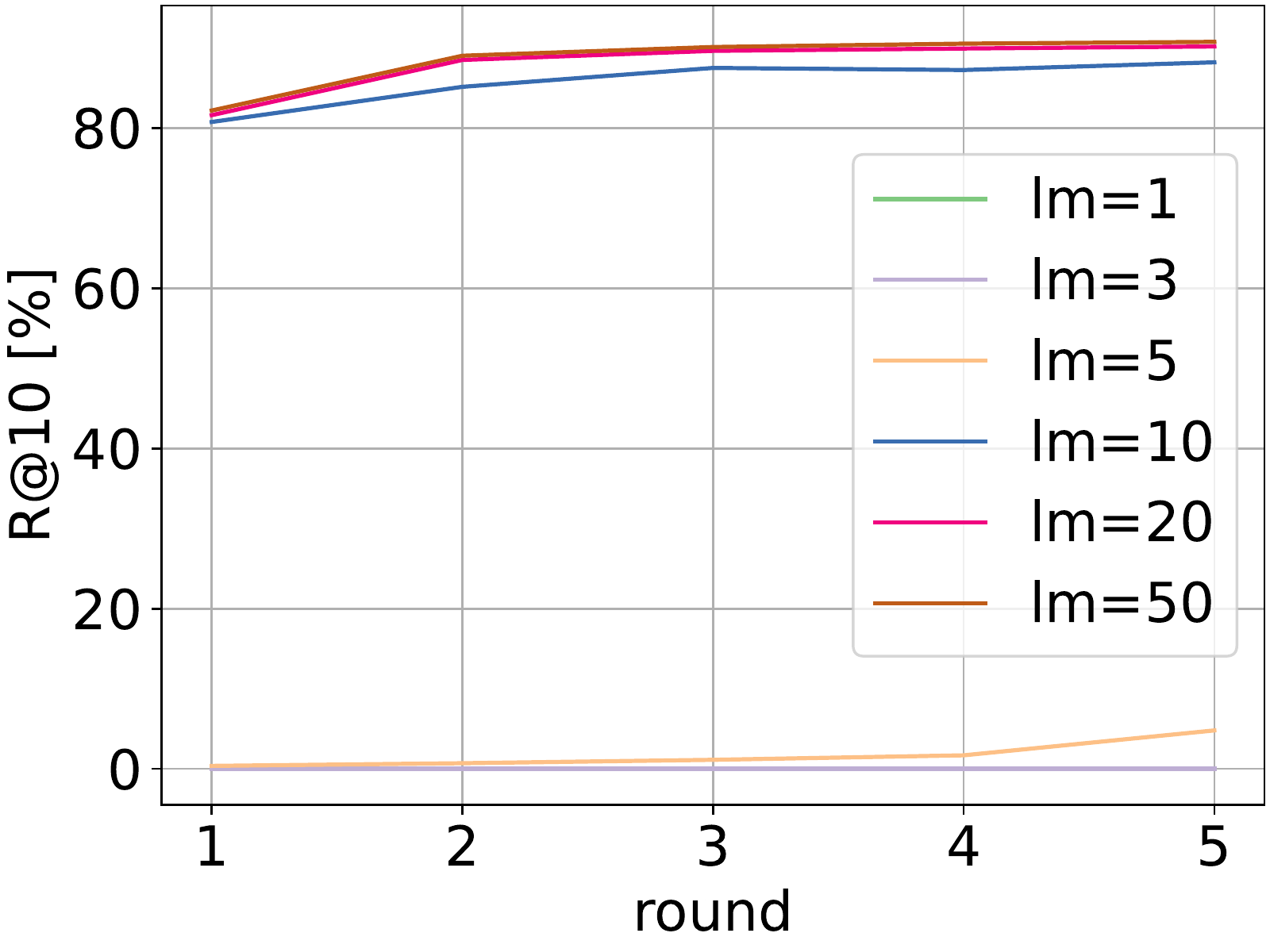}
    \caption{Results of experiments with varying logit multiplier. We try $6$ values $(1, 3, 5, 10, 20, 50)$ and evaluate models at R@1 and R@10 after each fine-tunig round. For smaller multipliers, the model does not learn anything meaningful. The model learns when we employ values of $10$ or larger. We see a slight improvement when increasing the value from $10$ to $20$. An increase above $20$ does not produce a difference.
    }
    \label{fig:logit_multipliers}
\end{figure}

In \Cref{subsub:loss} we explain why scaling the cosine similarities helps the model to learn.
Here we present results (\Cref{fig:logit_multipliers}) of several experiments with varying logit multipliers.

The takeaways from our experiments are the following:
\begin{itemize}
    \item Without a suitable logit multiplier, the model fails to learn.
    \item The exact value of the logit multiplier is not crucially important. It is only required for it to be sufficiently high.
\end{itemize}

Our results demonstrate that for multipliers $\leq 5$, the model cannot learn, and the training tends to converge to a point where the model predicts a vector of $1$ and $-1$ no matter the input.
We report success with values $\geq 10$.
We see a slight improvement when increasing the value from $10$ to $20$.
Increase above $20$ does not bear more fruit.

\subsection{Size}\label{subsec:size_finetune}
In \Cref{subsec:model_compar} we showed that the number of parameters affects the quality of generated embeddings.
Here we explore the same relation, but this time we fine-tune our model.
Model size is likely to affect its learning capacity during fine-tuning. Consequently, we expect that the models with more parameters should exhibit superior performance. 
However, as noted in \Cref{subsec:model_compar}, the largest model underperforms significantly on OLPEAT. 
Thus, the relationship between model size and performance may be more complex than initially assumed.

\begin{figure}[t]
    \centering
    \includegraphics[width=0.49\textwidth]{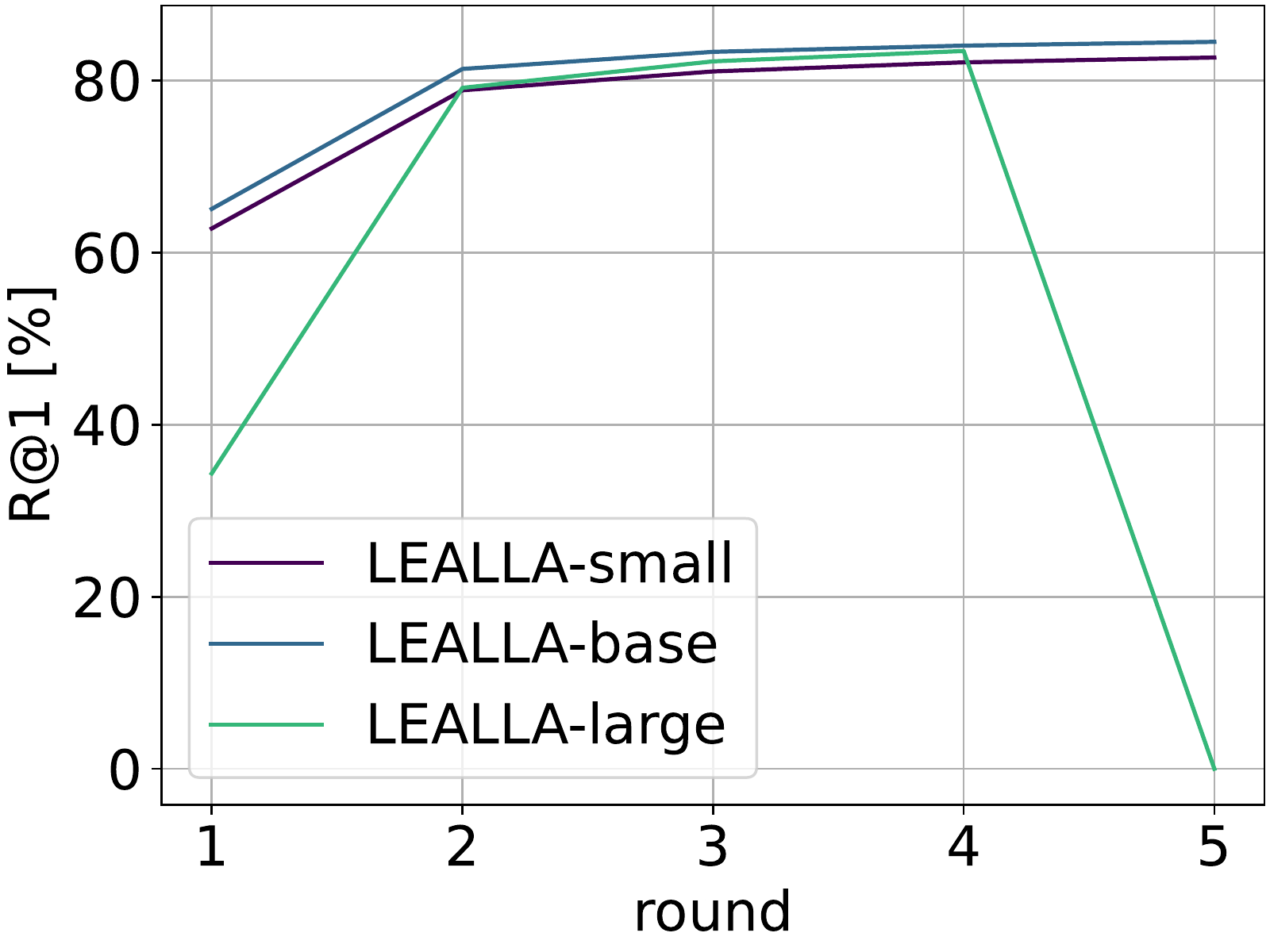}\kern.02\textwidth
    \includegraphics[width=0.49\textwidth]{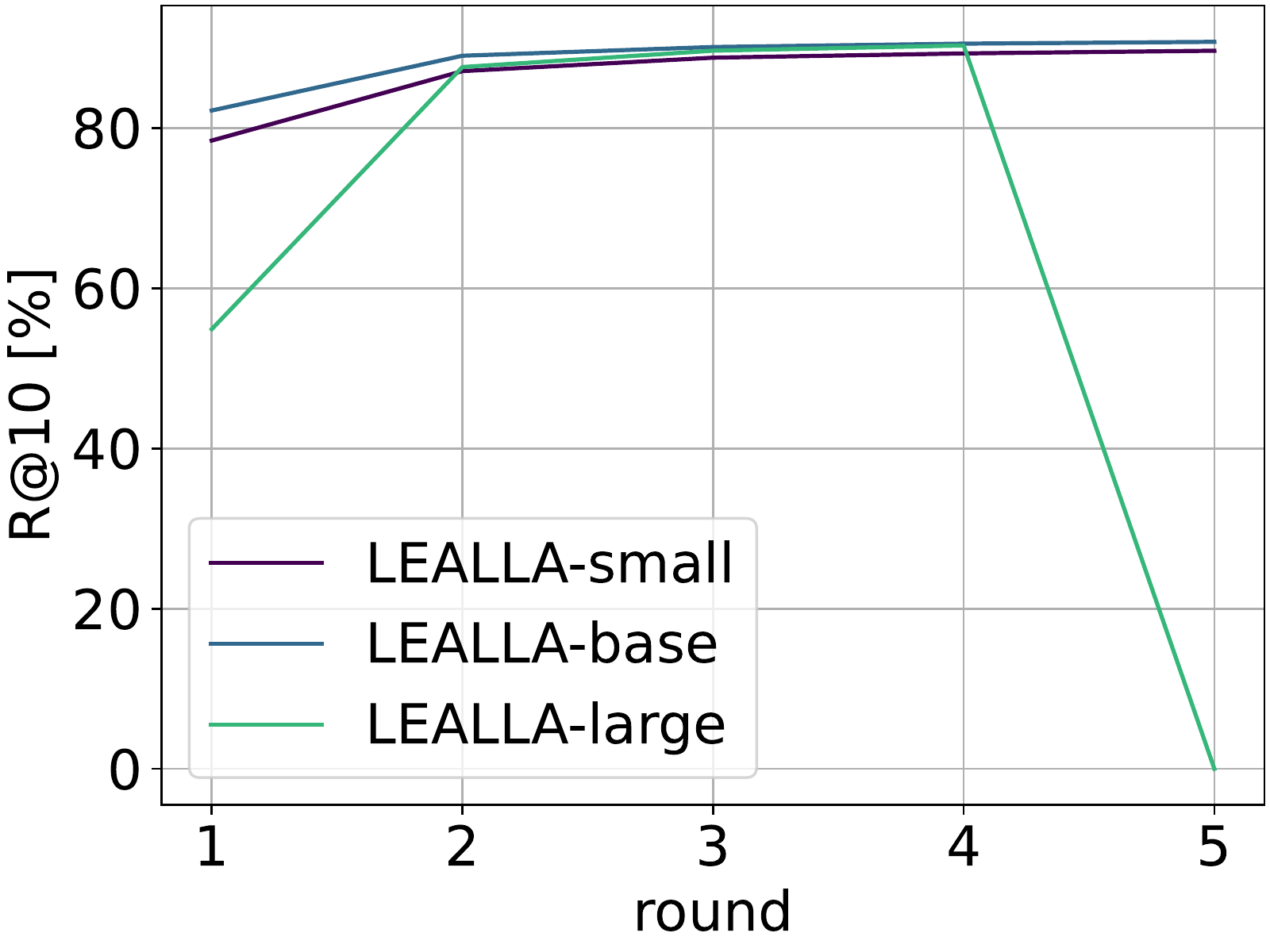}
    \caption{Results of experiments with three variations of the LEALLA model: small, base, and large. R@1 and R@10 were evaluated after each round. The results show that while all models perform comparably, the LEALLA-small is slightly weaker. Intriguingly, the LEALLA-large, despite its initial struggle, shows potential to outperform the base model in later rounds. However, its training stability remains a challenge. In our experiment, the model diverged in the last round and started predicting essentially the same embeddings for all inputs.}
    \label{fig:sizes}
\end{figure}

We present our results in \Cref{fig:sizes}.
All the models are comparable. LEALLA-small is slightly weaker than the other two, and it appears the the LEALLA-large could overtake the base version in later rounds.
However, in our experiment, the loss of LEALLA-large exploded in the last round, destroying any progress.

\subsubsection{Problem With Training LEALLA-large}

For the curious reader, we present a thorough discussion of problems of the
large model below.
Our results indicate that LEALLA-large is significantly more challenging to train than the rest of the herd.
It initially struggles during the first phase of learning, but after some time it can perform comparably to other models.
However, the training is sensitive to parameter choice, and the model can easily diverge. 
To improve its stability, we implement a linear learning rate warm-up~\citep{goyal2018accurate}, along with decreasing the final learning rate. 
These modifications result in either negligible effects or failure to converge, depending on the exact parameters we try.

\begin{figure}
\centering
\includegraphics[width=.8\linewidth]{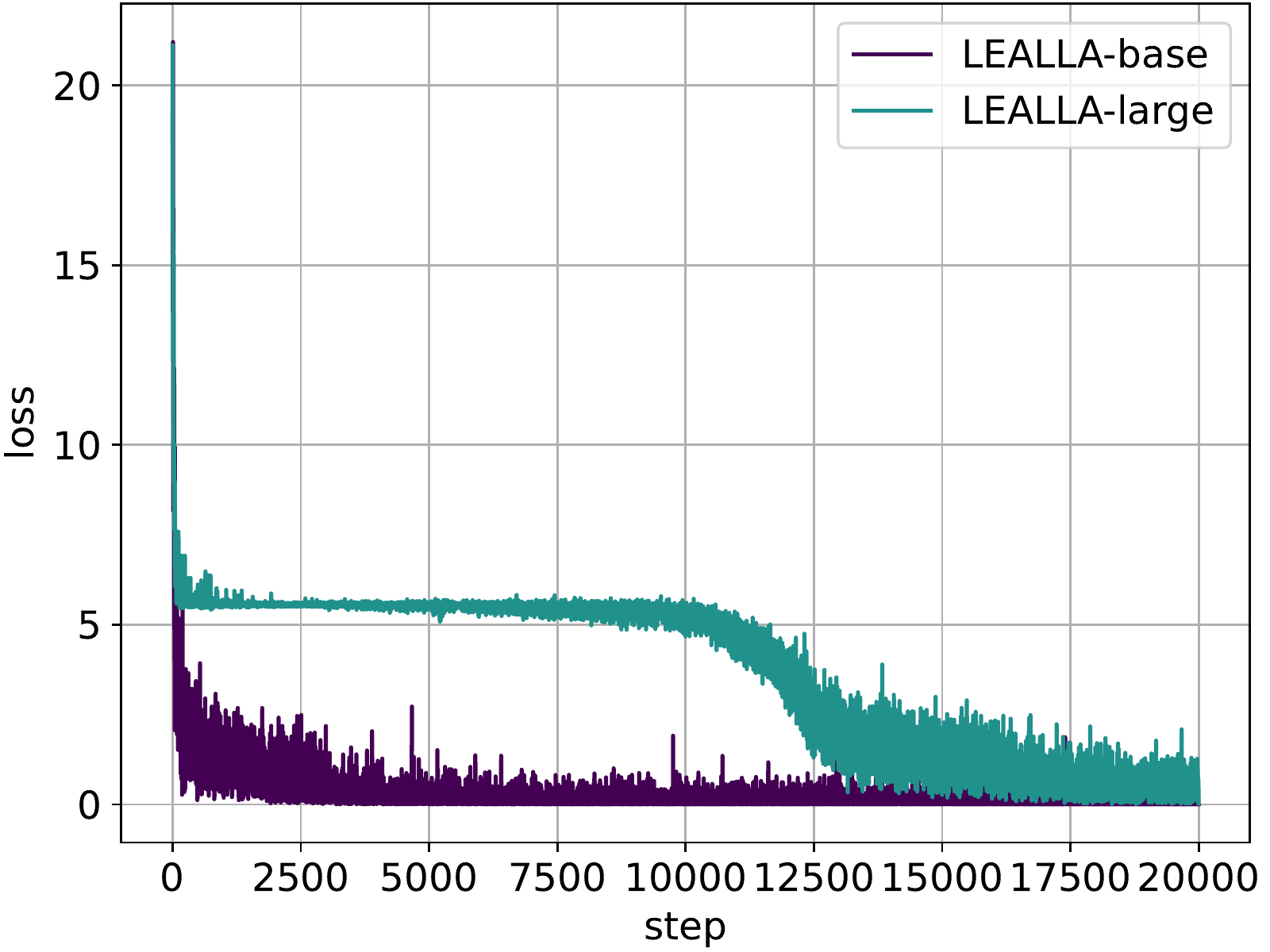}
  \caption{Comparison of loss of LEALLA-base and LEALLA-large during the first round of training. Whereas the base model learns without much difficulty, the large one hovers above $5$ for more than 10000 updates before successfully solving the task.}
\label{fig:size_losses}
\end{figure}

In \Cref{fig:size_losses} we present the comparison of losses of LEALLA-base and LEALLA-large during the first round.
We see that LEALLA-large plateaus slightly above the value of $5$.
We note that this is a loss that the model would achieve if it created a uniform distribution on the output because $5.5 \approx - \log{1/(32 \cdot 8)}$ --- assuming the batch size of $32$ and $1 + 7$ examples per a mention in the batch.
This behavior is unique to the first round; in later rounds, both loss and recall rates for LEALLA-large improve consistently, mirroring the behavior seen in other models.
This whole phenomenon can be explained by the fact that without any training, the large model is weaker at our task (\Cref{subsec:model_compar}) and needs some time to catch up.

The crash in recalls during the later stages of training coincides precisely with the point at which we begin to feel victorious. 
It appears that even a small number of noisy batches can significantly disrupt the training of LEALLA-large in our configuration. 
If this happens, the model can quickly revert to its championed loss value of $5.5$, wiping any understanding it acquired.
While we could attempt to stabilize the training by experimenting with the parameters a little bit more, we have decided not to pursue it to avoid over-optimizing our parameters for a single model.

\section{Cross-Lingual Transfer}
\label{sec:cros_lingual_transfer_res}

In \Cref{sec:cross_lingual_transfer} we explain cross-lingual transfer learning and discuss how it can be used to save computational resources by training only on set of languages that is smaller than the one on which we are evaluating.
In this section, we conduct an experiment with a model trained on Spanish and evaluate it on the complete Mewsli-9 benchmark.

We start fine-tuning from the checkpoint corresponding to the last model from \Cref{subsec:index_rebuilding_res}.
This model has a very good understanding of EL in Spanish, however, in \Cref{sec:hyperparameter_search} to save time we fine-tune models only on a KB that contains entities with a corresponding Wikipedia page.
Here, we are aiming for the best recalls, thus, we include \textit{all Spanish entities}.
Subsequently, we fine-tune the model for three more rounds, so that it encounters some of the newly added entities as negative examples.

\begin{table}[t]
  \centering

  \begin{tabular}{lc}
    \hline
    context size & \064\\
    rounds & \0\03 \\
    steps per round & \010\rlap{$^5$}\\
    logit multiplier & \050\\
    retrieval parameter $k$ & 100\\
    $\mathit{neg}$ & \0\07\\
    batch size & \032\\
    starting learning rate & \010\rlap{$^{-5}$}\\
    learning rate step decay & ~~~~~~\0\00\rlap{.998}~~~~~~\\
    \hline
  \end{tabular}
  \caption{Parameters for the cross-lingual experiment.}
  \label{tab:cross_params}
\end{table}

We employ a fairly standard set of hyperparameters (\Cref{tab:cross_params}) that is like those used in \Cref{sec:hyperparameter_search}.
Additionally, to stabilize convergence, we employ exponential learning rate decay in the ultimate round.
We start with a learning rate value of $10^{-5}$ and decrease it to $3\cdot 10^{-6}$ at the end of the training.

\subsection{Baseline With Labels}\label{subsec:cross_baseline}
Before we move to the experiments, we need to establish a baseline to which we can then compare our results.
Alias tables (\Cref{section:alias_table,subsec:at_damuel}) or OLPEAT (\Cref{subsec:we_alias_table,subsec:we_alias_table_res}) would not serve well because they are very different in their nature --- they can contain multiple different representations for one entity.
Contrary to them, all our neural network experiments have just one representation per entity.

We would therefore like to have a baseline that also adheres to one representation per entity. 
Therefore, for each entity, we take its label, embed it with LEALLA-base (without any fine-tuning), and use this embedding as the only representation of the entity.
When linking, we embed only the mention, ignoring the context, and look for the most similar labels.

As we did for OLPEAT (\Cref{subsec:we_alias_table_res}) and the string similarities (\Cref{subsec:ss_results}), we evaluate for each Mewsli-9 language separately.
We present our results together with the evaluations of the cross-lingual transfer in \Cref{tab:cross_per_lang}.

\begin{table}[t]
  \centering
  \catcode`!=13\def!{\bfseries}
  \begin{tabular}{lcccc}
    \hline\multirow{2}{*}{ Language} & \multicolumn{2}{c}{Baseline} & \multicolumn{2}{c}{CLT} \\
    \cline{2-5}
            &  R@1 &  R@10 &  R@1 &  R@10 \\
    \hline
        ar & 66.3 & 84.1 & !70.6 & !88.1\\
        de & 66.1 & 83.1 & !81.4 & !91.2\\
        en & 54.2 & 71.9 & !77.7 & !88.8\\
        es & 58.9 & 77.8 & !85.6 & !92.2\\
        fa & 66.5 & 82.4 & !80.6 & !90.7\\
        ja & 66.5 & 78.2 & !70.5 & !84.3\\
        sr & 58.2 & 88.4 & !87.3 & !94.5\\
        ta & 63.7 & 76.8 & !75.7 & !91.5\\
        tr & 71.4 & 85.4 & !81.1 & !92.2\\
    \hline
  \end{tabular}
  \caption{Results of cross-lingual transfer of model trained purely on Spanish EL to other languages. The results are compared to a simple baseline, for which we embedded only the labels with a default model.}
  \label{tab:cross_per_lang}
\end{table}

\subsection{Per-Language Evaluation}\label{subsec:clt_per}
Similarly to our previous experiments, we do a cross-lingual transfer learning evaluation for each language separately.

We present our results compared to the baseline in \Cref{tab:cross_per_lang} and to OLPEAT in \Cref{tab:cross_per_lang_we}.
We see that we can beat the baseline comfortably.
The model improves a lot even on the languages it did not see during training.
However, when we compare it to OLPEAT (thus far our strongest approach \Cref{subsec:we_alias_table_res}), we see that we must be careful with our optimism regarding the cross-lingual transfer.
In it, the OLPEAT outperforms cross-lingual transfer on R@1.
The results for R@10 are less decisive.
The fact that the bi-encoder works quite well on higher recalls could be further exploited.
For example, one could use our model to retrieve candidates, which can be then re-ranked with a simpler model.
We hypothesize that this might be a viable approach based on our results with $R@100$ and $R@1000$ (\Cref{section:clt_appendix}), which are noticeably close to the upper bounds established in \Cref{section:damuel_mewsli_intersection}.

\begin{table}[t]
  \centering
  \catcode`!=13\def!{\bfseries}
  \begin{tabular}{lcccc}
    \hline \multirow{2}{*}{Language}  & \multicolumn{2}{c}{ OLPEAT} & \multicolumn{2}{c}{ CLT} \\
    \cline{2-5}
            &  R@1 &  R@10 &  R@1 &  R@10 \\
    \hline
        ar & !90.1 & !93.3 & 70.6 & 88.1\\
        de & !88.8 & !92.9 & 81.4 & 91.2\\
        en & !80.1 & 88.2 & 77.7 & !88.8\\
        es & 83.9 & 90.3 & !85.6 & !92.2\\
        fa & !83.7 & 89.7 & 80.6 & !90.7\\
        ja & !85.2 & !92.6 & 70.5 & 84.3\\
        sr & !91.1 & !95.0 & 87.3 & 94.5\\
        ta & !91.2 & !95.5 & 75.7 & 91.5\\
        tr & !86.0 & !93.3 & 81.1 & 92.2\\
    \hline
  \end{tabular}
  \caption{Results of cross-lingual transfer of model trained purely on Spanish EL to other languages. We compare with OLPEAT (\Cref{subsec:we_alias_table,subsec:we_alias_table_res}) our strongest model yet.}
  \label{tab:cross_per_lang_we}
\end{table}

\subsection{Where To Go Next?}
Hitherto, we have seen that cross-lingual transfer is a decent approach but lacks OLPEAT, which has the advantage of having multiple different representations for one entity.
Nonetheless, we believe that our results with cross-lingual transfer can be further improved.
Therefore, here, we propose two ideas that could improve the results and conduct preliminary experiments.

\subsubsection{Entities from All Languages}
Perhaps limiting ourselves to entity descriptions in just one language is too constraining.
Thus, we propose creating one large KB and populating it with entity descriptions from all languages.

We are still working with our model trained only on Spanish; thus we prefer to use entities with Spanish descriptions.
Only when a Spanish description for an entity is not available, do we try to use one from a different language.
To do that, we must decide from what language we should take the entity.
Quite often we are going to face a situation, where an entity is not present in Spanish but exists in multiple other languages.
To select a language for entity description, we propose to use the same method as \citet{botha-100}, who rank the different-language representations based on the number of training mentions pointing to them.
The idea is that if there are a lot of links to an entity in some language, then the representation of that entity in the particular language also has high quality.
We discuss this approach in more detail in \Cref{subsub:el100}.

To get an idea of the strength of the proposed approach, we limit ourselves only to the nine Mewsli languages during the KB construction.
For our KB we take all Spanish entities and add to it all entities that do not have a Spanish DaMuEL entry but are present in any of the remaining eight languages.
When an entity has more than one representation in the eight languages, we use the one favored by the heuristic from \citet{botha-100}.
We use this KB to evaluate all the languages of Mewsli-9.

We present our results in \Cref{tab:cross_el100}.
We see that the approach that mixes multiple languages is significantly worse everywhere except for Spanish.
This leads us to a hypothesis that the model cannot link from a mention in one language to an entity in another.
This is plausible because it was trained to link only inside one particular language (Spanish to Spanish).
Nevertheless, a more in-depth evaluation is needed to fully verify it.

\begin{table}[t]
  \centering
  \catcode`!=13\def!{\bfseries}
  \begin{tabular}{lcccc}
    \hline \multirow{2}{*}{Language}  & \multicolumn{2}{c}{CLT multiple languages} & \multicolumn{2}{c}{CLT} \\
    \cline{2-5}
            & R@1 & R@10 & R@1 & R@10 \\
    \hline
        ar & 60.4 & 76.2 & 70.6 & 88.1\\
        de & 70.8 & 82.6 & 81.4 & 91.2\\
        en & 68.2 & 80.4 & 77.7 & 88.8\\
        es & 85.1 & 92.2 & 85.6 & 92.2\\
        fa & 50.7 & 75.7 & 80.6 & 90.7\\
        ja & 29.4 & 44.5 & 70.5 & 84.3\\
        sr & 72.2 & 87.6 & 87.3 & 94.5\\
        ta & 56.0 & 74.5 & 75.7 & 91.5\\
        tr & 67.1 & 80.4 & 81.1 & 92.2\\
    \hline
  \end{tabular}
  \caption{Cross-lingual transfer approach with different knowledge bases. For the first two recalls, the KB is constructed by taking all Spanish entities and adding to them entities that are missing in the Spanish part of DaMuEL. The CLT recalls use a KB that is dependent on language. For CLT, we construct the KB from all entities present in the specific part of DaMuEL corresponding to the language. Our results show that constructing the KBs per language is better. The model seems to have difficulties with linking to a KB in one language from mentions in another.}
  \label{tab:cross_el100}
\end{table}

\subsubsection{Training Contexts as Knowledge Base}
Utilizing contexts of training mentions as a KB can improve recalls over the standard approach, in which the descriptions and training mentions are separate~\citep{fitzgerald2022moleman}.
Our bi-encoder was designed with versatility in mind.
This allows for an intriguing experiment.
We can populate the KB with entity descriptions during training (which gives us an index of a reasonable size) and only during evaluation switch to the index populated with contexts of mentions.
We can do this because we use only one model to encode both entities and mentions, and the token used to mark an entity or a mention inside a text is the same for both.

Our preliminary results show that the recalls are improved by the addition of contexts.
We conduct per-language evaluations for Spanish and Turkish.
First, we populate the knowledge base \textit{only} with the contexts (without any descriptions), and we immediately see an improvement. 
For Spanish, we achieve R@1 and R@10 of $86.3$ and $92.2$, respectively.
In R@1 our score improved $0.7$ points compared to the version with descriptions.
For Turkish we achieve $85.9$ and $93.6$, respectively, slightly falling behind OLPEAT (\Cref{tab:pretrained_embeddings}) for R@1 and exceeding it at R@10.

Nevertheless, we can improve the score even more by populating the KB with \textit{both the training contexts and the entity descriptions}.
This yields an additional improvement.
We obtain $86.8$ in R@1 and $92.7$ in R@10 for Spanish. 
For Turkish we get $86.5$ and $94.4$, respectively.
With descriptions, we surpass OLPEAT even in Turkish, obtaining our best results yet.
Overall, with our flexible model design, we managed to improve our R@1s by $1.2$ percent points for Spanish and $5.4$ percent points for Turkish (compared to CLT, \Cref{tab:cross_per_lang_we}) \textit{with no additional training}.

\section{Comparing to the Current State of the Art}
Comparison with the best models of today is tricky, given the fact that we lack around $7 \%$ of entities in our training data (\Cref{section:damuel_mewsli_intersection}).
In \Cref{section:damuel_mewsli_intersection,subsec:at_damuel} we argue that the lack of entities is a lesser problem for R@1 than for R@10 because a lot of these entities are disambiguations, which are hard to link anyway.
Our results support this assumption.
We trail the current best result in Spanish~\citep{decao2021multilingual} by $3.3$ points in R@1 and $5.8$ in R@10~\citep{fitzgerald2022moleman}.
Our underperformance on other languages is greater because we did not train on them.

To get a full picture, it is essential to also compare the computation resources needed for training.
We train for $1120$k steps with a batch size of $32$ on a single NVIDIA GeForce GTX 1080 Ti GPU.\footnote{\scriptsize\url{https://www.nvidia.com/en-gb/geforce/graphics-cards/geforce-gtx-1080-ti/specifications/}}
This requires around six days.
The two approaches most similar to ours~\citep{botha-100, fitzgerald2022moleman} both train for a similar number of steps ($750$k and $500$k) but do so with batch size more than a hundred times larger ($8192$).
All their training is done on Google TPU v3 architecture.\footnote{\scriptsize\url{https://cloud.google.com/tpu/docs/v3}}
\citet{decao2021multilingual}, using a very different approach from bi-encoders, required over one thousand GPU days when training on Tesla V100 GPUs 32GB.\footnote{\scriptsize\url{https://www.nvidia.com/en-gb/data-center/tesla-v100/}}
In \Cref{tab:comparison_others} we provide a more structured comparison.

\begin{table}[t]
  \centering
  \catcode`!=13\def!{\bfseries}
  \begin{tabular}{lcccc}
    \hline
    Authors & R@1 & $b$ & Hardware & GPU/TPU days\\
    \hline
        this thesis & 85.6 & 32 & GTX 1080 Ti & 6 \\
        this thesis; mentions KB & 86.8 & 32 &  GTX 1080 Ti & 6 \\
        \citet{botha-100} & 89.0 & 8192 & TPU v3& $\sim 30$\\
        \citet{fitzgerald2022moleman} & 88.7 & 8192 & TPU v3& $\sim 20$\\
        \citet{decao2021multilingual} & 90.1 & --- & Tesla V100 & 1152\\
     \hline
  \end{tabular}
  \caption{A comparison of our approach to other works. For each, we provide the recall for Spanish, the batch size ($b$), the hardware, and estimates of the time needed. We do not provide the batch size for \citet{decao2021multilingual} because they utilize an entirely different approach. \citet{botha-100, fitzgerald2022moleman} note that they trained for less than two days on an unspecified number of TPUs. Subsequently, we cannot give a precise number of days required by their approach. The values provided are thus only our rough estimates based on the memory of TPUv3, known batch sizes, and our own experience.}
  \label{tab:comparison_others}
\end{table}

\section{To Fine-Tune or Not To Fine-Tune?}
So far, we have presented two different approaches.
The first is based solely on processing mentions, the second tries to also exploit the surrounding text.
Interestingly, the approaches based on mentions compare favorably with the more sophisticated fine-tuning.

There are two principal reasons why it is happening.
The first is that in entity linking the gap between good and excellent models is very small.
The primary usage of language is to communicate, thus it is not surprising that languages are optimized in a way that minimizes the number of entities that share a label.
Moreover, Mewsli-9 is sourced from news articles, which are meant to be unambiguous and understandable for a wide audience.
Subsequently, a simple baseline like an alias table performs well (\Cref{subsec:at_damuel}).
The second reason why the performance is comparable is that all the models based solely on mentions (\Cref{chap:baselines}) build their KB from training mentions, thus they have significantly more examples at their disposal.
Nevertheless, extending our model to utilize all training mentions as the knowledge base is possible, and our preliminary experiments suggest that it brings a decent performance boost.

To answer the question from the title, we recommend fine-tuning if one needs the extra performance boost.
Our results from \Cref{sec:cros_lingual_transfer_res} show that fine-tuning comfortably beats the other approaches in Spanish and can be further improved by enlarging the KB with training mentions.

Cross-lingual transfer is viable when the training data for some particular language is unavailable.
However, when the data are available cross-lingual transfer can be matched or even surpassed by simpler approaches that do not need fine-tuning.

\chapter{Conclusion}
Our goal in this thesis was to study entity linking and create models that are fast to train, operate in multiple languages, and are based solely on open-source data.

In the first half, we proposed three systems that are built solely on mentions, without the surrounding context.
We gave a detailed evaluation of two of them, the one based on string similarities and the one based on pre-trained embeddings. We showed that they easily outperform alias tables.
Perhaps surprisingly, they all scored well on Mewsli-9 despite their simplicity.
The beauty of both these systems is that they are easy to implement, and as we showed, capable of matching more sophisticated approaches.
Our results indicate that if we are willing to sacrifice a little bit of performance, it is possible to build entity linking systems without the need to set up the complex infrastructure required for fine-tuning.

Later, we fine-tuned a bi-encoder to disambiguate Spanish entities.
Our final model achieved a performance $3.3$ points worse in R@1 than the current state of the art on the Mewsli-9 benchmark. 
Even though we do not surpass the best results, our training time is one to two orders of magnitude smaller compared to the leading model.
Moreover, we argue that our results can be further improved by enlarging our knowledge base because the one built on top of DaMuEL lacked $6.6\%$ of Spanish Mewsli-9 entities.
Overall, our fine-tuning results demonstrate that it is possible to train an EL system close to SOTA using relatively limited resources.

Additionally, we examined cross-lingual transfer learning, which we consider unexplored for EL, and showed three different ways to use our Spanish model to evaluate on all Mewsli-9 languages.
The results of our preliminary evaluation with a KB populated with all training contexts and entity descriptions turned out to be very promising.
In it, we tested a novel idea in which a model is trained with a standard KB containing entity descriptions but evaluated with a KB enlarged with all training contexts.
Remarkably, the model can link to the contexts even though they are different from what was available during training.
Overall, using the enlarged KB only during evaluation allowed us to increase our Spanish R@1 by $1.2$ points with zero additional training cost.

Apart from evaluating multiple EL systems, we also conducted a detailed examination of various hyperparameters for training bi-encoders, hoping to make the work of anyone who comes after us a little easier.
Most notably, we discovered that the number of hard negatives is closely connected to batch size, and in some situations increasing the former over the latter can save significant computational resources.
Furthermore, we managed to show that it is possible to train with extremely small batch sizes, which opens up possibilities to train on commodity hardware.
To the best of our knowledge, we are also the first to use hard-negative mining from the start of the training; we achieve it due to a suitable choice of our model.

In addition, our work is the first comprehensive attempt to build an EL system on top of DaMuEL, which is a new dataset for this task.
Our work provides a clear picture of DaMuEL's current strengths and weaknesses.
Besides that, we discovered two minor flows, which were previously unnoticed: the exclusion of disambiguation Wikipedia pages, and the accidental removal of a small number of other entities.

In the future, there are multiple areas of research that we hope to explore.
First, we would like to train our model for more than one language.
To do that, we plan to update our infrastructure so that we can distribute the workload to multiple machines.
To allow others to directly use the models we train, we hope to incorporate our work into LinPipe --- a multilingual NLP framework that is currently in preparation.\footnote{\url{https://ufal.mff.cuni.cz/linpipe}}
Lastly, the authors of DaMuEL aim to address the discovered deficiencies in an upcoming release.
We hope to re-evaluate at least part of our experiments when the new version becomes available.

\def\bibfont{\hfuzz=2pt}

\printbibliography[heading=bibintoc]

\chapwithtoc{List of Abbreviations}

\begin{tabular}{ll}
    \abr{EL}{entity linking}
    \abr{NLP}{natural language processing}
    \abr{KB}{knowledge base}
    \abr{KG}{knowledge graph}
    \abr{MD}{mention detection}
    \abr{ED}{entity disambiguation}
    \abr{XEL}{cross-lingual entity linking}
    \abr{MEL}{multilingual entity linking}
    \abr{MIPS}{maximum inner product search}
    \abr{OLAT}{one language alias table}
    \abr{OLPEAT}{one language pre-trained embeddings alias table}
    \abr{CLT}{cross-lingual transfer}
\end{tabular}

\ifx\ThesisType\TypePhD
\chapwithtoc{List of Publications}
\fi

\appendix
\chapter{Additional Results}
\section{DaMuEL and Mewsli-9 Intersections}\label{appendix:intersection}
In \Cref{tab:damuel_mewsli_intersection} we upper bound recalls obtainable on Mewsli-9 when using DaMuEL. 
We point out that the recalls do not equate to the percentage of entities that are contained in both of the datasets because some entities in Mewsli-9 are linked multiple times.
Here, we provide the intersections that are calculated on the \textit{sets} of entities. 
Our results are in \Cref{tab:damuel_mewsli_intersection_per_part}.
When we compare this to our results from \Cref{tab:damuel_mewsli_intersection}, we see that the upper bounds on recalls are higher.
From this, we conclude that the entities that are present in DaMuEL are repeated in Mewsli-9 more often than the entities that are missing.

\begin{table}[t]
  \centering
  \begin{tabular}{lcc}
    \hline
     Mewsli-9 & \multicolumn{2}{c}{intersection} \\
    \cline{2-3}
     Language &  with specific (\%)       & with whole(\%) \\

    \hline
    ar & 94.6 & 94.6\\
    de & 93.5 & 93.9\\
    en & 93.2 & 93.2\\
    es & 93.3 & 94.0\\
    fa & 97.7 & 97.7\\
    ja & 95.6 & 95.8\\
    sr & 93.3 & 94.2\\
    ta & 97.3 & 97.4\\
    tr & 95.3 & 95.4\\
    \hline
  \end{tabular}
  \caption{Intersection between DaMuEL and Mewsli-9. In the second column, each Mewsli-9 language is intersected only with QIDs gathered from the specific DaMuEL part. In the third, QIDs from all of DaMuEL were gathered and then the intersections were computed.}
  \label{tab:damuel_mewsli_intersection_per_part}
\end{table}

\section{Alias Table}\label{appendix:alias_tables}

Here we provide more results for alias tables as described in \Cref{section:alias_table} and \Cref{subsec:at_damuel}.
We try three different settings:
\begin{itemize}
    \item In \Cref{tab:at_lowercase} we lowercase all aliases and queries.
    \item In \Cref{tab:at_higher_recalls} we repeat the alias table experiment from \Cref{subsec:at_damuel}, but this time we also provide R@100 and R@1000.
    \item DaMuEL contains also links that are not in Wikipedia but are added by
      a heuristic proposed by the authors. So far we always excluded those links
      to give a fair comparison to works that built directly on top of
      Wikipedia. In \Cref{tab:all_links}, we show how much would recalls change
      if we used \textit{all} links to create the aliases. 
\end{itemize}

\begin{table}[t]
  \centering
  \begin{tabular}{lcccc}
    \hline
     Language &  R@1 &  R@10 &  R@100 &  R@1000\\
    \hline
    ar &  87.4 &  89.7 &  89.8 & 89.8 \\
    de &  86.2 &  90.5 &  90.9 & 90.9\\
    en &  77.6 &  85.0 &  86.1 & 86.2\\
    es &  82.5 &  89.4 &  90.2 & 90.2\\
    fa &  72.9 &  76.8 &  77.0 & 77.0\\
    ja &  82.6 &  89.2 &  89.5 & 89.5\\
    sr &  87.4 &  90.6 &  90.8 & 90.8\\ 
    ta &  81.2 &  82.8 &  82.9 & 82.9\\
    tr &  72.2 &  81.6 &  81.8 & 81.8\\
    \hline
  \end{tabular}
  \caption{Results of an \textbf{uncased} alias table evaluated on Mewsli-9 with
  mentions from DaMuEL. R@1, R@10, R@100, and R@1000 are included. Aliases from
  DaMuEL were processed as outlined in \Cref{subsec:at_damuel} and then
  lowercased. The recalls are overall slightly better than values from our main
  cased variant. The only language that does not achieve the same or better recalls compared to the cased version is Turkish, where we see a notable drop of a few points in all columns.
}
  \label{tab:at_lowercase}
\end{table}

\begin{table}[t]
  \centering
  \begin{tabular}{lcccc}
    \hline
     Language &  R@1 &  R@10 &  R@100 &  R@1000\\
    \hline
        ar & 87.4 &  89.7 & 89.8 & 89.8\\
        de & 86.3 &  90.3 & 90.7 & 90.7\\
        en & 77.6 &  84.6 & 85.6 & 85.7\\
        es & 82.4 &  89.0 & 89.7 & 89.8\\
        fa & 72.9 &  76.8 & 77.0 & 77.0\\
        ja & 82.7 &  89.2 & 89.5 & 89.5\\
        sr & 87.2 &  90.3 & 90.5 & 90.5\\
        ta & 81.2 &  82.8 & 82.9 & 82.9\\
        tr & 81.7 &  88.3 & 88.5 & 88.5\\
    \hline
  \end{tabular}
  \caption{Results of standard alias table evaluated on Mewsli-9 with mentions from DaMuEL. Compared to results from \Cref{subsec:at_damuel}, here we also evaluate R@100 and R@1000. Aliases from DaMuEL were processed in the same way as described in \Cref{subsec:at_damuel}. Except for English and Spanish, there is no practical difference between R@100 and R@1000.}
  \label{tab:at_higher_recalls}
\end{table}

\begin{table}[t]
  \centering
  {
  \catcode`!=13\def!{\bfseries}
  \begin{tabular}{lcccccc}
    \hline
     \multirow{2}{*}{Language} & \multicolumn{2}{c}{DaMuEL wiki} & \multicolumn{2}{c}{DaMuEL all} \\
    \cline{2-5}
                     &   R@1       &  R@10      &  R@1       &  R@10\\ 
    \hline
    ar &  87.4  &  89.7 & !88.2  & !90.5 \\
    de & !86.3  &  90.3 &  85.8  & !90.5 \\
    en &  77.6  &  84.6 & !77.9  & !84.9 \\
    es & !82.4  &  89.0 & !82.4  & !89.2 \\
    fa &  72.9  &  76.8 & !75.0  & !78.1 \\
    ja & !82.7  &  89.2 &  82.5  & !89.4 \\
    sr &  87.2  &  90.3 & !88.3  & !91.3 \\
    ta & !81.2  &  82.8 &  80.9  & !84.1 \\
    tr &  81.7  &  88.3 & !82.8  & !88.9 \\

  \end{tabular}
  }
  \caption{Results for alias tables with automatically added entity mentions.
  \textbf{DaMuEL wiki} corresponds to aliases gathered from Wikipedia links,
  \textbf{DaMuEL all} also constructs aliases from Wikipedia links but enlarges
  them from new links added by the author's heuristic (\Cref{subsec:damuel}).
  Generally, the heuristic helps. We believe that this is mostly due to the fact
  that the heuristic expands links based on lemmatization. This can increase the number of aliases we have, thus allowing us to match previously unsolvable mentions. We report a clear improvement on R@10 for all languages. We also observe a slight improvement for R@1, although some languages give better recalls when only Wikipedia links are utilized.}
  \label{tab:all_links}
\end{table}

\section{Embeddings With Context and No Training}\label{appendix:context_embeddings}
During fine-tuning, we embed mentions with surrounding contexts and labels with their descriptions.
How well do these embeddings work when we do not fine-tune and immediately evaluate those from the LEALLA-base model?
This is an interesting question because were the embeddings random, the initial round of hard negative mining would not bear any fruit.
In \Cref{tab:embs_context_no_training}, we provide recalls that are achieved when embedding descriptions and using them as a KB  and then conducting a \textit{per language} evaluation with Mewsli-9.
We use the same tokenization as described in \Cref{sec:tokenization}.
Our results indicate that these embeddings are better than random ones but lag behind the trained ones.
The probable reason is that although descriptions and contexts share some similarities, it is generally unlikely that the context directly describes the mention. 
Thus, the similarity is often small, and the same is the R@1 metric. 

\begin{table}[t]
  \centering
  \begin{tabular}{lcc}
    \hline
     Language &  R@1 &  R@10\\
    \hline
    ar & 1.8 & 5.8\\
    de & 3.4 & 9.1\\
    en & 2.8 & 7.5\\
    es & 3.1 & 8.6\\
    fa & 3.3 & 7.8\\
    ja & 0.9 & 2.8\\
    sr & 1.5 & 4.5\\
    ta & 3.1 & 8.9\\
    tr & 2.8 & 8.0\\
    \hline
  \end{tabular}
  \caption{Recalls for the LEALLA-base model embeddings without fine-tuning, evaluated per language using the Mewsli-9 dataset. Embeddings were built from descriptions and contexts. The tokenization process used can be found in \Cref{sec:tokenization}. All recalls are small, which we hypothesize is caused by limited similarity between contexts and descriptions.
}
  \label{tab:embs_context_no_training}
\end{table}

\section{Cross-Lingual Transfer Additional Results}\label{section:clt_appendix}
We present our results of cross-lingual transfer for R@1, R@10, R@100, and R@1000 in \Cref{tab:clt_more_recalls}.

\begin{table}[t]
  \centering
  \begin{tabular}{lccccc}
    \hline
     Language
            &  R@1 &  R@10 &  R@100 &  R@1000 & upper bound \\
    \hline
        ar & 70.6 & 88.1 & 91.7 & 93.7 & 95.6\\
        de & 81.4 & 91.2 & 93.0 & 93.8 & 95.9\\
        en & 77.7 & 88.8 & 91.5 & 92.4 & 94.3\\
        es & 85.6 & 92.2 & 93.4 & 93.8 & 95.0\\
        fa & 80.6 & 90.7 & 94.2 & 97.4 & 98.3\\
        ja & 70.5 & 84.3 & 89.8 & 91.9 & 96.7\\
        sr & 87.3 & 94.5 & 96.5 & 96.9 & 97.3\\
        ta & 75.7 & 91.5 & 95.9 & 97.1 & 98.6\\
        tr & 81.1 & 92.2 & 93.9 & 94.4 & 96.0\\
    \hline
  \end{tabular}
  \caption{Cross-lingal transfer evaluated for each of the Mewsli-9 languages separately. We see that for higher recalls, the results get noticeably close to upper bounds established in \Cref{section:damuel_mewsli_intersection}.}
  \label{tab:clt_more_recalls}
\end{table}

\chapter{Examples of Spanish Links on Mewsli-9}
Below we present examples produced by the model from \Cref{subsec:clt_per}. 
All English translations were machine-generated with DeepL.\footnote{\url{https://www.deepl.com/translator}}

\renewcommand{\arraystretch}{1.3}
\setlength{\tabcolsep}{0.04\linewidth}
\begin{tabular}{p{0.85\linewidth}}
\textbf{Original}: ``este de Huetamo, Michoacán. Hasta el momento, no se reportan daños materiales ni pérdidas humanas. Los sistemas y servicios no se vieron afectados en la mayor parte de la ciudad. Sin embargo, el \textbf{[M]~Metro~[M]} suspendió transitoriamente sus actividades y diversas oficinas y centros de trabajo fueron evacuados.''\\

\textbf{English translation}: ``east of Huetamo, Michoacán. So far, no material damage or human losses have been reported. Systems and services were not affected in most of the city. However, the \textbf{[M]~Metro~[M]} temporarily suspended its activities and several offices and work centers were evacuated.''\\

\vspace{-.3em}

\textbf{Predicted}: Metro de la Ciudad de México (Q735042)\\

\textbf{Correct}: Metro de la Ciudad de México (Q735042)\\

\hline

\textbf{Original}: ``Chávez. " " Él está haciendo algo muy valioso para nosotros " ", dijo Hassan Bzaih, libanés propietario de un almacén en \textbf{[M] Isla Margarita [M]} y quien visitó su país a finales de julio. Mientras tanto, la comunidad judía reaccionó con indignación. " " Estamos''\\

\textbf{English translation}: ``Chavez. `He is doing something very valuable for us', said Hassan Bzaih, Lebanese owner of a store in \textbf{[M] Isla Margarita [M]} and who visited his country at the end of July. Meanwhile, the Jewish community reacted with indignation. `We are''\\

\vspace{-.3em}

\textbf{Predicted}: Isla de Margarita --- Venezuela (Q334738)\\

\textbf{Correct}: Isla Margarita --- disambiguation page (Q1182916)\\

\hline

\textbf{Original}: ``( 6 ) en su programa semanal ` Aló Presidente ', en el que dijo que el Estado de Israel era el culpable del nuevo \textbf{[M]~Holocausto~[M]} con el apoyo de los Estados Unidos, país al que calificó de terrorista . Chávez había ordenado el viernes el retiro''\\

\textbf{English translation}: ``( 6 ) in his weekly program `Aló Presidente', in which he said that the State of Israel was guilty of the `new \textbf{[M]~Holocaust~[M]}' with the support of the United States, country which he qualified as `terrorist'. Chavez had ordered on Friday the withdrawal of the''\\ 

\vspace{-.3em}

\textbf{Precited}: Holocausto (término) (Q881997)\\

\textbf{Correct}: Holocausto (Q2763)\\

\end{tabular}

\begin{tabular}{p{0.85\linewidth}}
\textbf{Original}: ``Esperamos que estas medidas, tomadas bajo revisión por el gobierno, sean necesarias sólo por tiempo limitado, reza un comunicado emitido por el \textbf{[M] Departamento de Transporte [M]}. Mientras tanto, el Departamento de Seguridad Interna de los Estados Unidos ha elevado el nivel de amenaza terrorista en los vuelos desde y hacia el Reino Unido''\\

\textbf{English translation}: ``We expect that these measures, taken under review by the government, will only be necessary for a limited time, reads a statement issued by the \textbf{[M]~Department for Transport~[M]}. Meanwhile, the U.S. Department of Homeland Security has raised the terror threat level on flights to and from the United Kingdom.''\\

\vspace{-.3em}

\textbf{Predicted}: Departamento de Transporte del Reino Unido (Q2982287)\\

\textbf{Correct}: Departamento de Transporte del Reino Unido (Q2982287)\\

\hline

\textbf{Original}: ``2 ) al ejército de su país que se oponga a la transición de poder tras la cesión de las funciones de Fidel Castro a su hermano \textbf{[M] Raúl [M]} por la enfermedad del primero. Jorge Mas Santos, presidente de la junta directiva de la FNCA, declaró que " " los militares tienen la oportunidad de''\\

\textbf{English translation}: ``2 ) his country's military to oppose the transition of power following the transfer of Fidel Castro's functions to his brother \textbf{[M] Raul [M]} due to the former's illness. Jorge Mas Santos, president of the board of directors of the FNCA, declared that the military has the opportunity to''\\

\vspace{-.3em}

\textbf{Predicted}: Raúl Castro (Q46809)\\

\textbf{Correct}: Raúl Castro (Q46809)\\
\end{tabular}

\end{document}